\pdfoutput=1
\documentclass[11pt]{article}

\usepackage[final]{acl}

\usepackage{times}
\usepackage{latexsym}
\usepackage{enumitem}
\usepackage{caption}
\usepackage[T1]{fontenc}
\usepackage[utf8]{inputenc}

\usepackage{microtype}

\usepackage{inconsolata}

\usepackage{graphicx}

\usepackage{multirow}
\usepackage{graphicx}
\usepackage{algorithm}
\usepackage{algorithmic}
\usepackage{booktabs}
\usepackage{tcolorbox}
\usepackage{amssymb}
\usepackage{amsmath}
\usepackage{array}
\usepackage{longtable}
\usepackage{xcolor}
\usepackage{makecell}
\usepackage{geometry}
\usepackage{longtable}
\usepackage{booktabs}
\usepackage{tabularx}

\renewcommand{\arraystretch}{1.2}

\title{Why Synthetic Isn't Real Yet: A Diagnostic Framework for Contact Center Dialogue Generation}

\author{
  Rishikesh Devanatha, Varun Nathan, Ayush Kumar \\
  \texttt{\{rishikesh.devanathan, varun.nathan, ayush\}@observe.ai} \\
  Observe.AI \\ Bangalore, India
}

\begin{document}
\maketitle
\begin{abstract}

Synthetic data is increasingly critical for contact centers, where privacy constraints and data scarcity limit the availability of real conversations. However, generating synthetic dialogues that are realistic and useful for downstream applications remains challenging. In this work, we benchmark multiple generation strategies guided by structured supervision on call attributes (Intent Summaries, Topic Flows, and Quality Assurance (QA) Forms) across multiple languages. To test downstream utility, we evaluate synthetic transcripts on an automated quality assurance (AutoQA) task, finding that prompts optimized on real transcripts consistently outperform those optimized on synthetic transcripts. These results suggest that current synthetic transcripts fall short in capturing the full realism of real agent–customer interactions. To highlight these downstream gaps, we introduce a diagnostic evaluation framework comprising 17 metrics across four dimensions: (1) Emotional and Sentiment Arcs, (2) Linguistic Complexity, (3) Interaction Style, and (4) Conversational Properties. Our analysis shows that even with structured supervision, current generation strategies exhibit measurable deficiencies in sentiment fidelity, disfluency modeling, behavioral variation, and conversational realism. Together, these results highlight the importance of diagnostic, metric-driven evaluation for synthetic conversation generation intended for downstream applications.

% Synthetic transcript generation is critical in contact center domains, where privacy and data scarcity limit model training and evaluation. Unlike prior synthetic dialogue generation work on open-domain or medical dialogues, contact center conversations are goal-oriented, role-asymmetric, and behaviorally complex, featuring disfluencies, ASR noise, and compliance-driven agent actions. In deployments where transcripts are unavailable, standard pipelines still yield derived call attributes such as Intent Summaries, Topic Flow, and QA Evaluation Forms. We leverage these as supervision signals to guide generation. To assess the quality of such outputs, we introduce a diagnostic framework of 18 linguistically and behaviorally grounded metrics for comparing real and synthetic transcripts. We benchmark four language-agnostic generation strategies, from simple prompting to characteristic-aware multi-stage approaches, alongside reference-free baselines. Results reveal persistent challenges: no method excels across all traits, with notable deficits in disfluency, sentiment, and behavioral realism. Our diagnostic tool exposes these gaps, enabling fine-grained evaluation and stress testing of synthetic dialogue across languages.

% Methods for synthetic data generation has been studied but the quality is not assessed 

\end{abstract}

\section{Introduction and Related Work}

Contact center environments generate massive volumes of dialogue that remain largely inaccessible for model development due to privacy constraints. Synthetic data offers a solution to this scarcity, with the potential to serve as a reliable foundation for training and optimization. Success in other domains, such as medical summarization \citep{binici2025medsageenhancingrobustnessmedical}, where synthetic data improved robustness of the summaries by 16.4\%, suggests that utility is possible when noise and errors are modeled effectively. However, capturing the linguisitic, conversational and behavioral realism of goal-oriented interactions remains a significant hurdle.

The difficulty stems from the fact that realism in contact center conversations is multi-dimensional, extending far beyond surface fluency. Authentic dialogues are structured by the interplay of several key dimensions: (1) \textbf{Emotional and Sentiment Arcs}, where real calls exhibit gradual escalation, de-escalation, and modulation of tone, with skilled agents actively managing customer affect, rather than remaining sentimentally flat or shifting abruptly; (2) \textbf{Linguistic Complexity}, where authentic conversations mix compliance-heavy phrasing and technical detail with colloquial reassurance, reflecting a natural balance between lexical richness and accessibility; (3) \textbf{Interaction Style}, where the subtle negotiation of control is encoded through initiative balance, question types, and politeness strategies, with customers who might interrupt or drive the agenda; and (4) \textbf{Conversational Properties}, where surface-level realism is conveyed through disfluencies, false starts, and ASR-induced artifacts that signal turn-taking irregularities.

These dimensions underscore why \textbf{plausibility is not realism}. Current methods are adept at producing plausible text, but they fail to capture the affective, linguistic, interactional, and procedural dynamics that define authentic conversations. Recent works such as NoteChat \citep{Wang_2024} and ConvoGen \citep{gody2025convogenenhancingconversationalai} synthesize dialogues using a multi-agent setup, while others target multi-turn or speech-level synthesis \citep{suresh2025diasynthsyntheticdialoguegeneration, wang2025speechdialoguefactorygeneratinghighqualityspeech}. While these approaches span multiple domains, they are \textbf{not tailored to the goal-oriented, behaviorally rich, and acoustically noisy environments of contact centers}, which demand fidelity to all dimensions of realism.

The limitations of current methods are exacerbated by a lack of diagnostic evaluation. Conventional metrics like BLEU \citep{bleu} reward lexical overlap but cannot surface the underlying representational failures that lead to downstream degradation. While prior works propose targeted evaluations for specific phenomena like fillers \citep{hassan2024enhancingnaturalnessllmgeneratedutterances} or ASR noise \citep{binici2025medsageenhancingrobustnessmedical}, they do not assess realism comprehensively. This work makes three primary contributions:
\begin{enumerate}[leftmargin=*, topsep=0pt, noitemsep=0.45mm]
    \item We introduce a \textbf{diagnostic evaluation framework} comprising 17 metrics across the core dimensions of realism to quantify where synthetic dialogues diverge from authenticity by comparing their distribution against real transcripts. \footnote{Code for evaluation framework attached as zip file.}
    
    \item We conduct a \textbf{downstream-task experiment} to measure the efficacy of synthetic transcripts for prompt optimization, demonstrating a significant performance gap compared to real transcripts.

    \item We \textbf{elucidate specific quality gaps} by benchmarking five generation strategy of increasing levels of structured supervision, across four languages. Our analysis identifies core challenges and contributes evidence that existing synthetic dialogue generation approaches remain insufficient for producing realistic call center conversations, thereby motivating further investigation by the scientific community.
\end{enumerate}

\section{Methodology}

\begin{figure*}[t]
    \centering
    \includegraphics[width=0.85\textwidth]{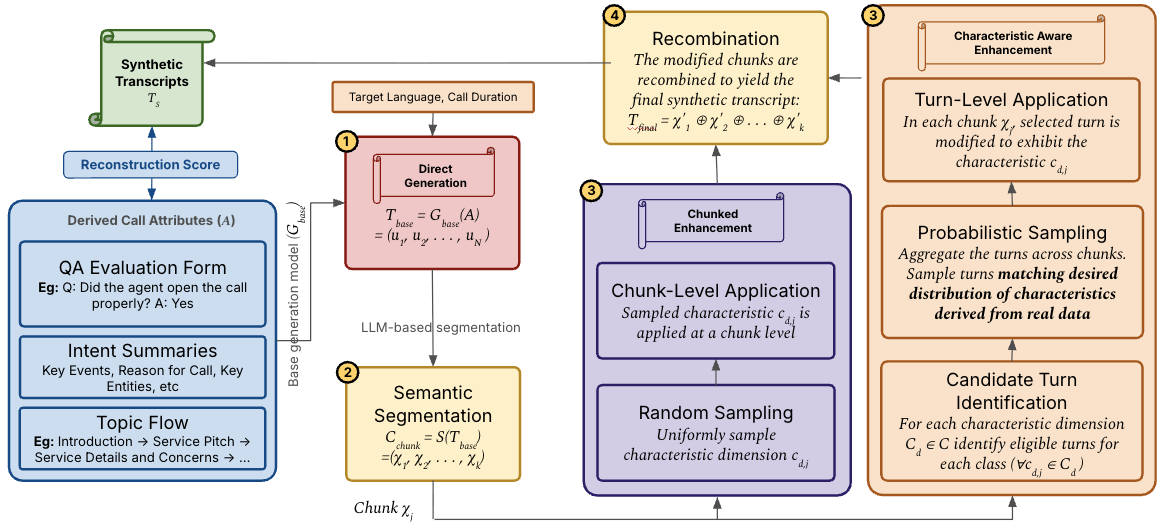}
    \vspace*{-3.2mm} % Add negative vertical space here
    \caption{Diagram depicting the three generation strategies. Section in red depicts Direct generation where transcripts are generated by simply conditioning on the input call attributes \((A)\). Sections in purple and orange depict the Chunked and Characteristic Aware enhancement stages respectively.}
    \label{fig:pipeline_diagram}
\end{figure*}

In this section, we briefly outline the structured attribute-guided generation strategies used for experimentation and the evaluation methodology designed to measure the fidelity of the synthetic data.

\subsection{Generation Methodology}
\label{generation_methodology}
To address challenges in synthetic contact center transcript generation and limitations of existing methods, our pipeline conditions generation on modular, interpretable supervision signals derived from real transcripts and routinely produced in call center operations: intent-specific summaries \citep{probing-cc-llm}, quality assurance (QA) forms \citep{probing-qa}, topic flow \citep{calls-enhancing-segmentation}. These signals, along with the target call length and language, constitute the input call attributes \((A)\) for our generation strategies. Using these structured attributes serves two key purposes: they act as privacy-preserving proxies, mitigating the need to handle raw, sensitive transcript data, and they ground the synthetic dialogues in the operational realities of the call center domain. We later perform an ablation study to examine the impact of these attributes on the realism of the generated data. Examples of these attributes are shown in Tables~\ref{examples_dual_stage_v1_1} - \ref{examples_dual_stage_v1_2}, and detailed descriptions are provided in section~\ref{desc_struc_inputs}.

\label{transcript_gen_pipeline}
As shown in Figure \ref{fig:pipeline_diagram}, we propose three generation strategies of increasing procedural depth, with each successive variation designed to incrementally enhance the realism of authentic dialogues through structured supervision and iterative refinements. The pipelines are as follows:

\subsubsection{Direct Generation}
\label{single-stage-generation}
Referred to as \textbf{Direct} in Tables~\ref{results_sent_and_emotion_english}-~\ref{results_conversational_properties_english} and ~\ref{results_sent_and_emotion_all_languages}-~\ref{results_conversational_properties_all_languages}, this strategy employs an LLM \( G_{\text{base}} \) to directly generate transcript \( T_{\text{base}} \) by conditioning on the input call attributes \((A)\) with simple prompt-level instructions to simulate disfluencies, ASR noise, and other call center-specific characteristics in a direct prompt. This straightforward approach provides a foundational transcript that serves as the baseline and upon which subsequent refinements are made.

\subsubsection{Chunked Enhancement}

Referred to as \textbf{Chunked} in Tables~\ref{results_sent_and_emotion_english}-~\ref{results_conversational_properties_english} and ~\ref{results_sent_and_emotion_all_languages}-~\ref{results_conversational_properties_all_languages}, this pipeline addresses limitations of direct single-pass generation in being able to generate long transcripts adhering to all the call attributes and instructions, by segmenting and enhancing the base transcript in smaller, semantically coherent chunks. 

The base transcript is divided into chunks \( \mathcal{C}_{\text{chunk}} = (\chi_1, \chi_2, \ldots, \chi_k) \) using LLM-derived boundaries. For each chunk \( \chi_j \), the characteristic dimension \( C_d \in \mathcal{C} \) (e.g., Sentiment, Question Type), are sampled uniformly and applied at the chunk level: this includes adding speech disfluencies, introducing plausible ASR errors, turns with specified sentiment type, yielding enhanced chunk \(\chi'_j = E(\chi_j, D_j)\). The final transcript is the concatenation \(T_{\text{final}} = \chi'_1 \oplus \chi'_2 \oplus \ldots \oplus \chi'_k\). 

This method produces transcripts that more accurately capture the noisy, imperfect, and interactive nature of real-world spoken dialogues by mitigating drift through localized enhancements. 

\subsubsection{Characteristic-Aware Enhancement}

Referred to as \textbf{Characteristic Aware} in Tables~\ref{results_sent_and_emotion_english}-~\ref{results_conversational_properties_english} and ~\ref{results_sent_and_emotion_all_languages}-~\ref{results_conversational_properties_all_languages}, this pipeline is a more targeted enhancement strategy that aligns the base transcript's turn-level features with those observed in real-world data. Unlike the chunk-level enhancements and uniform sampling of disfluencies in the previous method, this applies the characteristics at the turn-level through a controlled process involving candidate turn identification, probabilistic sampling based on real-data distributions, and targeted rewriting. This enables fine-grained control over stylistic and structural aspects of the output. Full details are provided in Section~\ref{characteristic_aware_pipeline}.

\noindent The prompts used for the different pipelines can be found in Tables \ref{prompts_for_transcript_generation_1} and \ref{examples_dual_stage_v1_2}.

\section{Evaluation Framework}

Here, we outline the methodology for evaluating and quantifying the distributional alignment of the proposed metrics between real transcript ($\mathcal{T}_R$) and a synthetic, generated transcript ($\mathcal{T}_S$). Prior work often represents entire conversations using generic vector embeddings \citep{lavi2021weveconversationbeforenovel} and compares them using cosine similarity, which can obscure fine-grained conversational structure and behavior specifically relevant to dialogue realism; our framework instead decomposes conversations along interpretable dimensions that more directly capture conversational phenomena. To assess the quality and realism of the conversations, we propose a comprehensive, multi-dimensional evaluation framework that goes beyond traditional lexical similarity metrics. The framework spans four core dimensions, with metrics applied at both the turn and transcript level:

\begin{enumerate}[leftmargin=*, topsep=0pt, partopsep=0pt, parsep=0pt]
    \item \textbf{Interaction and Operational Style}: Measures the nature of engagement between participants at the turn level using metrics like proactivity, emphasis, and question type.

    \item \textbf{Conversational Properties}: Evaluates the naturalness of the conversation at the turn level through metrics such as repetition, disfluency, and the presence/type of ASR noise. We account for the fact that multiple disfluency and ASR noise types can be present in a turn.

    \item \textbf{Emotional and Sentiment}: Captures the turn-level sentiment of the conversation and transcript-level progression (arc) of sentiment and emotion of the agent and customer separately. Unlike approaches that analyse the sentiment at a turn level \citep{fu2022entitylevelsentimentanalysiscontact}, our metrics enable analysis of sentiment shifts over time ("arcs"), which more directly reflects conversational dynamics.

    \item \textbf{Linguistic Complexity and Content Density}: Assesses the richness and accessibility of language. Transcript-level metrics include technical density, sentence complexity, discourse flow, and overall readability. Language complexity is evaluated for each turn. While traditional linguistic metrics such as Flesch–Kincaid \citep{flesch1948new} readability scores have been used to quantify text complexity in other domains \citep{rooein2024conversationssourceteachingscientific}, they are generally coarse and statistical; our framework complements such measures with richer, task-informed features like technical density and discourse flow to capture nuanced structural and stylistic properties of dialogues.
\end{enumerate}

Full definitions of all metrics and it's categories along with examples are provided in Table~\ref{tab:comprehensive_eval_metrics_original} and Section~\ref{eval_metric_detailed_desc}. The analysis pipeline proceeds through the following stages:

\begin{enumerate}[leftmargin=*, topsep=-1pt, noitemsep]
    % \item \textbf{Programmatic Transcript Chunking:} To create an efficient tradeoff between analytical accuracy and computational cost, each transcript is segmented into smaller, contiguous chunks. This method avoids the context-length limitations of processing entire transcripts at once, as well as the high cost of analyzing each turn individually. The chunking algorithm dynamically determines the number of segments to ensure that all chunks are of approximately equal size

    \item \textbf{Turn-Level Feature Annotation:} A judge LLM classifies the turns within each transcript segment for a multi-dimensional taxonomy of conversational characteristics ($\mathcal{C}$) at turn level. For \textit{Solution} and \textit{Proactivity} which are agent-specific turns, only agent turns are classified. To ensure classifications are contextually aware despite the segmentation, the model is provided with a context window of surrounding turns for each chunk.

    \item \textbf{Transcript-Level Feature Annotation:} Concurrently, a separate LLM-based model analyzes each full transcript to derive readability scores and descriptive emotion and sentiment arcs.

    \item \textbf{Empirical Frequency Distribution Construction:} Following annotation, we construct empirical frequency distributions for each characteristic dimension $C_d \in \mathcal{C}$. This is done for both the real data ($\mathbf{O}_d$) and synthetic data ($\mathbf{E}_d$) by aggregating classifications across all turns from the respective corpora ($\mathcal{T}_R$ and $\mathcal{T}_S$).
\end{enumerate}

\noindent \textbf{Statistical Comparison of Distributions}: Generated transcripts from each method, along with real transcripts, are analyzed at turn level and transcript level across these categories. We compare observed frequencies ($\mathbf{O}_d$) from real data and expected frequencies ($\mathbf{E}_d$) from synthetic data using Pearson's Chi-squared test, G-test (likelihood-ratio), and Jensen-Shannon divergence. With the frequency distributions, p-values from chi-square \citep{Pearson01071900} or G-test \citep{McDonald2014} (depending on the number of categories) and Jensen–Shannon (JS) Divergence \citep{MENENDEZ1997307} are calculated. Low p-values or high JS scores indicate significant differences, quantifying fidelity gaps.

% To validate our LLM-based annotation framework and the clarity of our metric definitions, we conducted a human evaluation study on the English dataset. We employed two in-house, English speaking annotators with backgrounds in linguistics who followed the same detailed guidelines provided as prompts to the language model. They underwent three iterative rounds of training and calibration on task-specific guidelines before commencing the main annotation phase. For both turn-level and transcript-level metrics, we created a stratified sample of approximately 50-60 annotations per metric, ensuring balanced representation from both the original and generated transcripts from all the strategies. 

% Alignment between the human and LLM annotations using Cohen's Kappa is calculated to be within the substantial range (0.6-0.8) for all the metrics (See Table \ref{tab:kappa_scores}). This alignment validates two critical aspects of our framework: (1) the metric categories are sufficiently granular and distinguishable, and (2) the annotation instructions are clear and unambiguous, enabling both human annotators and LLMs to consistently identify and classify conversational characteristics. 

 To validate our LLM-based annotation framework and the clarity of metric definitions, we conducted a human evaluation on the English dataset using two in-house English-speaking annotators with linguistics backgrounds who followed the same detailed guidelines used as LLM prompts. Annotators completed three iterative rounds of training and calibration prior to the main annotation phase. We constructed stratified samples of 972 annotations (50-60 per metric) at both turn and transcript levels, ensuring balanced coverage of original and generated transcripts across all strategies. \textbf{Agreement between human and LLM annotations, measured using Cohen’s Kappa, falls within the substantial agreement range (0.6–0.8)} for all metrics (Table~\ref{tab:kappa_scores}), validating that the metric categories are sufficiently granular and distinguishable and that the annotation guidelines are clear and unambiguous, enabling consistent identification of characteristics by both humans and LLMs.

\subsection{Evaluation on Downstream Task}

We evaluate three synthetic transcript generation pipelines ($m \in \{$\textit{Direct, Chunked, Characteristic-Aware}$\}$) on the downstream AutoQA task \citep{probing-qa, 10.3115/1225785.1225796}, which is formulated as a binary classification over $\mathcal{Y}={\texttt{Yes},\texttt{No}}$ given a transcript $x$ and a question $q$ (e.g., ``Did the agent verify the customer's name?''). To obtain model-agnostic evidence of synthetic transcript utility, experiments are conducted across four models: GPT-4.1-mini, GPT-4.1, GPT-4o, and Claude-v3-Haiku. We choose prompt optimization over fine-tuning for this study as (1) it applies directly to widely deployed commercial models that lack fine-tuning support, and (2) it achieves strong performance improvements at a fraction of the cost and time of traditional fine-tuning.

Specifically, we optimize $(x,q,y)$ triplets from real and synthetic transcripts, where the question–label pairs $(q,y)$ are shared but the transcript $x$ differs between real and synthetic sources, using \texttt{DSPy MIPROv2} \citep{opsahlong2024optimizinginstructionsdemonstrationsmultistage} with \textbf{Claude-3.5-Sonnet} as the prompt-generating model and macro-F1 as the loss function. With sufficiently large search budgets ($\texttt{num\_trials}=50$, $\texttt{num\_candidates}=5$; Table~\ref{tab:dspy_hyperparameters}), the optimizer exhaustively explores the instruction–demonstration space, reducing sensitivity to arbitrary sample selection. The detailed methodology is outlined in Section~\ref{downstream_eval_methodology} and results are discussed in Section~\ref{sec:downstream_results}.

% We construct training datasets comprising $(x, q, y)$ triplets for both real transcripts ($\mathcal{D}_{\text{train}}^{\text{real}}$) and synthetic transcripts ($\{\mathcal{D}_{\text{train}}^{m}\}_{m}$), where the question-label pairs $(q, y)$ are identical across sources but the transcript $x$ differs. These triplets serve as labeled demonstrations for prompt optimization using the \texttt{DSPy MIPROv2} optimizer \citep{opsahlong2024optimizinginstructionsdemonstrationsmultistage}. We optimize each of the 4 models independently with \textbf{Claude-3.5-Sonnet} \citep{claude-3_5-sonnet} as the prompt-generating model and macro-F1 as the loss function. With sufficiently large hyperparameters ($\texttt{num\_trials}=50$, $\texttt{num\_candidates}=5$; Table~\ref{tab:dspy_hyperparameters}), the optimizer exhaustively explores the joint space of task instructions and few-shot configurations, reducing sensitivity to arbitrary sample selection. Because the transcript modality $x$ differs between real and synthetic sources, the optimizer produces distinct prompts and few-shot examples that reflect the characteristics of the training transcripts.

%%%%% We also consider an un-optimized baseline where the prompt before optimizing is considered for evaluation.

% We evaluate all methods on a shared held-out test set $\mathcal{D}_{\text{test}}^{\text{real}}$, reporting the macro F1 scores averaged across all model variants. The detailed methodology is outlined in Section~\ref{downstream_eval_methodology} and results are discussed in Section~\ref{sec:downstream_results}.

\section{Experimental Setup}
\subsection{Dataset}  
\label{dataset}

The \textbf{in-house} multilingual synthetic transcript dataset\footnote{Due to proprietary restrictions, this dataset cannot be released.} is constructed by sampling real call center conversations across multiple domains (e.g., retail, logistics, telecom), four languages (English, Spanish, French, French-Canadian), and call length categories to ensure broad linguistic and domain coverage. Only permissible data approved for experimental use are included, with all sensitive personally identifiable and payment card information redacted; details on structured call attributes are provided in Section~\ref{desc_struc_inputs}. For transcript generation evaluation, the dataset comprises 200 test examples (50 per language) and 400 tuning examples (100 per language), with a 70--30 split used for prompt tuning based on reconstruction score. For downstream AutoQA evaluation, we construct a separate, disjoint dataset by flattening each call’s QA form into $(x,q,y)$ triplets after cross-lingual normalization. The resulting dataset contains 612 training examples from 168 calls, 263 validation examples from 121 calls, and 4,184 test examples from 756 calls. Following prompt optimization guidelines \citep{opsahlong2024optimizinginstructionsdemonstrationsmultistage}, we adopt a split favoring smaller training sets and larger validation and test sets, with a shared validation set across all optimization methods to ensure fair model selection. Full dataset construction and filtering details are provided in Section~\ref{downstream_eval_methodology}.

\subsection{Models}

% All synthetic transcript generation pipelines use \textbf{GPT-4.1-mini} \citep{gpt4_1_mini}, while \textbf{Claude-3.5-Sonnet} \citep{claude-3_5-sonnet} is used for computing evaluation metrics and reconstruction loss. Separate models are used for generation and evaluation to reduce model-specific biases and ensure a more reliable assessment of output quality. For the downstream AutoQA task, \textbf{GPT-4.1-mini} is used for answer generation. For all prompt optimization procedures (both generation pipeline tuning and downstream task optimization), we use DSPy MIPROv2 \citep{opsahlong2024optimizinginstructionsdemonstrationsmultistage} with \textbf{GPT-4.1-mini} as the task model (the model being optimized, matching the generation model) and \textbf{Claude-3.5-Sonnet} as the prompt model (the meta-optimizer that proposes candidate prompts). This ensures that the optimized prompts are tailored to the same model used for generation. Ablations are done using \textbf{GPT-4.1} \citep{openai2025gpt41}, \textbf{GPT-4o} \citep{openai2024gpt4o}, \textbf{Claude-v3-Haiku} \citep{rahman2024claudehaiku}. LLM configuration details are provided in Table~\ref{tab:llm_pipeline_parameters}.

All synthetic transcript generation pipelines use \textbf{GPT-4.1-mini} \citep{gpt4_1_mini}, while \textbf{Claude-3.5-Sonnet} \citep{claude-3_5-sonnet} is used for metric computation and reconstruction loss, with separate models employed for generation and evaluation to reduce model-specific bias. All prompt optimization procedures covering both generation pipeline tuning and downstream task optimization are performed using DSPy MIPROv2 \citep{opsahlong2024optimizinginstructionsdemonstrationsmultistage}, with \textbf{GPT-4.1-mini} as the task model and \textbf{Claude-3.5-Sonnet} as the meta-optimizer prompt model. This ensures that the optimized prompts are tailored to the same model used for generation. Model sensitivity study additionally use \textbf{GPT-4.1} \citep{openai2025gpt41}, \textbf{GPT-4o} \citep{openai2024gpt4o}, and \textbf{Claude-v3-Haiku} \citep{rahman2024claudehaiku}. Downstream task uses all 4 models with full configuration details provided in Table~\ref{tab:llm_pipeline_parameters}.

\subsection{Baselines}

We adapt two recent methods for synthetic dialogue generation to the call center domain: \textbf{NoteChat} \citep{Wang_2024}, originally designed for clinical conversations, and \textbf{ConvoGen} \citep{gody2025convogenenhancingconversationalai}, a multi-agentic approach for synthetic conversation generation. For fair comparison, same set of call attributes and system level prompts used in the our approaches are provided to the baselines ensuring structured supervision. Rationale for choosing the two baselines and their full adaptation details are provided in Section~\ref{baselines}.

\subsection{Prompt Optimization For Generation}
\label{prompt_tuning_for_transcript_gen}

% We define a composite \textbf{Reconstruction Score} to measure how well a synthetic transcript reflects its intended structure, content, and conversational style. It combines multiple LLM evaluated metrics: topic flow, intent fulfillment, QA consistency, and speech realism into a single score via weighted aggregation after normalization. \textbf{This score helps measure the fidelity to the input call attributes and instructions and thus is used as the objective function for prompt optimization.} Full metric definitions and computation details are provided in Sections~\ref{loss_function_prompt_tuning} and ~\ref{reconstruction_score_calculation}.

To enable effective prompt optimization, we define a composite \textbf{Reconstruction Score} that aggregates LLM-evaluated metrics including topic flow, intent fulfillment, QA consistency, and speech realism to quantify how well a synthetic transcript adheres to the intended structure, content, and style. This score measures fidelity to input call attributes and instructions and serves as the objective function for prompt optimization. Full metric definitions and details are provided in Sections~\ref{loss_function_prompt_tuning} and ~\ref{reconstruction_score_calculation}.

\subsection{Implementation Overview}

Our generation and evaluation pipelines (for transcript generation and downstream AutoQA task) are implemented in Python.
% and run entirely on local infrastructure for reproducibility. 
Prompt optimization is performed using \texttt{DSPy} \cite{khattab2023dspycompilingdeclarativelanguage} with the \texttt{MIPROv2} optimizer \cite{opsahlong2024optimizinginstructionsdemonstrationsmultistage}. Model access is managed via \texttt{Bedrock} and OpenAI API.  All LLM configurations and hyperparameters used for prompt tuning are detailed in Section~\ref{implementation_details}.

\begin{table*}[h!]
\centering
  \fontsize{8}{9}\selectfont
\begin{tabular}{lcccccccc}
\toprule
\multirow{2}{*}{Method} &
  \multicolumn{2}{c}{Proactivity} &
  \multicolumn{2}{c}{Emphasis} &
  \multicolumn{2}{c}{Question Type} &
  \multicolumn{2}{c}{Solution} \\
\cmidrule(lr){2-3} \cmidrule(lr){4-5} \cmidrule(lr){6-7} \cmidrule(lr){8-9}
&
\multicolumn{1}{c}{$\chi^2$/G (p)} &
\multicolumn{1}{c}{JS-Div} &
\multicolumn{1}{c}{$\chi^2$/G (p)} &
\multicolumn{1}{c}{JS-Div} &
\multicolumn{1}{c}{$\chi^2$/G (p)} &
\multicolumn{1}{c}{JS-Div} &
\multicolumn{1}{c}{$\chi^2$/G (p)} &
\multicolumn{1}{c}{JS-Div} \\
\midrule
  Direct  & 0.000 & 0.041 & 0.000 & 0.023 & 0.000 & 0.008 & 0.000 & 0.020  \\
  Chunked  & 0.000 & 0.008 & 0.000 & 0.023 & 0.000 & 0.003 & 0.000 & 0.021  \\
  Characteristic Aware  & 0.000 & 0.026 & 0.000 & 0.026 & 0.000 & 0.012 & 0.000 & 0.024  \\
  ConvoGen  & 0.000 & 0.019 & 0.000 & 0.068 & 0.000 & 0.006 & 0.000 & 0.044  \\
  NoteChat  & 0.000 & 0.008 & 0.000 & 0.008 & 0.000 & 0.022 & 0.000 & 0.021  \\
 \bottomrule
\end{tabular}%
\vspace*{-2.5mm} % Add negative vertical space here
\caption{Comparison of methods for transcript generation across \textbf{English} language and metrics in the \textbf{Interaction Style and Operational} category.}
\label{results_interaction_style_english}
\end{table*}

% \begin{table*}[h!]
%   \centering
%   \fontsize{6}{8}
% \resizebox{0.8\textwidth}{!}{%
%   \tiny
% \begin{tabular}{lcccccc}
% \toprule
% \multirow{2}{*}{Method} &
%   \multicolumn{2}{c}{Disfluency} &
%   \multicolumn{2}{c}{Repetition} &
%   \multicolumn{2}{c}{ASR Noise} \\
% \cmidrule(lr){2-3} \cmidrule(lr){4-5} \cmidrule(lr){6-7}
% &
% \multicolumn{1}{c}{$\chi^2$/G (p)} &
% \multicolumn{1}{c}{JS-Div} &
% \multicolumn{1}{c}{$\chi^2$/G (p)} &
% \multicolumn{1}{c}{JS-Div} &
% \multicolumn{1}{c}{$\chi^2$/G (p)} &
% \multicolumn{1}{c}{JS-Div} \\
% \midrule
%   Direct  & 0.000 & 0.126 & 0.000 & 0.003 & 0.000 & 0.106  \\
%   Chunked  & 0.000 & 0.086 & 0.001 & 0.001 & 0.000 & 0.065  \\
%   Characteristic Aware  & 0.000 & 0.075 & 0.000 & 0.002 & 0.000 & 0.065  \\
%   ConvoGen  & 0.000 & 0.321 & \textcolor{magenta}{\textbf{0.234}} & 0.000 & 0.000 & 0.128  \\
%   NoteChat  & 0.000 & 0.069 & 0.000 & 0.036 & 0.000 & 0.066  \\
%  \bottomrule
% \end{tabular}%
% }
% \caption{Comparison of methods for transcript generation across \textbf{English} language and metrics in the \textbf{Conversational Properties} category.}
% \label{results_conversational_properties_english}
% \end{table*}

\begin{table*}[t]
  \centering
  \fontsize{8}{9}\selectfont
\begin{tabular}{lcccccc}
\toprule
\multirow{2}{*}{Method} &
  \multicolumn{2}{c}{Disfluency} &
  \multicolumn{2}{c}{Repetition} &
  \multicolumn{2}{c}{ASR Noise} \\
\cmidrule(lr){2-3} \cmidrule(lr){4-5} \cmidrule(lr){6-7}
&
{$\chi^2$/G (p)} & JS-Div &
{$\chi^2$/G (p)} & JS-Div &
{$\chi^2$/G (p)} & JS-Div \\
\midrule
Direct  & 0.000 & 0.126 & 0.000 & 0.003 & 0.000 & 0.106 \\
Chunked & 0.000 & 0.086 & 0.001 & 0.001 & 0.000 & 0.065 \\
Characteristic Aware & 0.000 & 0.075 & 0.000 & 0.002 & 0.000 & 0.065 \\
ConvoGen & 0.000 & 0.321 & \textcolor{magenta}{\textbf{0.234}} & 0.000 & 0.000 & 0.128 \\
NoteChat & 0.000 & 0.069 & 0.000 & 0.036 & 0.000 & 0.066 \\
\bottomrule
\end{tabular}
\vspace*{-2.5mm} % Add negative vertical space here
\caption{Comparison of methods for transcript generation across \textbf{English} language and metrics in the \textbf{Conversational Properties} category.}
\label{results_conversational_properties_english}
\end{table*}

\begin{table*}[ht]
  \centering
\resizebox{0.9\textwidth}{!}{%
\large
\begin{tabular}{lcccccccccc}
\toprule
\multirow{2}{*}{Method} &
  \multicolumn{2}{c}{Sentiment} &
  \multicolumn{2}{c}{Customer Emotion Arc$^{\dagger}$} &
  \multicolumn{2}{c}{Agent Emotion Arc$^{\dagger}$} &
  \multicolumn{2}{c}{Customer Sentiment Arc$^{\dagger}$} &
  \multicolumn{2}{c}{Agent Sentiment Arc$^{\dagger}$} \\
\cmidrule(lr){2-3} \cmidrule(lr){4-5} \cmidrule(lr){6-7}
\cmidrule(lr){8-9} \cmidrule(lr){10-11}
&
\multicolumn{1}{c}{$\chi^2$/G (p)} &
\multicolumn{1}{c}{JS-Div} &
\multicolumn{1}{c}{$\chi^2$/G (p)} &
\multicolumn{1}{c}{JS-Div} &
\multicolumn{1}{c}{$\chi^2$/G (p)} &
\multicolumn{1}{c}{JS-Div} &
\multicolumn{1}{c}{$\chi^2$/G (p)} &
\multicolumn{1}{c}{JS-Div} &
\multicolumn{1}{c}{$\chi^2$/G (p)} &
\multicolumn{1}{c}{JS-Div} \\
\midrule
  Direct  & 0.000 & 0.024 & \textcolor{magenta}{\textbf{0.255}} & 0.206 & 0.023 & 0.159 & \textcolor{magenta}{\textbf{0.118}} & 0.059 & 0.005 & 0.138  \\
  Chunked  & 0.000 & 0.021 & \textcolor{magenta}{\textbf{0.381}} & 0.162 & \textcolor{magenta}{\textbf{0.065}} & 0.169 & \textcolor{magenta}{\textbf{0.087}} & 0.073 & 0.032 & 0.108  \\
  Characteristic Aware  & 0.000 & 0.020 & 0.038 & 0.275 & 0.007 & 0.182 & 0.008 & 0.110 & 0.011 & 0.118  \\
  ConvoGen  & 0.000 & 0.028 & 0.004 & 0.330 & 0.027 & 0.170 & 0.011 & 0.085 & 0.013 & 0.123  \\
  NoteChat  & 0.000 & 0.030 & 0.000 & 0.390 & 0.002 & 0.210 & 0.000 & 0.221 & 0.001 & 0.162  \\
 \bottomrule
\end{tabular}%
}
\vspace*{-2.5mm} % Add negative vertical space here
\caption{Comparison of methods for transcript generation across \textbf{English} language and metrics in the \textbf{Sentiment and Emotion} category. $^{\dagger}$ denotes transcript-level metrics; unmarked metrics are turn-level traits.}
\label{results_sent_and_emotion_english}
\end{table*}

\begin{table*}[ht]
  \centering
\resizebox{0.9\textwidth}{!}{%
\large
\begin{tabular}{lcccccccccc}
\toprule
\multirow{2}{*}{Method} &
  \multicolumn{2}{c}{Language Complexity} &
  \multicolumn{2}{c}{Technical Density$^{\dagger}$} &
  \multicolumn{2}{c}{Sentence Complexity$^{\dagger}$} &
  \multicolumn{2}{c}{Overall Readability$^{\dagger}$} &
  \multicolumn{2}{c}{Discourse Flow$^{\dagger}$} \\
\cmidrule(lr){2-3} \cmidrule(lr){4-5} \cmidrule(lr){6-7}
\cmidrule(lr){8-9} \cmidrule(lr){10-11}
&
\multicolumn{1}{c}{$\chi^2$/G (p)} &
\multicolumn{1}{c}{JS-Div} &
\multicolumn{1}{c}{$\chi^2$/G (p)} &
\multicolumn{1}{c}{JS-Div} &
\multicolumn{1}{c}{$\chi^2$/G (p)} &
\multicolumn{1}{c}{JS-Div} &
\multicolumn{1}{c}{$\chi^2$/G (p)} &
\multicolumn{1}{c}{JS-Div} &
\multicolumn{1}{c}{$\chi^2$/G (p)} &
\multicolumn{1}{c}{JS-Div} \\
\midrule
  Direct  & 0.000 & 0.057 & \textcolor{magenta}{\textbf{0.136}} & 0.034 & 0.001 & 0.107 & 0.001 & 0.110 & 0.000 & 0.526  \\
  Chunked  & 0.000 & 0.018 & 0.041 & 0.042 & 0.019 & 0.067 & 0.023 & 0.055 & 0.000 & 0.116  \\
  Characteristic Aware  & 0.000 & 0.047 & 0.014 & 0.058 & \textcolor{magenta}{\textbf{0.428}} & 0.014 & \textcolor{magenta}{\textbf{0.135}} & 0.031 & 0.000 & 0.270  \\
  ConvoGen  & 0.000 & 0.009 & \textcolor{magenta}{\textbf{0.275}} & 0.025 & 0.019 & 0.067 & 0.006 & 0.084 & 0.014 & 0.072  \\
  NoteChat  & 0.000 & 0.050 & \textcolor{magenta}{\textbf{0.090}} & 0.038 & 0.033 & 0.053 & 0.005 & 0.086 & 0.000 & 0.201  \\
 \bottomrule
\end{tabular}%
}
\vspace*{-2.5mm} % Add negative vertical space here
\caption{Comparison of methods for transcript generation across \textbf{English} language and metrics in the \textbf{Linguistic Complexity and Content Density} category. $^{\dagger}$ denotes transcript-level metrics; unmarked metrics are turn-level.}
\label{results_linguistic_complexity_english}
\end{table*}

\section{Results}

\begin{figure}[t]
    \centering
    \includegraphics[width=0.5\textwidth]{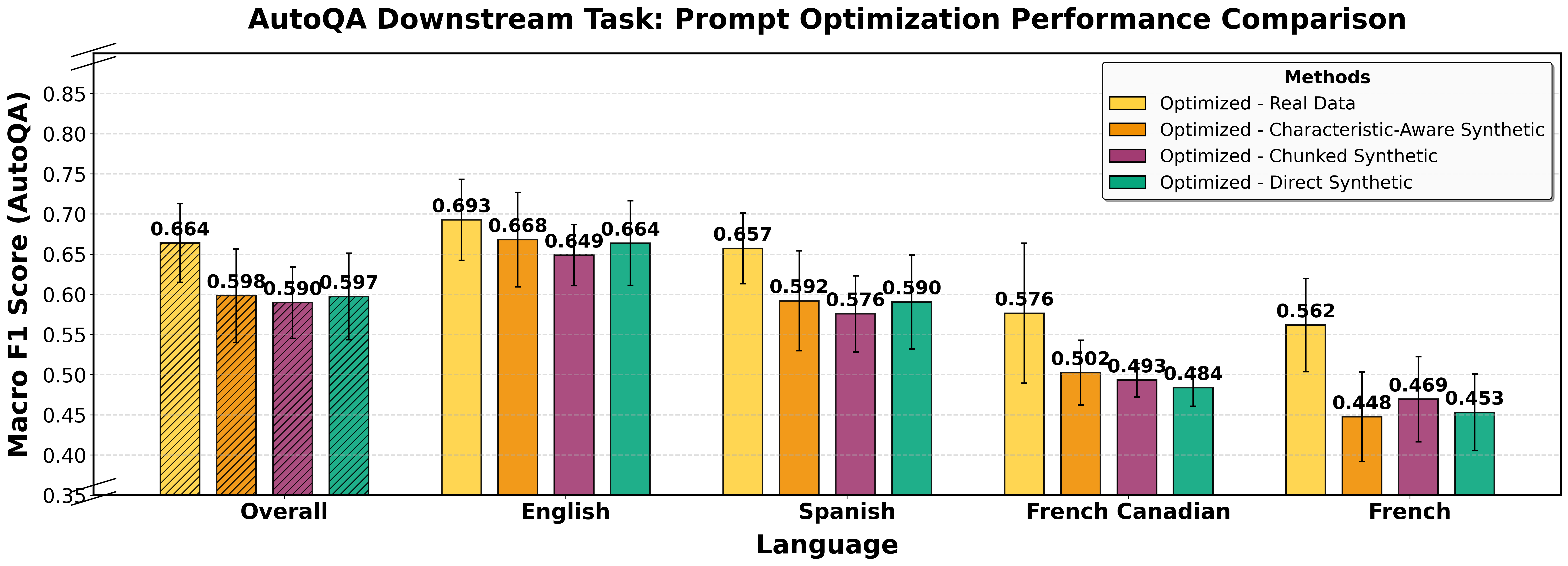}
    \vspace*{-7mm} % Add negative vertical space here
    \caption{Comparison of AutoQA performance using real and synthetic transcripts for prompt optimization.}
    \label{fig:downstream_language_f1}
\end{figure}

We first present results from the downstream AutoQA task evaluation (\S\ref{sec:downstream_results}), which reveals a consistent performance gap between prompts optimized on real versus synthetic transcripts. This gap motivates a detailed analysis of synthetic transcript quality across four categories of evaluation metrics: (a) Interaction Style,  (b) Conversational Properties, (c) Sentiment/Emotion, and (d) Linguistic Complexity. We evaluate five generation methods (Direct, Chunked, Characteristic-Aware, ConvoGen, and NoteChat) across four languages (English, Spanish, French, French-Canadian). 
% The null hypothesis ($H_0$) for both chi-square and G-tests of independence is that the distributions are the same (equivalently, the categories are independent / follow the expected distribution). Values in \textcolor{magenta}{\textbf{{magenta bold}}} in the table indicate $p\text{-value} > 0.05$, denoting statistical similarity or independence as we fail to reject the null hypothesis. In other words, it denotes statistical similarity between generated and real transcripts for the specified metric, suggesting the synthetic transcript successfully captures that aspect of real conversations.
For both the chi-square and G-tests, $p<0.05$ indicates a statistically significant divergence between the distributions of real and synthetic transcripts for the corresponding metric. Conversely, values highlighted in \textcolor{magenta}{\textbf{magenta bold}} correspond to $p>0.05$, representing the desirable outcome in which real and synthetic transcripts exhibit statistically similar distributions.

\subsection{Downstream Task Performance}
\label{sec:downstream_results}

Figure~\ref{fig:downstream_language_f1} shows overall and language-wise Macro F1 scores for the AutoQA task, comparing prompts optimized on real versus synthetic transcripts. Bar heights denote average Macro F1 across four model variants (GPT-4.1-mini, GPT-4.1, Claude Haiku, GPT-4o), with error bars indicating inter-model standard deviation. \textbf{Prompts optimized using real transcripts (\textit{Optimized--Real Data}) consistently outperform all synthetic methods} across languages, achieving an overall Macro F1 of 0.664 compared to 0.598 for Characteristic-Aware Synthetic, the best-performing synthetic approach, corresponding to an 11.0\% advantage. Notably, the \textbf{performance gap is substantially larger for non-English languages} than for English, indicating that synthetic transcripts struggle to capture language-specific conversational patterns and cultural nuances. These consistent deficits across synthetic methods motivate a deeper analysis of transcript quality, which we undertake in the following subsections using our evaluation framework.

\subsection{Interaction Style and Operational Traits}

Table~\ref{results_interaction_style_english} and \ref{results_interaction_style_all_languages} presents results for interaction-level traits: \textit{Emphasis}, \textit{Question Type}, \textit{Solution}, and \textit{Proactivity}. \textbf{Most methods struggle with key interaction traits.} No method consistently aligns with real data on \textit{Proactivity}, \textit{Emphasis}, \textit{Question Type} and \textit{Solution}  across all languages. A deeper analysis of the frequency distribution (Table \ref{freq_dist_proactivity}-\ref{freq_dist_solution}) of these metrics categories reveal the following: \textbf{(i) Proactivity: agent turns that are \texttt{Understated Proactivity} are much less (4-15\%) in synthetic transcripts compared to real ones (20-30\%).} This shows the reluctance of the generation models and methods in being able to produce transcripts where the agents are not proactive or helpful. \textbf{(ii) Solution: In reality, agents tend to listen more, than proactively offer a solution.} Most of the agent turns in real transcript are not oriented towards providing a solution (~40\% \texttt{No Solution}) whereas the generated transcripts have the agent providing solution (~30\% \texttt{Solution Oriented}) or explaining the process (~45\% \texttt{Process Oriented}) in most turns.

\subsection{Conversational Properties}
Table~\ref{results_conversational_properties_english} and \ref{results_conversational_properties_all_languages} reports results on: \textit{Repetition}, \textit{Disfluency}, and \textit{ASR Noise Type}.
\textbf{(a) Conversational properties remain one of the hardest to model.} Upon analyzing the Disfluency frequency distributions (Table \ref{freq_dist_disfluency}), it's evident that none of the generation strategies are able to replicate the \texttt{interactional\_disfluency} to the level that it's present in real transcripts (15-30\% in real conversations vs 1-6\% in synthetic). A similar analysis on ASR Noise distributions (Table \ref{freq_dist_asr_noise_type}) reveal that while \texttt{insertion} errors are modeled appropriately, the presence of \texttt{deletion} and \texttt{substitution} errors are very sparse in generated transcripts.
\textbf{(b) Convogen \citep{gody2025convogenenhancingconversationalai}, a multi-agentic approach is the only approach that's able to simulate repetition realistically (Table \ref{results_conversational_properties_all_languages})}. A possible reason could be that the nature of multi-agentic approach where individual turns are generated one-at-a-time is conducive for modeling customer and agent repetition.

\subsection{Sentiment and Emotion Fidelity}
From Table \ref{results_sent_and_emotion_english} and \ref{results_sent_and_emotion_all_languages}, we observe following patterns:

\textbf{(a) Modeling turn-level sentiment remains challenging compared to sentiment/emotion arcs.} No method achieves statistical similarity on the \textit{Sentiment} (turn level) metric, highlighting the difficulty of capturing local emotional nuance despite better global alignment. As indicated by Table \ref{freq_dist_sentiment}, all generation methods produce turns that have disproportionately more \texttt{Positive Sentiment} (15-25\%) compared to real transcript distribution where it is only in the range of 5-6\%. 

\textbf{(b) Customer sentiment/emotion arc is easier to model compared to that of the agent's.} Both the Direct and Chunked pipeline are able to capture customer emotion and sentiment in most of the languages as outlined in Table \ref{results_sent_and_emotion_all_languages}. Agent's emotion arc is misrepresented in the generated transcripts as \texttt{factual\_to\_gratitude} (60-70\% in generated vs 30-50\% in real transcripts) in most cases whereas real transcripts have more varied emotion arcs for agents (Table \ref{freq_dist_agent_emotion}). 

\textbf{(c) Baselines lack alignment and fidelity, failing to capture both the agent and customer arcs.} ConvoGen \citep{gody2025convogenenhancingconversationalai} and NoteChat \citep{Wang_2024} though adapted with structured supervision, underperform - likely due to the lack of constraint adherence mechanisms or optimization via evaluation-guided prompting used in our methods.

\subsection{Linguistic Complexity \& Content Density}

Table~\ref{results_linguistic_complexity_english} and \ref{results_linguistic_complexity_all_languages} summarizes the linguistic fidelity of synthetic transcripts across dimensions such as 
\textit{Language Complexity}, \textit{Technical Density}, \textit{Sentence Complexity}, \textit{Overall Readability} and \textit{Discourse Flow}. Key findings:

% \textbf{(a) Language complexity and discourse flow is the hardest to replicate.} No method matches the distribution in any language (Table \ref{results_linguistic_complexity_all_languages}), reflecting the difficulty of modeling stylistic nuance and discourse structure with structured guidance. Analysis of the frequency distribution for language complexity (Table \ref{freq_dist_language_complexity}) reveals that although both real and synthetic conversation turns are in simple language, real conversation turns tend to be more informal (\texttt{Simple Informal Language}). There is a consistent over-representation of formal language in all the generation methods with distribution ranging 40-50\% compared to 15-20\% in real conversations highlighting the innate bias of the model and generation strategies in generating overly formal conversations. Discourse flow (Table \ref{freq_dist_discourse_flow}) also tends to be more linear in generated transcripts (nearly 90\% in \texttt{4-5} range which indicate smooth transitions as opposed to 50-60\% in real conversations). This is especially apparent in Direct generation strategy with discourse flow becoming progressively complex in Dual, Characteristic Aware and baseline strategies. This highlights the efficacy of chunk-wise enhancements in modeling in-coherent conversations with poor logical progression reflective of the real nature of call center transcripts.

\textbf{(a) Language complexity and discourse flow are the hardest to replicate.}
No method matches the real distribution across languages (Table~\ref{results_linguistic_complexity_all_languages}), underscoring the difficulty of modeling stylistic nuance and discourse structure with structured guidance. Frequency analysis of language complexity (Table~\ref{freq_dist_language_complexity}) shows that while both real and synthetic turns are predominantly simple, real conversations are more informal (\texttt{Simple Informal Language}), whereas all generation methods systematically overproduce formal language (40--50\% vs.\ 15--20\% in real data), revealing an inherent bias toward overly formal dialogue. Discourse flow (Table~\ref{freq_dist_discourse_flow}) is substantially more linear in generated transcripts, with nearly 90\% exhibiting excellent coherence (\texttt{4--5} range) compared to only 50--60\% in real conversations; this effect is strongest for Direct generation and is progressively attenuated in Dual, Characteristic-Aware, and baseline strategies.  These trends highlight the effectiveness of chunk-wise enhancements in comparatively better capturing the incoherent and irregular discourse patterns characteristic of real call center transcripts.

\textbf{(b) Overall readability is harder for real transcripts.}
As shown in Table~\ref{freq_dist_overall_readability_score}, synthetic transcripts exhibit substantially higher easy-to-read scores (90--100\% in \texttt{4--5} score) than real transcripts (60--80\%). This pattern aligns with the discourse flow findings in (a), as simpler, more linear trajectories are easier to read. Qualitative inspection further reveals that real transcripts contain frequent interruptions and topic shifts, which are largely absent in synthetic.

% \textbf{(b) The Overall Readability of real is harder compared to synthetic transcripts.} This is indicated by the higher \texttt{4-5} (easy to read) values in Table~\ref{freq_dist_overall_readability_score} for synthetic (90-100\%) compared to real transcripts (60-80\%). This result also follows from the observations regarding Discourse Flow outlined in (a): simple discourse trajectories are easier to read. Upon examining the real and synthetic transcripts, it is apparent that real transcripts are marked with frequent interruptions and topic jumps which are much less in synthetic counterparts. 

\textbf{(c) Across all the languages, most of the methods are able to generate conversations with similar levels of technical density as that of real conversations.} This also follows from the investigation of language complexity in (a), which indicate that the turns in synthetic and real transcripts tend to be simple in nature, lacking technical jargon.

\subsection{Reconstruction Score}
Table \ref{reconstruction_score_table} reports \textit{Reconstruction Score}, a weighted composite metric (see Section~\ref{reconstruction_score_calculation}) measuring how well generated transcripts reflect the structured inputs: intent summaries, topic flows, QA evaluations, disfluencies, and ASR noise. The key observation from the analysis is that \textbf{(high reconstruction fidelity does not guarantee conversational realism.} As detailed in Table~\ref{reconstruction_score_table}, the direct generation strategy consistently achieves the highest reconstruction scores across most languages. However, this strict adherence is not indicative of realistic conversations compared to other methods which score better in the evaluation metrics (see Table~\ref{results_sent_and_emotion_all_languages}-~\ref{results_conversational_properties_all_languages}). Even a smaller model like GPT-4.1-mini can achieve high reconstruction scores. This indicates that capturing predefined call attributes is a relatively simple task. However, as noted, this high fidelity does not inherently translate into generating realistic, human-like conversations. \textbf{This highlights a fundamental insight: conversation realism does not originate from the call attributes but is rather a property of the LLM and generation strategy.} Thus, explicit supervision on the input call attribute is not sufficient and necessitates architectural and model changes. This result is further strengthened by the following sections on model sensitivity and call attribute ablation. 

\section{Analysis of Model Sensitivity}
\label{sec:ablation_model}

We conduct a comparative analysis across four models (Table~\ref{tab:model_ablation_pvalue}) to assess the impact of model capacity on synthetic transcript realism, benchmarking GPT-4.1-mini against GPT-4.1, GPT-4o, and Claude Haiku v3. \textbf{(a) More capable models consistently achieve statistical independence on Linguistic Complexity and Content Density metrics}, with all three showing improvements in Technical Density, Sentence Complexity, and Overall Readability, indicating a stronger implicit understanding of these dimensions without explicit supervision. \textbf{(b) In contrast, Interaction Style, Operational metrics, and Conversational Properties remain dependent across all models}, as Proactivity, Solution, Question Type, and Emphasis retain near-zero $p$-values regardless of model capacity, reflecting their reliance on pragmatic, conversation-level context not encoded in call attributes; similarly, ASR Noise and Disfluency persist as surface-level phenomena largely orthogonal to language modeling capability. Overall, GPT-4o achieves the strongest results with six metric improvements and no degradations, followed by GPT-4.1 (five) and Claude Haiku v3 (four). \textbf{While larger models perform better, high-quality generation with smaller models such as GPT-4.1-mini remains essential, as large-scale synthetic data generation with more capable models is prohibitively expensive.}

\section{Call Attribute Ablation}
\label{sec:ablation_call_attribute}
% Our analysis (Table \ref{tab:multi_pipeline_ablation_pvalue}) reveals how the call input attributes serve as a foundational but limited mechanism for enhancing conversational realism, with distinct effects on different metric categories.
% \textbf{(a) The necessity of a complete attribute set is most evident in simpler strategies like the direct generation pipeline, which lacks the capacity to model complex dynamics without explicit guidance.} The removal of all attributes causes Direct generation pipeline to completely collapse in realism with 3 independent distributions to no independence in any metric. 
% \textbf{(b) High-level agent-centric metrics like Proactivity and Solution exhibit notable stability across all ablations.} These metrics remained consistently dependent on the reference distribution regardless of which attributes were removed. This suggests that these features are emergent properties of the conversational strategy learned by the language model, rather than behaviors that can be explicitly supervised or directly controlled by the presence or absence of individual input signals like Summary or Topic Flow.

Our analysis (Table~\ref{tab:multi_pipeline_ablation_pvalue}) illustrates that call input attributes provide a necessary but limited foundation for improving conversational realism, with heterogeneous effects across metric categories. \textbf{(a) The importance of a complete attribute set is most pronounced for simpler strategies such as Direct generation}, which lack the capacity to model complex dynamics without explicit guidance; removing all attributes causes the Direct pipeline to collapse from three independent distributions to no independence across any metric. \textbf{(b) In contrast, high-level agent-centric metrics such as Proactivity and Solution remain stable across all ablations}, consistently exhibiting dependence on the reference distribution regardless of which attributes are removed, suggesting that these behaviors emerge from the model’s learned conversational strategy rather than being directly controlled by individual input signals such as Summary or Topic Flow.

\section{Conclusion}

We introduced a diagnostic evaluation framework to quantify the realism of synthetic transcripts in contact center settings. By benchmarking multiple generation strategies, we showed that even with structured supervision, synthetic data fails to match the utility of real data on a downstream AutoQA task. Our 17-metric analysis highlights that these deficits are most prominent in areas such as disfluency modeling and sentiment fidelity. By surfacing these specific deficiencies, our diagnostic tool provides a principled foundation for improving synthetic generation methods and reaching downstream-ready transcript quality at scale.

\section{Limitations}

While our structured pipeline surfaces several useful insights into multilingual synthetic transcript generation, it is not without limitations. Firstly, although we evaluate multiple prompting strategies in isolation, we do not investigate hybrid approaches that selectively combine their strengths - e.g., using Direct prompting for semantic fidelity and characteristic-aware generation for behavioral induction. Secondly, our modular pipeline currently applies rule-based supervision and sequential prompting but does not leverage reinforcement learning or differentiable objectives to induce traits like disfluency, noise, or emotion more robustly. Future work could explore policy-gradient methods to iteratively refine these characteristics. Thirdly, we limit our analysis to four languages due to resource constraints. Extending the pipeline to lower-resource or morphologically rich languages would test its generalizability. Lastly, our structured supervision—though diverse—may not cover all latent cues necessary for high-fidelity generation, especially for nuanced metrics like emphasis and ASR noise. Additional metadata, acoustic signals, or fine-grained annotation might be needed to bridge this gap.

\bibliography{custom}

\clearpage
\appendix
\onecolumn
\section{Appendix}
\label{sec:appendix}

\begin{table*}[h!]
\centering
\resizebox{\textwidth}{!}{%
\begin{tabular}{llccccccc}
\toprule
Language & Method & \shortstack{Intent Summary \\ Score} & \shortstack{Topic Flow \\ Score} & \shortstack{QA Eval \\ Score} & \shortstack{Disfluency \\ Score} & \shortstack{ASR Noise \\ Score} & \shortstack{Interruption \\ Score} & \shortstack{Overall \\ Score} \\ \midrule
\multirow{5}{*}{\shortstack{\texttt{English}}} & Direct & \textcolor{magenta}{\textbf{0.97}} & \textcolor{magenta}{\textbf{0.86}} & 0.86 & 0.24 & 0.08 & 0.17 & 0.74 \\
& Chunked & \textcolor{magenta}{\textbf{0.97}} & 0.8 & \textcolor{magenta}{\textbf{0.88}} & 0.8 & 0.6 & 0.64 & \textcolor{magenta}{\textbf{0.83}} \\
& Characteristic Aware & 0.95 & 0.84 & 0.86 & 0.55 & 0.36 & 0.44 & 0.8 \\
& ConvoGen & 0.77 & 0.66 & 0.86 & 0.78 & 0.49 & 0.35 & 0.67 \\
& NoteChat & 0.91 & 0.8 & 0.8 & 0.79 & 0.55 & 0.65 & 0.79 \\ \midrule
\multirow{5}{*}{\shortstack{\texttt{Spanish}}} & 
Direct & \textcolor{magenta}{\textbf{0.98}} & \textcolor{magenta}{\textbf{0.88}} & 0.72 & 0.23 & 0.11 & 0.19 & 0.68 \\
& Chunked & \textcolor{magenta}{\textbf{0.98}} & 0.81 & 0.69 & 0.84 & 0.62 & 0.65 & 0.78 \\
& Characteristic Aware & 0.96 & 0.84 & \textcolor{magenta}{\textbf{0.75}} & 0.68 & 0.44 & 0.55 & \textcolor{magenta}{\textbf{0.80}} \\
& ConvoGen & 0.85 & 0.68 & 0.61 & 0.74 & 0.41 & 0.39 & 0.65 \\
& NoteChat & 0.95 & 0.82 & 0.71 & 0.81 & 0.54 & 0.63 & \textcolor{magenta}{\textbf{0.80}} \\ \midrule
\multirow{5}{*}{\shortstack{\texttt{French}}} & 
Direct & \textcolor{magenta}{\textbf{0.98}} & \textcolor{magenta}{\textbf{0.90}} & \textcolor{magenta}{\textbf{0.94}} & 0.25 & 0.14 & 0.21 & 0.75 \\
& Chunked & \textcolor{magenta}{\textbf{0.98}} & 0.83 & 0.88 & 0.89 & 0.69 & 0.71 & 0.84 \\
& Characteristic Aware & 0.96 & 0.88 & \textcolor{magenta}{\textbf{0.94}} & 0.65 & 0.51 & 0.5 & 0.84 \\
& ConvoGen & 0.85 & 0.70 & 0.87 & 0.75 & 0.46 & 0.35 & 0.71 \\
& NoteChat & 0.96 & 0.87 & 0.90 & 0.85 & 0.6 & 0.64 & \textcolor{magenta}{\textbf{0.86}} \\ \midrule
\multirow{5}{*}{\shortstack{\texttt{French-Canadian}}} & 
Direct & \textcolor{magenta}{\textbf{0.98}} & \textcolor{magenta}{\textbf{0.85}} & \textcolor{magenta}{\textbf{0.82}} & 0.34 & 0.15 & 0.24 & 0.73 \\
& Chunked & 0.97 & 0.77 & 0.78 & 0.86 & 0.67 & 0.67 & 0.79 \\
& Characteristic Aware & 0.97 & 0.82 & \textcolor{magenta}{\textbf{0.82}} & 0.63 & 0.43 & 0.48 & 0.8 \\
& ConvoGen & 0.88 & 0.73 & \textcolor{magenta}{\textbf{0.82}} & 0.83 & 0.54 & 0.35 & 0.74 \\
& NoteChat & 0.92 & 0.84 & 0.8 & 0.86 & 0.59 & 0.67 & \textcolor{magenta}{\textbf{0.81}} \\ \bottomrule
\end{tabular}%
}
\caption{Reconstruction score $(0\text{--}1$, $\uparrow$ better) for generating synthetic transcripts in the evaluation dataset, reported by language and method. For each language and metric, the highest score is highlighted in bold and color to indicate the top-performing method. The overall score is used to optimize prompts during prompt tuning and is computed as a weighted sum of the individual reconstruction scores. Details on the computation of the overall score are provided in Section~\ref{loss_function_prompt_tuning}.}
\label{reconstruction_score_table}
\end{table*}

\begin{table*}[h!]
\centering
\large
\resizebox{\textwidth}{!}{%
\begin{tabular}{clcccccccccc}
\toprule
\multirow{2}{*}{Language}  &
  Method &
  \multicolumn{2}{c}{Sentiment} &
  \multicolumn{2}{c}{Customer Emotion Arc$^{\dagger}$} &
  \multicolumn{2}{c}{Agent Emotion Arc$^{\dagger}$} &
  \multicolumn{2}{c}{Customer Sentiment Arc$^{\dagger}$} &
  \multicolumn{2}{c}{Agent Sentiment Arc$^{\dagger}$} \\
\cmidrule(lr){3-4} \cmidrule(lr){5-6} \cmidrule(lr){7-8} \cmidrule(lr){9-10} \cmidrule(lr){11-12}
&
&
\multicolumn{1}{c}{$\chi^2$/G (p)} &
\multicolumn{1}{c}{JS-Div} &
\multicolumn{1}{c}{$\chi^2$/G (p)} &
\multicolumn{1}{c}{JS-Div} &
\multicolumn{1}{c}{$\chi^2$/G (p)} &
\multicolumn{1}{c}{JS-Div} &
\multicolumn{1}{c}{$\chi^2$/G (p)} &
\multicolumn{1}{c}{JS-Div} &
\multicolumn{1}{c}{$\chi^2$/G (p)} &
\multicolumn{1}{c}{JS-Div} \\
\midrule
% --- your rows continue here ---
  \multirow{5}{*}{\shortstack{\texttt{English}}}       & Direct  & 0.000 & 0.024 & \textcolor{magenta}{\textbf{0.255}} & 0.206 & 0.023 & 0.159 & \textcolor{magenta}{\textbf{0.118}} & 0.059 & 0.005 & 0.138  \\
    & Chunked  & 0.000 & 0.021 & \textcolor{magenta}{\textbf{0.381}} & 0.162 & \textcolor{magenta}{\textbf{0.065}} & 0.169 & \textcolor{magenta}{\textbf{0.087}} & 0.073 & 0.032 & 0.108  \\
    & Characteristic Aware  & 0.000 & 0.020 & 0.038 & 0.275 & 0.007 & 0.182 & 0.008 & 0.110 & 0.011 & 0.118  \\
    & ConvoGen  & 0.000 & 0.028 & 0.004 & 0.330 & 0.027 & 0.170 & 0.011 & 0.085 & 0.013 & 0.123  \\
    & NoteChat  & 0.000 & 0.030 & 0.000 & 0.390 & 0.002 & 0.210 & 0.000 & 0.221 & 0.001 & 0.162  \\
 \midrule
  \multirow{5}{*}{\shortstack{\texttt{French}}}       & Direct  & 0.000 & 0.032 & \textcolor{magenta}{\textbf{0.213}} & 0.138 & 0.003 & 0.128 & 0.016 & 0.099 & 0.014 & 0.064  \\
    & Chunked  & 0.000 & 0.020 & \textcolor{magenta}{\textbf{0.633}} & 0.084 & \textcolor{magenta}{\textbf{0.470}} & 0.028 & \textcolor{magenta}{\textbf{0.100}} & 0.059 & \textcolor{magenta}{\textbf{0.459}} & 0.011  \\
    & Characteristic Aware  & 0.000 & 0.023 & \textcolor{magenta}{\textbf{0.379}} & 0.115 & 0.040 & 0.069 & 0.024 & 0.106 & 0.007 & 0.059  \\
    & ConvoGen  & 0.000 & 0.039 & \textcolor{magenta}{\textbf{0.051}} & 0.217 & 0.033 & 0.098 & 0.016 & 0.104 & \textcolor{magenta}{\textbf{0.597}} & 0.009  \\
    & NoteChat  & 0.000 & 0.018 & \textcolor{magenta}{\textbf{0.182}} & 0.136 & \textcolor{magenta}{\textbf{0.095}} & 0.068 & 0.007 & 0.113 & 0.028 & 0.055  \\
 \midrule
  \multirow{5}{*}{\shortstack{\texttt{French-Canadian}}}       & Direct  & 0.000 & 0.064 & 0.000 & 0.300 & 0.000 & 0.203 & 0.006 & 0.080 & 0.000 & 0.175  \\
    & Chunked  & 0.000 & 0.032 & \textcolor{magenta}{\textbf{0.203}} & 0.142 & 0.039 & 0.145 & \textcolor{magenta}{\textbf{0.080}} & 0.044 & 0.012 & 0.069  \\
    & Characteristic Aware  & 0.000 & 0.037 & \textcolor{magenta}{\textbf{0.089}} & 0.175 & 0.008 & 0.155 & \textcolor{magenta}{\textbf{0.273}} & 0.027 & 0.001 & 0.146  \\
    & ConvoGen  & 0.000 & 0.050 & 0.008 & 0.227 & 0.037 & 0.125 & 0.020 & 0.073 & 0.029 & 0.074  \\
    & NoteChat  & 0.000 & 0.031 & 0.000 & 0.246 & 0.000 & 0.253 & 0.000 & 0.199 & 0.000 & 0.237  \\
 \midrule
  \multirow{5}{*}{\shortstack{\texttt{Spanish}}}       & Direct  & 0.000 & 0.048 & \textcolor{magenta}{\textbf{0.157}} & 0.169 & \textcolor{magenta}{\textbf{0.080}} & 0.097 & \textcolor{magenta}{\textbf{0.077}} & 0.071 & 0.024 & 0.082  \\
    & Chunked  & 0.000 & 0.024 & \textcolor{magenta}{\textbf{0.080}} & 0.197 & \textcolor{magenta}{\textbf{0.279}} & 0.086 & 0.041 & 0.063 & \textcolor{magenta}{\textbf{0.170}} & 0.052  \\
    & Characteristic Aware  & 0.000 & 0.024 & \textcolor{magenta}{\textbf{0.190}} & 0.201 & \textcolor{magenta}{\textbf{0.158}} & 0.091 & \textcolor{magenta}{\textbf{0.081}} & 0.077 & \textcolor{magenta}{\textbf{0.058}} & 0.076  \\
    & ConvoGen  & 0.000 & 0.034 & 0.019 & 0.365 & \textcolor{magenta}{\textbf{0.726}} & 0.078 & 0.002 & 0.194 & \textcolor{magenta}{\textbf{0.488}} & 0.054  \\
    & NoteChat  & 0.000 & 0.031 & 0.001 & 0.287 & 0.010 & 0.149 & 0.000 & 0.188 & 0.006 & 0.106  \\
 \bottomrule
\end{tabular}%
}
\caption{Comparison of methods for transcript generation across multiple languages and metrics in the \textbf{Sentiment and Emotion} category. Legend: $\chi^2$/G (p) = Chi-Square / G-Test p-value; JS-Div = Jensen–Shannon Divergence. $^{\dagger}$ denotes transcript-level metrics; unmarked metrics are turn-level traits.}
\label{results_sent_and_emotion_all_languages}
\end{table*}

\setlength{\tabcolsep}{4pt}          % reduce column spacing
\renewcommand{\arraystretch}{1.1}    % (optional) slightly taller rows

\begin{table*}[h!]
\centering
\large
\resizebox{\textwidth}{!}{%
\begin{tabular}{clcccccccccc}
\toprule
\multirow{2}{*}{Language}  &
  \multicolumn{1}{l}{Method} &
  \multicolumn{2}{c}{Language Complexity} &
  \multicolumn{2}{c}{Technical Density$^{\dagger}$} &
  \multicolumn{2}{c}{Sentence Complexity$^{\dagger}$} &
  \multicolumn{2}{c}{Overall Readability$^{\dagger}$} &
  \multicolumn{2}{c}{Discourse Flow$^{\dagger}$} \\
\cmidrule(lr){3-4} \cmidrule(lr){5-6} \cmidrule(lr){7-8} \cmidrule(lr){9-10} \cmidrule(lr){11-12}
&
&
\multicolumn{1}{c}{$\chi^2$/G (p)} &
\multicolumn{1}{c}{JS-Div} &
\multicolumn{1}{c}{$\chi^2$/G (p)} &
\multicolumn{1}{c}{JS-Div} &
\multicolumn{1}{c}{$\chi^2$/G (p)} &
\multicolumn{1}{c}{JS-Div} &
\multicolumn{1}{c}{$\chi^2$/G (p)} &
\multicolumn{1}{c}{JS-Div} &
\multicolumn{1}{c}{$\chi^2$/G (p)} &
\multicolumn{1}{c}{JS-Div} \\ 
\midrule

\multirow{5}{*}{\texttt{English}} &
  Direct  & 0.000 & 0.057 & \textcolor{magenta}{\textbf{0.136}} & 0.034 & 0.001 & 0.107 & 0.001 & 0.110 & 0.000 & 0.526  \\
& Chunked  & 0.000 & 0.018 & 0.041 & 0.042 & 0.019 & 0.067 & 0.023 & 0.055 & 0.000 & 0.116  \\
& Characteristic Aware  & 0.000 & 0.047 & 0.014 & 0.058 & \textcolor{magenta}{\textbf{0.428}} & 0.014 & \textcolor{magenta}{\textbf{0.135}} & 0.031 & 0.000 & 0.270  \\
& ConvoGen  & 0.000 & 0.009 & \textcolor{magenta}{\textbf{0.275}} & 0.025 & 0.019 & 0.067 & 0.006 & 0.084 & 0.014 & 0.072  \\
& NoteChat  & 0.000 & 0.050 & \textcolor{magenta}{\textbf{0.090}} & 0.038 & 0.033 & 0.053 & 0.005 & 0.086 & 0.000 & 0.201  \\
\midrule

\multirow{5}{*}{\texttt{French}} &
  Direct  & 0.000 & 0.051 & \textcolor{magenta}{\textbf{0.236}} & 0.015 & 0.002 & 0.105 & 0.000 & 0.193 & 0.000 & 0.423  \\
& Chunked  & 0.000 & 0.011 & 0.019 & 0.057 & \textcolor{magenta}{\textbf{0.064}} & 0.050 & 0.000 & 0.147 & 0.011 & 0.075  \\
& Characteristic Aware  & 0.000 & 0.044 & \textcolor{magenta}{\textbf{0.377}} & 0.018 & \textcolor{magenta}{\textbf{0.124}} & 0.029 & 0.000 & 0.124 & 0.000 & 0.303  \\
& ConvoGen  & 0.000 & 0.028 & 0.023 & 0.064 & 0.019 & 0.085 & 0.000 & 0.171 & 0.000 & 0.227  \\
& NoteChat  & 0.000 & 0.103 & \textcolor{magenta}{\textbf{0.333}} & 0.019 & 0.041 & 0.044 & 0.000 & 0.174 & 0.000 & 0.219  \\
\midrule

\multirow{5}{*}{\texttt{French-Canadian}} &
  Direct  & 0.000 & 0.043 & 0.033 & 0.053 & 0.000 & 0.121 & 0.000 & 0.516 & 0.000 & 0.511  \\
& Chunked  & 0.000 & 0.007 & 0.013 & 0.063 & 0.003 & 0.092 & 0.000 & 0.511 & 0.000 & 0.328  \\
& Characteristic Aware  & 0.000 & 0.035 & 0.009 & 0.067 & \textcolor{magenta}{\textbf{0.168}} & 0.028 & 0.000 & 0.463 & 0.000 & 0.459  \\
& ConvoGen  & 0.000 & 0.032 & 0.008 & 0.073 & 0.003 & 0.096 & 0.000 & 0.577 & 0.000 & 0.439  \\
& NoteChat  & 0.000 & 0.089 & \textcolor{magenta}{\textbf{0.052}} & 0.042 & 0.002 & 0.100 & 0.000 & 0.511 & 0.000 & 0.484  \\
\midrule

\multirow{5}{*}{\texttt{Spanish}} &
  Direct  & 0.000 & 0.052 & 0.022 & 0.063 & 0.000 & 0.131 & 0.000 & 0.243 & 0.000 & 0.536  \\
& Chunked  & 0.000 & 0.014 & \textcolor{magenta}{\textbf{0.125}} & 0.033 & 0.050 & 0.034 & 0.006 & 0.082 & 0.000 & 0.119  \\
& Characteristic Aware  & 0.000 & 0.034 & 0.021 & 0.055 & \textcolor{magenta}{\textbf{0.052}} & 0.046 & 0.001 & 0.117 & 0.000 & 0.269  \\
& ConvoGen  & 0.000 & 0.042 & \textcolor{magenta}{\textbf{0.297}} & 0.047 & \textcolor{magenta}{\textbf{0.054}} & 0.072 & 0.009 & 0.142 & 0.000 & 0.322  \\
& NoteChat  & 0.000 & 0.081 & \textcolor{magenta}{\textbf{0.792}} & 0.007 & 0.028 & 0.041 & 0.000 & 0.120 & 0.000 & 0.234  \\
\bottomrule
\end{tabular}%
}
\caption{Comparison of methods for transcript generation across multiple languages in the \textbf{Linguistic Complexity and Content Density} category. 
Legend: $\chi^2$/G (p) = Chi-Square / G-Test p-value; JS-Div = Jensen–Shannon Divergence. 
$^{\dagger}$ denotes transcript-level metrics; unmarked metrics are turn-level traits.}
\label{results_linguistic_complexity_all_languages}
\end{table*}

\setlength{\tabcolsep}{4pt}
\renewcommand{\arraystretch}{1.1}

\begin{table*}[h!]
\centering
% \large
\resizebox{0.8\textwidth}{!}{%
\begin{tabular}{clcccccccc}
\toprule
\multirow{2}{*}{Language}  &
  \multicolumn{1}{l}{Method} &
  \multicolumn{2}{c}{Proactivity} &
  \multicolumn{2}{c}{Emphasis} &
  \multicolumn{2}{c}{Question Type} &
  \multicolumn{2}{c}{Solution} \\
\cmidrule(lr){3-4} \cmidrule(lr){5-6} \cmidrule(lr){7-8} \cmidrule(lr){9-10}
&
&
\multicolumn{1}{c}{$\chi^2$/G (p)} &
\multicolumn{1}{c}{JS-Div} &
\multicolumn{1}{c}{$\chi^2$/G (p)} &
\multicolumn{1}{c}{JS-Div} &
\multicolumn{1}{c}{$\chi^2$/G (p)} &
\multicolumn{1}{c}{JS-Div} &
\multicolumn{1}{c}{$\chi^2$/G (p)} &
\multicolumn{1}{c}{JS-Div} \\
\midrule

\multirow{5}{*}{\texttt{English}} &
  Direct  & 0.000 & 0.041 & 0.000 & 0.023 & 0.000 & 0.008 & 0.000 & 0.020  \\
& Chunked  & 0.000 & 0.008 & 0.000 & 0.023 & 0.000 & 0.003 & 0.000 & 0.021  \\
& Characteristic Aware  & 0.000 & 0.026 & 0.000 & 0.026 & 0.000 & 0.012 & 0.000 & 0.024  \\
& ConvoGen  & 0.000 & 0.019 & 0.000 & 0.068 & 0.000 & 0.006 & 0.000 & 0.044  \\
& NoteChat  & 0.000 & 0.008 & 0.000 & 0.008 & 0.000 & 0.022 & 0.000 & 0.021  \\
\midrule

\multirow{5}{*}{\texttt{French}} &
  Direct  & 0.000 & 0.044 & 0.000 & 0.037 & 0.000 & 0.005 & 0.000 & 0.032  \\
& Chunked  & 0.000 & 0.007 & 0.000 & 0.030 & 0.000 & 0.003 & 0.000 & 0.016  \\
& Characteristic Aware  & 0.000 & 0.034 & 0.000 & 0.031 & \textcolor{magenta}{\textbf{0.087}} & 0.001 & 0.000 & 0.035  \\
& ConvoGen  & 0.000 & 0.026 & 0.000 & 0.068 & 0.000 & 0.005 & 0.000 & 0.052  \\
& NoteChat  & 0.000 & 0.029 & 0.000 & 0.003 & 0.000 & 0.049 & 0.000 & 0.020  \\
\midrule

\multirow{5}{*}{\texttt{French-Canadian}} &
  Direct  & 0.000 & 0.051 & 0.000 & 0.057 & 0.000 & 0.013 & 0.000 & 0.022  \\
& Chunked  & 0.000 & 0.019 & 0.000 & 0.037 & 0.000 & 0.006 & 0.000 & 0.010  \\
& Characteristic Aware  & 0.000 & 0.043 & 0.000 & 0.030 & 0.000 & 0.013 & 0.000 & 0.015  \\
& ConvoGen  & 0.000 & 0.042 & 0.000 & 0.076 & 0.003 & 0.004 & 0.000 & 0.048  \\
& NoteChat  & 0.000 & 0.055 & 0.000 & 0.008 & 0.000 & 0.037 & 0.000 & 0.010  \\
\midrule

\multirow{5}{*}{\texttt{Spanish}} &
  Direct  & 0.000 & 0.047 & 0.000 & 0.043 & 0.000 & 0.007 & 0.000 & 0.025  \\
& Chunked  & 0.000 & 0.011 & 0.000 & 0.036 & 0.000 & 0.004 & 0.000 & 0.020  \\
& Characteristic Aware  & 0.000 & 0.022 & 0.000 & 0.037 & 0.000 & 0.003 & 0.000 & 0.032  \\
& ConvoGen  & 0.000 & 0.016 & 0.000 & 0.070 & \textcolor{magenta}{\textbf{0.072}} & 0.002 & 0.000 & 0.053  \\
& NoteChat  & 0.000 & 0.026 & 0.000 & 0.010 & 0.000 & 0.048 & 0.000 & 0.016  \\
\bottomrule
\end{tabular}%
}
\caption{Comparison of methods for transcript generation across multiple languages in the \textbf{Interaction Style and Operational} category. 
Legend: $\chi^2$/G (p) = Chi-Square / G-Test p-value; JS-Div = Jensen–Shannon Divergence.}
\label{results_interaction_style_all_languages}
\end{table*}

\setlength{\tabcolsep}{4pt}

\begin{table*}[h!]
\centering
\resizebox{0.75\textwidth}{!}{%
% \medium
\begin{tabular}{clcccccc}
\toprule
\multirow{2}{*}{\shortstack{Language}}  &
  \multicolumn{1}{l}{Method} &
  \multicolumn{2}{c}{Disfluency} &
  \multicolumn{2}{c}{Repetition} &
  \multicolumn{2}{c}{ASR Noise} \\
\cmidrule(lr){3-4} \cmidrule(lr){5-6} \cmidrule(lr){7-8}
&
&
\multicolumn{1}{c}{$\chi^2$/G (p)} &
\multicolumn{1}{c}{JS-Div} &
\multicolumn{1}{c}{$\chi^2$/G (p)} &
\multicolumn{1}{c}{JS-Div} &
\multicolumn{1}{c}{$\chi^2$/G (p)} &
\multicolumn{1}{c}{JS-Div} \\
\midrule
  \multirow{5}{*}{\shortstack{\texttt{English}}}       & Direct  & 0.000 & 0.126 & 0.000 & 0.003 & 0.000 & 0.106  \\
    & Chunked  & 0.000 & 0.086 & 0.001 & 0.001 & 0.000 & 0.065  \\
    & Characteristic Aware  & 0.000 & 0.075 & 0.000 & 0.002 & 0.000 & 0.065  \\
    & ConvoGen  & 0.000 & 0.321 & \textcolor{magenta}{\textbf{0.234}} & 0.000 & 0.000 & 0.128  \\
    & NoteChat  & 0.000 & 0.069 & 0.000 & 0.036 & 0.000 & 0.066  \\
 \midrule
  \multirow{5}{*}{\shortstack{\texttt{French}}}       & Direct  & 0.000 & 0.125 & \textcolor{magenta}{\textbf{0.158}} & 0.001 & 0.000 & 0.123  \\
    & Chunked  & 0.000 & 0.082 & 0.000 & 0.003 & 0.000 & 0.100  \\
    & Characteristic Aware  & 0.000 & 0.071 & \textcolor{magenta}{\textbf{0.075}} & 0.000 & 0.000 & 0.100  \\
    & ConvoGen  & 0.000 & 0.234 & \textcolor{magenta}{\textbf{0.647}} & 0.000 & 0.000 & 0.119  \\
    & NoteChat  & 0.000 & 0.074 & 0.000 & 0.031 & 0.000 & 0.095  \\
 \midrule
  \multirow{5}{*}{\shortstack{\texttt{French-Canadian}}}       & Direct  & 0.000 & 0.102 & \textcolor{magenta}{\textbf{0.632}} & 0.000 & 0.000 & 0.264  \\
    & Chunked  & 0.000 & 0.059 & 0.000 & 0.002 & 0.000 & 0.233  \\
    & Characteristic Aware  & 0.000 & 0.066 & \textcolor{magenta}{\textbf{0.669}} & 0.000 & 0.000 & 0.220  \\
    & ConvoGen  & 0.000 & 0.198 & \textcolor{magenta}{\textbf{0.152}} & 0.001 & 0.000 & 0.266  \\
    & NoteChat  & 0.000 & 0.057 & 0.000 & 0.040 & 0.000 & 0.229  \\
 \midrule
  \multirow{5}{*}{\shortstack{\texttt{Spanish}}}       & Direct  & 0.000 & 0.099 & 0.000 & 0.009 & 0.000 & 0.100  \\
    & Chunked  & 0.000 & 0.050 & 0.000 & 0.001 & 0.000 & 0.069  \\
    & Characteristic Aware  & 0.000 & 0.028 & 0.000 & 0.002 & 0.000 & 0.061  \\
    & ConvoGen  & 0.000 & 0.155 & 0.027 & 0.002 & 0.000 & 0.069  \\
    & NoteChat  & 0.000 & 0.043 & 0.000 & 0.037 & 0.000 & 0.066  \\
 \bottomrule
\end{tabular}%
}
\caption{Comparison of methods for transcript generation across multiple languages and metrics in the \textbf{Conversational Properties} category. Legend: $\chi^2$/G (p) = Chi-Square / G-Test p-value; JS-Div = Jensen–Shannon Divergence.}
\label{results_conversational_properties_all_languages}
\end{table*}

\begin{table*}
  \centering
  \setlength{\tabcolsep}{2pt}
  \tiny
  \begin{tabular}{l|ccccc|ccccc|ccccc}
    \hline
    \textbf{} & \multicolumn{5}{c}{\textbf{Direct}} & \multicolumn{5}{c}{\textbf{Chunked}} & \multicolumn{5}{c}{\textbf{Char-Aware}} \\
    \textbf{Metric} & \textbf{Baseline} & \textbf{\shortstack[c]{w/o \\ All \\ Attr}} & \textbf{\shortstack[c]{w/o \\ QA \\ Eval}} & \textbf{\shortstack[c]{w/o \\ Summary}} & \textbf{\shortstack[c]{w/o \\ Topic \\ Flow}} & \textbf{Baseline} & \textbf{\shortstack[c]{w/o \\ All \\ Attr}} & \textbf{\shortstack[c]{w/o \\ QA \\ Eval}} & \textbf{\shortstack[c]{w/o \\ Summary}} & \textbf{\shortstack[c]{w/o \\ Topic \\ Flow}} & \textbf{Baseline} & \textbf{\shortstack[c]{w/o \\ All \\ Attr}} & \textbf{\shortstack[c]{w/o \\ QA \\ Eval}} & \textbf{\shortstack[c]{w/o \\ Summary}} & \textbf{\shortstack[c]{w/o \\ Topic \\ Flow}} \\
    \hline
    \textbf{Sentiment} & 0.000 & 0.000 & 0.000 & 0.000 & 0.000 & 0.000 & 0.000 & 0.000 & 0.000 & 0.000 & 0.000 & 0.000 & 0.000 & 0.000 & 0.000 \\
    \textbf{Customer Emotion Arc} & \textcolor{magenta}{\textbf{0.255}} & 0.000 & \textcolor{magenta}{\textbf{0.143}} & 0.039 & \textcolor{magenta}{\textbf{0.212}} & \textcolor{magenta}{\textbf{0.381}} & \textcolor{magenta}{\textbf{0.283}} & \textcolor{magenta}{\textbf{0.268}} & \textcolor{magenta}{\textbf{0.575}} & \textcolor{magenta}{\textbf{0.469}} & 0.038 & 0.043 & \textcolor{magenta}{\textbf{0.144}} & \textcolor{magenta}{\textbf{0.115}} & \textcolor{magenta}{\textbf{0.309}} \\
    \textbf{Agent Emotion Arc} & 0.023 & 0.010 & \textcolor{magenta}{\textbf{0.087}} & \textcolor{magenta}{\textbf{0.063}} & 0.041 & \textcolor{magenta}{\textbf{0.065}} & \textcolor{magenta}{\textbf{0.411}} & \textcolor{magenta}{\textbf{0.571}} & \textcolor{magenta}{\textbf{0.665}} & \textcolor{magenta}{\textbf{0.597}} & 0.007 & \textcolor{magenta}{\textbf{0.176}} & \textcolor{magenta}{\textbf{0.129}} & \textcolor{magenta}{\textbf{0.251}} & \textcolor{magenta}{\textbf{0.310}} \\
    \textbf{Cust. Sentiment Arc} & \textcolor{magenta}{\textbf{0.118}} & 0.000 & \textcolor{magenta}{\textbf{0.088}} & 0.015 & \textcolor{magenta}{\textbf{0.133}} & \textcolor{magenta}{\textbf{0.087}} & \textcolor{magenta}{\textbf{0.121}} & \textcolor{magenta}{\textbf{0.175}} & \textcolor{magenta}{\textbf{0.497}} & \textcolor{magenta}{\textbf{0.665}} & 0.008 & 0.007 & \textcolor{magenta}{\textbf{0.066}} & 0.022 & \textcolor{magenta}{\textbf{0.102}} \\
    \textbf{Agent Sentiment Arc} & 0.005 & 0.007 & 0.016 & 0.011 & 0.012 & 0.032 & \textcolor{magenta}{\textbf{0.173}} & \textcolor{magenta}{\textbf{0.220}} & \textcolor{magenta}{\textbf{0.272}} & \textcolor{magenta}{\textbf{0.433}} & 0.011 & 0.002 & 0.022 & 0.031 & 0.030 \\
    \hline
    \textbf{Language Complexity} & 0.000 & 0.000 & 0.000 & 0.000 & 0.000 & 0.000 & 0.000 & 0.000 & 0.010 & 0.000 & 0.000 & 0.000 & 0.000 & 0.000 & 0.000 \\
    \textbf{Technical Density} & \textcolor{magenta}{\textbf{0.136}} & 0.000 & \textcolor{magenta}{\textbf{0.546}} & \textcolor{magenta}{\textbf{0.198}} & \textcolor{magenta}{\textbf{0.819}} & 0.041 & 0.000 & 0.000 & \textcolor{magenta}{\textbf{1.000}} & \textcolor{magenta}{\textbf{1.000}} & 0.014 & \textcolor{magenta}{\textbf{0.365}} & 0.015 & 0.024 & \textcolor{magenta}{\textbf{0.101}} \\
    \textbf{Sentence Complexity} & 0.001 & 0.000 & 0.002 & 0.000 & 0.046 & 0.019 & 0.010 & 0.002 & \textcolor{magenta}{\textbf{0.739}} & \textcolor{magenta}{\textbf{0.268}} & \textcolor{magenta}{\textbf{0.428}} & \textcolor{magenta}{\textbf{0.293}} & \textcolor{magenta}{\textbf{0.532}} & \textcolor{magenta}{\textbf{0.901}} & \textcolor{magenta}{\textbf{0.784}} \\
    \textbf{Overall Readability} & 0.001 & 0.000 & 0.005 & 0.000 & \textcolor{magenta}{\textbf{0.075}} & 0.023 & 0.003 & 0.002 & \textcolor{magenta}{\textbf{1.000}} & \textcolor{magenta}{\textbf{1.000}} & \textcolor{magenta}{\textbf{0.135}} & 0.002 & 0.006 & \textcolor{magenta}{\textbf{0.285}} & \textcolor{magenta}{\textbf{0.215}} \\
    \textbf{Discourse Flow} & 0.000 & 0.000 & 0.000 & 0.000 & 0.000 & 0.000 & 0.000 & 0.000 & \textcolor{magenta}{\textbf{0.147}} & \textcolor{magenta}{\textbf{0.577}} & 0.000 & 0.000 & 0.000 & 0.000 & 0.000 \\
    \hline
    \textbf{Proactivity} & 0.000 & 0.000 & 0.000 & 0.000 & 0.000 & 0.000 & 0.000 & 0.000 & 0.000 & 0.020 & 0.000 & 0.000 & 0.000 & 0.000 & 0.000 \\
    \textbf{Emphasis} & 0.000 & 0.000 & 0.000 & 0.000 & 0.000 & 0.000 & 0.000 & 0.000 & \textcolor{magenta}{\textbf{0.378}} & 0.000 & 0.000 & 0.000 & 0.000 & 0.000 & 0.000 \\
    \textbf{Question Type} & 0.000 & 0.000 & 0.000 & 0.000 & 0.000 & 0.000 & 0.000 & 0.008 & 0.000 & 0.002 & 0.000 & 0.000 & 0.000 & 0.000 & 0.000 \\
    \textbf{Solution} & 0.000 & 0.000 & 0.000 & 0.005 & 0.000 & 0.000 & 0.000 & 0.000 & 0.041 & \textcolor{magenta}{\textbf{0.553}} & 0.000 & 0.000 & 0.000 & 0.000 & 0.000 \\
    \hline
    \textbf{Disfluency} & 0.000 & 0.000 & 0.000 & 0.000 & 0.000 & 0.000 & 0.000 & 0.000 & 0.001 & 0.026 & 0.000 & 0.000 & 0.000 & 0.000 & 0.000 \\
    \textbf{Repetition} & 0.000 & 0.018 & 0.000 & 0.000 & 0.000 & 0.001 & \textcolor{magenta}{\textbf{0.527}} & \textcolor{magenta}{\textbf{0.886}} & 0.000 & 0.008 & 0.000 & 0.016 & 0.000 & 0.000 & 0.000 \\
    \textbf{ASR Noise} & 0.000 & 0.000 & 0.000 & 0.000 & 0.000 & 0.000 & 0.002 & 0.002 & \textcolor{magenta}{\textbf{0.172}} & \textcolor{magenta}{\textbf{0.424}} & 0.000 & 0.000 & 0.000 & 0.000 & 0.000 \\
    \hline
    \textbf{Summary} &  & 0 / 3 / 14 & 1 / 0 / 16 & 1 / 2 / 14 & 1 / 0 / 16 &  & 2 / 0 / 15 & 2 / 0 / 15 & 7 / 0 / 10 & 7 / 0 / 10 &  & 2 / 1 / 14 & 3 / 1 / 13 & 2 / 0 / 15 & 4 / 0 / 13 \\
    \hline
  \end{tabular}
  \caption{Ablation study p-values for Direct, Chunked, Char-Aware pipelines on English data. Each pipeline section shows baseline (all call attributes) p-values followed by ablation variant p-values from independence tests (Chi-square/G-test/Fisher's Exact). Values in \textcolor{magenta}{\textbf{magenta bold}} indicate p-value $> 0.05$ (independent distribution at $\alpha=0.05$). Summary row format: improved (Dependent$\rightarrow$Independent) / degraded (Independent$\rightarrow$Dependent) / unchanged.}
  \label{tab:multi_pipeline_ablation_pvalue}
\end{table*}

\begin{table*}
  \centering
  \small
  \setlength{\tabcolsep}{2pt}
  \begin{tabular}{l|cccc}
    \hline
    \textbf{Metric} & \textbf{GPT-4.1-mini} & \textbf{GPT-4.1} & \textbf{GPT-4o} & \textbf{Claude Haiku v3} \\
    \hline
    \textbf{Sentiment} & 0.000 & 0.000 & 0.000 & 0.000 \\
    \textbf{Customer Emotion Arc} & \textcolor{magenta}{\textbf{0.381}} & \textcolor{magenta}{\textbf{0.175}} & \textcolor{magenta}{\textbf{0.468}} & \textcolor{magenta}{\textbf{0.219}} \\
    \textbf{Agent Emotion Arc} & \textcolor{magenta}{\textbf{0.065}} & \textcolor{magenta}{\textbf{0.103}} & \textcolor{magenta}{\textbf{0.329}} & \textcolor{magenta}{\textbf{0.252}} \\
    \textbf{Customer Sentiment Arc} & \textcolor{magenta}{\textbf{0.087}} & \textcolor{magenta}{\textbf{0.127}} & \textcolor{magenta}{\textbf{0.300}} & \textcolor{magenta}{\textbf{0.213}} \\
    \textbf{Agent Sentiment Arc} & 0.032 & \textcolor{magenta}{\textbf{0.089}} & \textcolor{magenta}{\textbf{0.342}} & \textcolor{magenta}{\textbf{0.082}} \\
    \hline
    \textbf{Language Complexity} & 0.000 & 0.000 & 0.000 & 0.000 \\
    \textbf{Technical Density} & 0.041 & \textcolor{magenta}{\textbf{0.489}} & \textcolor{magenta}{\textbf{0.173}} & \textcolor{magenta}{\textbf{0.593}} \\
    \textbf{Sentence Complexity} & 0.019 & \textcolor{magenta}{\textbf{0.125}} & \textcolor{magenta}{\textbf{0.075}} & \textcolor{magenta}{\textbf{0.603}} \\
    \textbf{Overall Readability Score} & 0.023 & \textcolor{magenta}{\textbf{0.125}} & \textcolor{magenta}{\textbf{0.565}} & \textcolor{magenta}{\textbf{0.351}} \\
    \textbf{Discourse Flow} & 0.000 & \textcolor{magenta}{\textbf{0.071}} & \textcolor{magenta}{\textbf{0.125}} & 0.042 \\
    \hline
    \textbf{Proactivity} & 0.000 & 0.000 & 0.001 & 0.000 \\
    \textbf{Emphasis} & 0.000 & 0.000 & 0.000 & 0.000 \\
    \textbf{Question Type} & 0.000 & 0.000 & 0.000 & 0.000 \\
    \textbf{Solution} & 0.000 & 0.000 & 0.000 & 0.000 \\
    \hline
    \textbf{Disfluency} & 0.000 & 0.000 & 0.000 & 0.000 \\
    \textbf{Repetition} & 0.001 & 0.000 & \textcolor{magenta}{\textbf{0.111}} & 0.000 \\
    \textbf{ASR Noise} & 0.000 & 0.000 & 0.000 & 0.000 \\
    \hline
    \textbf{Summary} &  & 5 / 0 / 12 & 6 / 0 / 11 & 4 / 0 / 13 \\
    \hline
  \end{tabular}
  \caption{Model ablation study p-values comparing GPT-4.1-mini (baseline) with GPT-4.1, GPT-4o, Claude Haiku v3 on English data for Chunked Enhancement pipeline. Each column shows p-values from independence tests (Chi-square/G-test/Fisher's Exact). Values in \textcolor{magenta}{\textbf{magenta bold}} indicate p-value $> 0.05$ (independent distribution at $\alpha=0.05$). Summary row format: improved (Dependent$\rightarrow$Independent) / degraded (Independent$\rightarrow$Dependent) / unchanged.}
  \label{tab:model_ablation_pvalue}
\end{table*}

\begin{table*}[t]
\centering
\small  % Reduce font size to help with fit
\begin{tabularx}{\textwidth}{|>{\raggedright\arraybackslash}p{0.8in}|
                               >{\raggedright\arraybackslash}p{1.3in}|
                               >{\raggedright\arraybackslash}X|}
\hline
\textbf{Metric Name} & \textbf{Brief Description} & \textbf{Original Labels} \\
\hline

\multicolumn{3}{|l|}{\textbf{1. Emotional \& Sentiment Metrics}} \\
\hline
Agent \& Customer Emotion Arc & Tracks emotional trajectory from conversation start to end. & An arc from 8 start emotions to 8 end emotions: \texttt{gratitude}, \texttt{relief}, \texttt{factual}, \texttt{curiosity}, \texttt{confusion}, \texttt{frustration}, \texttt{anger}, \texttt{anxiety} \\
Agent \& Customer Sentiment Arc &  Tracks sentiment trajectory from conversation start to end. & An arc from positive/neutral/negative to positive/neutral/negative \\
Sentiment (Turn-Level) & Classifies each turn's tone. & \texttt{positive\_sentiment}, \texttt{neutral\_sentiment}, \texttt{negative\_sentiment} \\
\hline

\multicolumn{3}{|l|}{\textbf{2. Linguistic Complexity \& Content Density}} \\
\hline
Language Complexity (Turn-Level) & Language structure. & \texttt{simple\_informal\_language}, \texttt{simple\_formal\_language}, \texttt{complex\_informal\_language}, \texttt{complex\_formal\_language} \\
Technical Density & Transcript score measuring jargon prevalence. & Score from 1-5 (1: high density, 5: low density) \\
Sentence Complexity & Transcript score evaluating structural complexity. & Score from 1-5 (1: highly complex, 5: very simple) \\
Discourse Flow & Transcript score assessing coherence. & Score from 1-5 (1: poor, 5: excellent) \\
Overall Readability Score & Combined transcript metric. & Score from 1-5 (1: very hard to read, 5: very easy to read) \\
\hline

\multicolumn{3}{|l|}{\textbf{3. Interaction Style and Operational}} \\
\hline
Proactivity (Agent Turns) & Agent's initiative level. & \texttt{neutral\_proactivity}, \texttt{overstated\_proactivity}, \texttt{understated\_proactivity} \\
Emphasis & Turn's focus. & \texttt{emotion\_focused}, \texttt{fact\_focused}, \texttt{balanced} \\
Question Type & Classifies questions. & \texttt{no\_question}, \texttt{closed\_question}, \texttt{informational\_question}, \texttt{conversational\_question} \\
Solution & Agent's contribution to resolving issues. & \texttt{solution\_oriented}, \texttt{process\_oriented}, \texttt{no\_solution} \\
\hline

\multicolumn{3}{|l|}{\textbf{4. Conversational Properties}} \\
\hline
Repetition & Information repetition patterns. & \texttt{no\_repetition}, \texttt{agent\_repetition}, \texttt{customer\_repetition} \\
Disfluency & Detects speech disruptions (multi-select). & \texttt{speech\_repair\_repetition}, \texttt{hesitation\_fillers}, \texttt{interactional\_disfluency}, \texttt{comprehension\_clarity\_issues}, \texttt{no\_disfluency} \\
ASR Noise Type & Simulates transcription errors (multi-select). & \texttt{no\_noise}, \texttt{substitution}, \texttt{insertion}, \texttt{deletion} \\
\hline
\end{tabularx}
\caption{Comprehensive Evaluation Metrics and their corresponding categories}
\label{tab:comprehensive_eval_metrics_original}
\end{table*}

\clearpage

\clearpage
\begin{table*}[t]
\centering
\small  % Reduce font size to help with fit
\begin{tabularx}{\textwidth}{|>{\raggedright\arraybackslash}p{0.8in}|
                                >{\raggedright\arraybackslash}p{1.1in}|
                                >{\raggedright\arraybackslash}X|}
\hline
\textbf{Stage} & \textbf{Inputs Considered} & \textbf{Prompt} \\
\hline

\hline
\textbf{Base Conversation Generation} & 
call\_duration, language, key\_events, next\_steps, reason\_for\_call, customer\_complaints, key\_entities, hold\_and\_transfer, resolution, topic\_flow and qa\_evaluation &
\ttfamily Generate a realistic, conversational call transcript based on the provided call attributes. The transcript must accurately reflect the topic flow, incorporate details from the QA evaluation, and intent summaries, match the expected call duration, and be in the specified language. IMPORTANT: YOU MUST ENCORPORATE ALL AND THE GIVEN CALL ATTRIBUTES. Consider cultural nuances, appropriate sentiment expressions, and formal/informal registers (e.g., 'tu' vs. 'vous' in French, 'tú' vs. 'usted' in Spanish) suitable for the language and call center context. If relevant for the language and context, include domain-specific vocabulary (e.g., banking, telecom) and be mindful of potential mixed-language use or code-switching (e.g., English terms in Spanish/French/Turkish conversations). Note that the transcript should resemble a output from an ASR system so the transcript should contain numbers only in words. Crucially, the transcript should include natural human conversational disfluencies like filler words ('um', 'uh', 'like'), pauses (represented as '...'), repetitions, and self-corrections to mimic a real conversation.
\\\hline

\textbf{Transcript Segmentation} & 
current\_transcript &
\ttfamily You are an AI assistant who can segment call center conversations between an agent and a customer into coherent topics. List each topic along with its name, brief description, and the start and end turn indices that define its chunk in the conversation.

\ttfamily Important rules for segmentation:
\ttfamily 1. Topics must be in disjoint chunks - no overlapping turns between topics
\ttfamily 2. All the turns in the transcript should be included
\ttfamily 3. Each chunk should be defined by a start\_turn and end\_turn index (inclusive)
\ttfamily 4. Chunks must be sequential - there should be no gaps between chunks
\ttfamily 5. A topic can appear multiple times in the conversation, but as separate non-overlapping chunks
\ttfamily 6. Try to maximize the size of each chunk (atmost 25 turns) by categorizing turns into the most appropriate topic

\ttfamily Response format should be standard across all input conversations and should be json parsable programmatically. The input transcript is given as a json array of turns. Each topic should have: name, description, start\_turn, and end\_turn.
\\\hline

\textbf{Candidate Turn Identification - Characteristic Aware approach} & 
transcript\_chunk, characteristic\_type and characteristic\_categories & 
\ttfamily Analyze a transcript chunk and identify which turns can be categorized into each class of the given characteristic type.
    
    \ttfamily For the characteristic type {characteristic\_type}, examine each turn and classify it into one of the available categories.
    Provide the turn numbers (as they appear in the transcript) for each category.
    
    \ttfamily Important:
    \ttfamily - Consider the speaker context (agent vs customer) when relevant
    \ttfamily - A turn can only belong to one category per characteristic type
    \ttfamily - Provide turn numbers as they appear in the transcript (with parentheses)
    \ttfamily - If no turns fit a category, return an empty list for that category
\\\hline

\textbf{Characteristic Application To Identified Turns - Characteristic Aware approach} & 
transcript\_chunk, modification\_instructions and language & 
\ttfamily Apply specific characteristics to designated turns in a transcript chunk.

    \ttfamily You will receive a transcript chunk and specific instructions for which turns should be modified with which characteristics.

    \ttfamily CRITICAL RULES:
    \ttfamily 1. Only modify the turns specified in the modification instructions
    \ttfamily 2. Non-target turns must remain exactly unchanged
    \ttfamily 3. Apply the characteristics naturally and authentically for the target language
    \ttfamily 4. Maintain conversation flow and context
    \ttfamily 5. Each turn can have multiple characteristics applied simultaneously

    \ttfamily For each characteristic, follow these guidelines:
    \ttfamily - DISFLUENCIES: Insert natural speech patterns, fillers, hesitations appropriate for the language
    \ttfamily - ASR\_NOISE: Introduce plausible speech recognition errors
    \ttfamily - QUESTION\_TYPE: Ensure the turn exhibits the specified question type
    \ttfamily - SENTIMENT: Adjust language to reflect the specified sentiment
    \ttfamily - etc.
\\\hline
\end{tabularx}
\caption{Prompts used in the Characteristic-Aware Enhancement pipeline.}
\label{prompts_for_transcript_generation_1}
\end{table*}

\begin{table*}[t]
\centering
\small  % Reduce font size to help with fit
\begin{tabularx}{\textwidth}{|>{\raggedright\arraybackslash}p{0.8in}|
                                >{\raggedright\arraybackslash}p{1.1in}|
                                >{\raggedright\arraybackslash}X|}
\hline
\textbf{Stage} & \textbf{Inputs Considered} & \textbf{Prompt} \\
\hline

\hline
\textbf{Measuring Fulfillment of Intent Summary in Synthetic Transcript} & 
synthetic\_transcript, key\_events, next\_steps, reason\_for\_call, customer\_complaints, key\_entities, hold\_and\_transfer and resolution & 
\ttfamily Evaluate how well a synthetic transcript aligns with each provided intent summary (e.g., key events, next steps, resolution, etc.).

    \ttfamily Each intent summary describes a specific aspect of the call. The evaluator should compare the provided summary with the synthetic transcript and assign a score from 1 to 10 based on how completely and accurately the intent is reflected.

    \ttfamily **Scoring Guide (per intent):**
    \ttfamily - Score of 10: All and only the elements described in the intent summary are clearly and accurately reflected in the transcript.
    \ttfamily - Score of 1-9: Partial or incorrect coverage (e.g., missing, paraphrased incorrectly, added extra content). For each missing/partially missing or extra element, reduce the score by 1.
    \ttfamily - Score of 0: The input summary for this intent is empty or unavailable.
\\\hline

\textbf{Replication of QA Scenario in Synthetic Transcript} & 
synthetic\_transcript and qa\_evaluation & 
\ttfamily Evaluate whether specific QA evaluation scenarios are reflected in the transcript. QA evaluation is a list of questions regarding the call quality and the corresponding answers.
    
    \ttfamily The output is a list of scores for each question. The score is 1 if the evaluated answer is matching with the provided answer in the qa\_evaluation.
\\\hline

\textbf{Naturalness Assessment in Synthetic Transcript} & 
synthetic\_transcript & 
\ttfamily Evaluate the overall naturalness of a transcript by assessing three key speech characteristics:
    \ttfamily 1. Interruptions: Where one speaker starts talking before the other finishes, creating overlapping speech
    \ttfamily 2. Disfluencies: Natural speech patterns like fillers, hesitations, repetitions, and self-corrections.
    \ttfamily 3. ASR Noise: Plausible speech recognition errors that would occur in real transcription systems

    \ttfamily A high-quality transcript should incorporate these elements naturally to simulate realistic human conversation.
    
    \ttfamily Examples to look for:
    
    \ttfamily \# Interruptions:
    - Speaker overlaps where one person starts talking before another finishes
    - Turn sequences that indicate interruption patterns (e.g abrupt ending of one's turn)
    - Truncated sentences or phrases that suggest someone was cut off
    
    \ttfamily \# Disfluencies:
    - Filler words appropriate to the language (e.g., 'um', 'uh', 'euh', 'ben' in French, 'em', 'eh' in Spanish)
    - Repetitions of words or phrases ("I think, I think we should...")
    - Self-corrections ("We'll send it on Monday- I mean Tuesday")
    - Hesitations indicated by pauses or incomplete thoughts
    - Short acknowledgment turns ("Ok", "Yeah", "Mmhmm") that show active listening
    
    \ttfamily \# ASR Noise:
    - Phonetically similar word substitutions (e.g., "two" vs "too")
    - Plausible homophone errors
    - Slight word deletions or insertions
    - Mis-segmentations or omitted diacritics (in applicable languages)
    - Regional accent-based misrecognitions (e.g., "twos day" for "Tuesday")
    
    \ttfamily The assessment should consider both the presence and the natural integration of these elements.
\\\hline
\end{tabularx}
\caption{Prompts used in Reconstruction score calculation}
\label{prompts_for_transcript_generation_3}
\end{table*}
%\end{comment}

% ========================== Example ======================
\clearpage

\begin{table*}[t]
\centering
\small  % Reduce font size to help with fit
\begin{tabularx}{\textwidth}{|>{\raggedright\arraybackslash}p{0.8in}|
                                >{\raggedright\arraybackslash}p{1.1in}|
                                >{\raggedright\arraybackslash}X|}
\hline
\textbf{Stage} & \textbf{Inputs Considered} & \textbf{Output of Stage} \\
\hline

\hline

\textbf{Topic Flow Generation} & 
original\_transcript & 
\begin{flushleft}
\texttt{Initial Greeting → Life Insurance Verification → Policy Quote Discussion →}\\
\texttt{Call Transfer Preparation → Life Insurance Application → Accidental Death Insurance → Closing Conversation}
\end{flushleft}
\\\hline

\textbf{Intent Summary Input Generation: Key Events} & 
original\_transcript & 
\begin{flushleft}
\texttt{- Agent contacts customer to discuss life insurance options.}\\
\texttt{- Customer confirms personal details and existing coverage.}\\
\texttt{- Customer agrees to new \$25k policy with monthly premium.}\\
\texttt{- Additional accidental death policy added.}\\
\texttt{- Beneficiaries designated; agent confirms application details.}
\end{flushleft}
\\\hline

\textbf{Intent Summary Input Generation: Next Steps} & 
original\_transcript & 
\begin{flushleft}
\texttt{- Customer will receive policy documents by mail.}\\
\texttt{- Initial premiums to be drafted in next 3 business days.}\\
\texttt{- Coverage confirmation via email.}\\
\texttt{- Contact info shared for follow-up or questions.}
\end{flushleft}
\\\hline

\textbf{Intent Summary Input Generation: Reason for Call} & 
original\_transcript & 
\begin{flushleft}
\texttt{Agent initiates call to review customer's current life insurance status and offer updated, more affordable policy options.}
\end{flushleft}
\\\hline

\textbf{Intent Summary Input Generation: Customer Complaints} & 
original\_transcript & 
\begin{flushleft}
\texttt{Customer expresses concern about having sufficient coverage to leave for their children. Requests increased policy value.}
\end{flushleft}
\\\hline

\textbf{Intent Summary Input Generation: Key Entities} & 
original\_transcript & 
\begin{flushleft}
\texttt{Names: [REDACTED],}\\
\texttt{Policy IDs, Contact Info, DOBs, Addresses: [MASKED FOR PRIVACY]}\\
\texttt{Used for validation and beneficiary designation steps.}
\end{flushleft}
\\\hline

\textbf{Intent Summary Input Generation: Hold and Transfer} & 
original\_transcript & 
\begin{flushleft}
\texttt{Class: Transfer}\\
\texttt{Evidence: Initial agent offers to transfer call to licensed Illinois agent.}\\
\texttt{Next speaker is a different agent, indicating successful transfer.}
\end{flushleft}
\\\hline

\textbf{Intent Summary Input Generation: Resolution} & 
original\_transcript & 
\begin{flushleft}
\texttt{Class: Resolved}\\
\texttt{Customer successfully applies for new insurance policies. Agent confirms submission and provides policy numbers.}
\end{flushleft}
\\\hline

\textbf{Intent Summary Generation} & 
original\_transcript & 
\begin{flushleft}
\texttt{agent intends to verify customer information and share updated life insurance options.}\\
\texttt{customer expresses interest in understanding rate changes and verifying recent charges.}
\end{flushleft}
\\\hline

\textbf{Topic Flow Generation} & 
original\_transcript & 
\begin{flushleft}
\texttt{(1) agent greeting and identification,}\\
\texttt{(2) introduction of updated insurance offerings,}\\
\texttt{(3) customer inquiries about recent charges,}\\
\texttt{(4) verification process,}\\
\texttt{(5) resolution and call closure.}
\end{flushleft}
\\\hline
\end{tabularx}
\caption{Example prompts used in generating the input call attributes from real transcripts.}
\label{examples_dual_stage_v1_1}
\end{table*}

\begin{table*}[t]
\centering
\small  % Reduce font size to help with fit
\begin{tabularx}{\textwidth}{|>{\raggedright\arraybackslash}p{0.8in}|
                                >{\raggedright\arraybackslash}p{1.1in}|
                                >{\raggedright\arraybackslash}X|}
\hline
\textbf{Stage} & \textbf{Inputs Considered} & \textbf{Output of Stage} \\
\hline

\hline
\textbf{Base Conversation Generation} & 
call\_duration, language, key\_events, next\_steps, reason\_for\_call, customer\_complaints, key\_entities, hold\_and\_transfer, resolution, topic\_flow, qa\_evaluation &
\begin{flushleft}
\texttt{agent: hello, may i speak with robin frederick, please? this is natalie calling from ins insure life.}\\
\texttt{customer: yes, this is robin. how can i help you?}\\
\texttt{agent: hi robin, i’m reaching out because we’ve recently updated our life insurance rates, and i wanted to make sure you have the most affordable coverage that fits your needs.}\\ 
\textbf{...}\\
\texttt{customer: no, that covers everything for now.}\\
\texttt{agent: alright, thank you for your time. have a great day!}\\
\texttt{customer: you too, goodbye.}
\end{flushleft}
\\\hline

\textbf{Transcript Segmentation} & 
base\_transcript & 
\begin{flushleft}
\texttt{\textbf{chunk 1:}}\\
\texttt{agent: hello, may i speak with robin frederick, please? this is natalie calling from ins insure life.}\\
\texttt{customer: yes, this is robin. how can i help you?}\\
\texttt{agent: hi robin, i’m reaching out because we’ve recently updated our life insurance rates, and i wanted to make sure you have the most affordable coverage that fits your needs.}\\ 
\textbf{...}\\
\texttt{\textbf{chunk 7:}}\\
\texttt{customer: no, that covers everything for now.}\\
\texttt{agent: alright, thank you for your time. have a great day!}\\
\texttt{customer: you too, goodbye.}
\end{flushleft}
\\\hline

\textbf{Combined Transcript Enhancement} & 
transcript\_chunks, disfluency\_types, language & 
\begin{flushleft}
\texttt{\textbf{chunk 1:}}\\
\texttt{agent: hello may i speak with robin frederick please this is natalie calling from ins insure life}\\
\texttt{agent: hi robin i’m reaching out because we’ve recently updated our life insurance rates}\\
\texttt{agent: and i wanted to make sure you have the most affordable coverage that fits your needs}\\
\texttt{\textbf{chunk 2:}}\\
\texttt{agent: so um first we need to we need to verify your account details to make sure everything is up to date}\\
\texttt{customer: sorry i’m a bit lost could you repeat that}
\end{flushleft}
\\\hline

\textbf{Final synthetic transcript} & 
transcript\_chunks & 
\begin{flushleft}
\texttt{agent: hello may i speak with robin frederick please this is natalie calling from ins insure life}\\
\texttt{customer: yes this is robin how can i help you today}\\
\texttt{agent: hi robin i’m reaching out because we’ve recently updated our life insurance rates}\\
\texttt{agent: and i wanted to make sure you have the most affordable coverage that fits your needs}\\
\texttt{customer: okay that sounds good what do you need from me}\\
\texttt{agent: so um first we need to we need to verify your account details to make sure everything is up to date}\\
\texttt{...}\\
\texttt{agent: sorry i’m a bit lost could you repeat that}\\
\texttt{customer: sure i thought i saw a charge last week but maybe it was something else}\\
\texttt{agent: ah okay let me check that for you}\\
\texttt{agent: umm just a moment please}
\end{flushleft}
\\\hline
\end{tabularx}
\caption{Example of the various generation stages for Chunked Enhancement pipeline}
\label{examples_dual_stage_v1_2}
\end{table*}

\begin{table*}[h!]
\centering
\fontsize{6}{8}\selectfont
\begin{tabular}{llccccccc}
\toprule
\multirow{2}{*}{\textbf{Language}} & \multirow{2}{*}{\textbf{Category}} & \multicolumn{2}{c}{\textbf{Original Transcripts}} & \multicolumn{5}{c}{\textbf{Generation Methods}} \\
\cmidrule(lr){3-4} \cmidrule(lr){5-9}
 &  & Test Set & Train Set & Direct & Chunked & Characteristic Aware & ConvoGen & NoteChat \\
\midrule
\multirow{3}{*}{\texttt{English}} & Positive Sentiment & 6.52\% & 5.43\% & 21.23\% & 20.32\% & 19.06\% & 22.37\% & 21.25\% \\
 & Neutral Sentiment & 86.20\% & 88.33\% & 71.19\% & 73.57\% & 71.59\% & 68.73\% & 76.69\% \\
 & Negative Sentiment & 7.28\% & 6.24\% & 7.57\% & 6.11\% & 9.35\% & 8.90\% & 2.06\% \\
\midrule
\multirow{3}{*}{\texttt{French}} & Positive Sentiment & 6.43\% & 5.93\% & 21.66\% & 16.91\% & 18.10\% & 20.41\% & 15.44\% \\
 & Neutral Sentiment & 87.47\% & 89.76\% & 67.24\% & 71.95\% & 70.65\% & 63.70\% & 83.20\% \\
 & Negative Sentiment & 6.10\% & 4.31\% & 11.10\% & 11.14\% & 11.25\% & 15.89\% & 1.36\% \\
\midrule
\multirow{3}{*}{\texttt{French-Canadian}} & Positive Sentiment & 3.98\% & 3.57\% & 22.86\% & 16.49\% & 18.46\% & 17.85\% & 17.78\% \\
 & Neutral Sentiment & 82.69\% & 82.53\% & 52.01\% & 62.33\% & 61.46\% & 54.95\% & 76.23\% \\
 & Negative Sentiment & 13.33\% & 13.90\% & 25.13\% & 21.18\% & 20.08\% & 27.20\% & 5.99\% \\
\midrule
\multirow{3}{*}{\texttt{Spanish}} & Positive Sentiment & 6.77\% & 7.00\% & 28.60\% & 20.14\% & 19.31\% & 19.91\% & 21.94\% \\
 & Neutral Sentiment & 88.02\% & 85.90\% & 63.30\% & 71.23\% & 70.81\% & 68.38\% & 77.15\% \\
 & Negative Sentiment & 5.21\% & 7.10\% & 8.10\% & 8.63\% & 9.88\% & 11.71\% & 0.91\% \\
\bottomrule
\end{tabular}
\caption{Frequency distribution of \textbf{Sentiment} categories for real and synthetic transcripts on the evaluation dataset across generation pipelines (used for G-test and Chi-Square Test). Distributions from the training set of original transcripts are also provided for reference.}
\label{freq_dist_sentiment}
\end{table*}

\begin{table*}[h!]
\centering
\fontsize{6}{8}\selectfont
\begin{tabular}{lccccccc}
\toprule
\multirow{2}{*}{\textbf{Category}} & \multicolumn{2}{c}{\textbf{Original Transcripts}} & \multicolumn{5}{c}{\textbf{Generation Methods}} \\
\cmidrule(lr){2-3} \cmidrule(lr){4-8}
 & Test Set & Train Set & Direct & Chunked & Characteristic Aware & ConvoGen & NoteChat \\
\midrule
Anxiety To Anxiety & 4.26\% & 3.85\% & 0.00\% & 0.00\% & 0.00\% & 0.00\% & 0.00\% \\
Anxiety To Factual & 0.00\% & 1.81\% & 2.13\% & 0.00\% & 0.00\% & 4.26\% & 0.00\% \\
Anxiety To Frustration & 2.13\% & 2.22\% & 0.00\% & 0.00\% & 0.00\% & 0.00\% & 0.00\% \\
Anxiety To Gratitude & 0.00\% & 1.14\% & 0.00\% & 0.00\% & 4.26\% & 8.51\% & 0.00\% \\
Anxiety To Relief & 0.00\% & 0.30\% & 4.26\% & 2.13\% & 10.64\% & 25.53\% & 25.53\% \\
Confusion To Anxiety & 0.00\% & 0.30\% & 0.00\% & 0.00\% & 0.00\% & 2.13\% & 0.00\% \\
Confusion To Factual & 4.26\% & 3.34\% & 0.00\% & 6.38\% & 2.13\% & 2.13\% & 2.13\% \\
Confusion To Frustration & 4.26\% & 4.65\% & 2.13\% & 0.00\% & 6.38\% & 2.13\% & 0.00\% \\
Confusion To Gratitude & 0.00\% & 1.14\% & 2.13\% & 0.00\% & 8.51\% & 4.26\% & 8.51\% \\
Confusion To Relief & 6.38\% & 6.58\% & 2.13\% & 4.26\% & 0.00\% & 0.00\% & 12.77\% \\
Curiosity To Anxiety & 0.00\% & 0.04\% & 0.00\% & 0.00\% & 2.13\% & 0.00\% & 0.00\% \\
Curiosity To Confusion & 0.00\% & 1.85\% & 2.13\% & 0.00\% & 0.00\% & 0.00\% & 0.00\% \\
Curiosity To Curiosity & 2.13\% & 2.30\% & 0.00\% & 0.00\% & 0.00\% & 0.00\% & 0.00\% \\
Curiosity To Factual & 2.13\% & 1.79\% & 6.38\% & 6.38\% & 0.00\% & 6.38\% & 10.64\% \\
Curiosity To Frustration & 2.13\% & 1.77\% & 0.00\% & 0.00\% & 4.26\% & 2.13\% & 0.00\% \\
Curiosity To Gratitude & 4.26\% & 3.54\% & 12.77\% & 8.51\% & 8.51\% & 14.89\% & 14.89\% \\
Curiosity To Relief & 4.26\% & 3.74\% & 2.13\% & 8.51\% & 6.38\% & 4.26\% & 8.51\% \\
Factual To Anxiety & 2.13\% & 2.05\% & 0.00\% & 0.00\% & 0.00\% & 0.00\% & 0.00\% \\
Factual To Confusion & 2.13\% & 1.97\% & 0.00\% & 4.26\% & 0.00\% & 4.26\% & 0.00\% \\
Factual To Curiosity & 2.13\% & 1.86\% & 2.13\% & 0.00\% & 0.00\% & 0.00\% & 0.00\% \\
Factual To Factual & 6.38\% & 6.35\% & 8.51\% & 4.26\% & 0.00\% & 4.26\% & 0.00\% \\
Factual To Frustration & 14.89\% & 12.13\% & 2.13\% & 6.38\% & 2.13\% & 2.13\% & 0.00\% \\
Factual To Gratitude & 21.28\% & 18.58\% & 29.79\% & 27.66\% & 25.53\% & 8.51\% & 6.38\% \\
Factual To Relief & 8.51\% & 7.67\% & 8.51\% & 10.64\% & 12.77\% & 2.13\% & 2.13\% \\
Frustration To Confusion & 2.13\% & 1.99\% & 0.00\% & 0.00\% & 0.00\% & 0.00\% & 0.00\% \\
Frustration To Factual & 2.13\% & 2.26\% & 4.26\% & 0.00\% & 0.00\% & 0.00\% & 0.00\% \\
Frustration To Frustration & 0.00\% & 0.38\% & 2.13\% & 0.00\% & 0.00\% & 0.00\% & 0.00\% \\
Frustration To Gratitude & 2.13\% & 2.04\% & 0.00\% & 0.00\% & 0.00\% & 0.00\% & 0.00\% \\
Frustration To Relief & 0.00\% & 1.13\% & 4.26\% & 8.51\% & 4.26\% & 2.13\% & 8.51\% \\
Gratitude To Gratitude & 0.00\% & 0.09\% & 2.13\% & 0.00\% & 2.13\% & 0.00\% & 0.00\% \\
Gratitude To Relief & 0.00\% & 1.16\% & 0.00\% & 2.13\% & 0.00\% & 0.00\% & 0.00\% \\
\bottomrule
\end{tabular}
\caption{Frequency distribution of \textbf{Customer Emotion Arc} categories for real and synthetic transcripts on the evaluation dataset across generation pipelines (used for G-test and Chi-Square Test) - English. Distributions from the training set of original transcripts are also provided for reference.}
\label{freq_dist_customer_emotion_english}
\end{table*}

\begin{table*}[h!]
\centering
\fontsize{6}{8}\selectfont
\begin{tabular}{lccccccc}
\toprule
\multirow{2}{*}{\textbf{Category}} & \multicolumn{2}{c}{\textbf{Original Transcripts}} & \multicolumn{5}{c}{\textbf{Generation Methods}} \\
\cmidrule(lr){2-3} \cmidrule(lr){4-8}
 & Test Set & Train Set & Direct & Chunked & Characteristic Aware & ConvoGen & NoteChat \\
\midrule
Anxiety To Gratitude & 0.00\% & 0.00\% & 2.17\% & 2.17\% & 0.00\% & 2.50\% & 0.00\% \\
Anxiety To Relief & 8.70\% & 0.00\% & 10.87\% & 6.52\% & 10.87\% & 22.50\% & 19.57\% \\
Confusion To Confusion & 2.17\% & 0.00\% & 0.00\% & 0.00\% & 0.00\% & 0.00\% & 0.00\% \\
Confusion To Factual & 2.17\% & 0.00\% & 0.00\% & 0.00\% & 0.00\% & 0.00\% & 0.00\% \\
Confusion To Frustration & 8.70\% & 0.00\% & 0.00\% & 2.17\% & 0.00\% & 0.00\% & 0.00\% \\
Confusion To Gratitude & 6.52\% & 0.00\% & 13.04\% & 15.22\% & 13.04\% & 5.00\% & 10.87\% \\
Confusion To Relief & 6.52\% & 8.33\% & 4.35\% & 15.22\% & 6.52\% & 15.00\% & 4.35\% \\
Curiosity To Anxiety & 4.35\% & 0.00\% & 0.00\% & 0.00\% & 0.00\% & 0.00\% & 0.00\% \\
Curiosity To Gratitude & 2.17\% & 0.00\% & 6.52\% & 2.17\% & 4.35\% & 5.00\% & 2.17\% \\
Curiosity To Relief & 0.00\% & 0.00\% & 0.00\% & 0.00\% & 0.00\% & 7.50\% & 2.17\% \\
Factual To Curiosity & 0.00\% & 0.00\% & 2.17\% & 0.00\% & 0.00\% & 0.00\% & 0.00\% \\
Factual To Factual & 2.17\% & 0.00\% & 2.17\% & 0.00\% & 0.00\% & 0.00\% & 0.00\% \\
Factual To Frustration & 0.00\% & 8.33\% & 0.00\% & 0.00\% & 0.00\% & 0.00\% & 0.00\% \\
Factual To Gratitude & 8.70\% & 0.00\% & 6.52\% & 8.70\% & 10.87\% & 0.00\% & 6.52\% \\
Factual To Relief & 8.70\% & 8.33\% & 2.17\% & 8.70\% & 10.87\% & 0.00\% & 2.17\% \\
Frustration To Anxiety & 2.17\% & 0.00\% & 0.00\% & 0.00\% & 0.00\% & 2.50\% & 0.00\% \\
Frustration To Factual & 4.35\% & 16.67\% & 0.00\% & 2.17\% & 0.00\% & 0.00\% & 0.00\% \\
Frustration To Frustration & 0.00\% & 16.67\% & 0.00\% & 0.00\% & 0.00\% & 0.00\% & 0.00\% \\
Frustration To Gratitude & 15.22\% & 33.33\% & 32.61\% & 17.39\% & 26.09\% & 22.50\% & 21.74\% \\
Frustration To Relief & 17.39\% & 8.33\% & 17.39\% & 19.57\% & 15.22\% & 17.50\% & 30.43\% \\
Gratitude To Relief & 0.00\% & 0.00\% & 0.00\% & 0.00\% & 2.17\% & 0.00\% & 0.00\% \\
\bottomrule
\end{tabular}
\caption{Frequency distribution of \textbf{Customer Emotion Arc} categories for real and synthetic transcripts on the evaluation dataset across generation pipelines (used for G-test and Chi-Square Test) - French. Distributions from the training set of original transcripts are also provided for reference.}
\label{freq_dist_customer_emotion_french}
\end{table*}

\begin{table*}[h!]
\centering
\fontsize{6}{8}\selectfont
\begin{tabular}{lccccccc}
\toprule
\multirow{2}{*}{\textbf{Category}} & \multicolumn{2}{c}{\textbf{Original Transcripts}} & \multicolumn{5}{c}{\textbf{Generation Methods}} \\
\cmidrule(lr){2-3} \cmidrule(lr){4-8}
 & Test Set & Train Set & Direct & Chunked & Characteristic Aware & ConvoGen & NoteChat \\
\midrule
Anger To Frustration & 0.00\% & 0.00\% & 0.00\% & 0.00\% & 2.04\% & 0.00\% & 0.00\% \\
Anxiety To Anxiety & 2.04\% & 0.00\% & 0.00\% & 0.00\% & 0.00\% & 0.00\% & 0.00\% \\
Anxiety To Frustration & 0.00\% & 0.00\% & 2.04\% & 0.00\% & 0.00\% & 2.08\% & 0.00\% \\
Anxiety To Gratitude & 0.00\% & 0.00\% & 8.16\% & 0.00\% & 2.04\% & 4.17\% & 0.00\% \\
Anxiety To Relief & 2.04\% & 0.00\% & 10.20\% & 6.12\% & 10.20\% & 10.42\% & 30.61\% \\
Confusion To Anxiety & 0.00\% & 0.00\% & 0.00\% & 0.00\% & 0.00\% & 6.25\% & 0.00\% \\
Confusion To Confusion & 4.08\% & 0.00\% & 0.00\% & 0.00\% & 0.00\% & 0.00\% & 0.00\% \\
Confusion To Curiosity & 0.00\% & 0.00\% & 0.00\% & 2.04\% & 0.00\% & 0.00\% & 0.00\% \\
Confusion To Factual & 4.08\% & 0.00\% & 8.16\% & 2.04\% & 0.00\% & 0.00\% & 0.00\% \\
Confusion To Frustration & 20.41\% & 70.00\% & 0.00\% & 2.04\% & 8.16\% & 4.17\% & 0.00\% \\
Confusion To Gratitude & 6.12\% & 0.00\% & 18.37\% & 10.20\% & 14.29\% & 14.58\% & 10.20\% \\
Confusion To Relief & 28.57\% & 20.00\% & 18.37\% & 42.86\% & 18.37\% & 35.42\% & 34.69\% \\
Curiosity To Frustration & 0.00\% & 0.00\% & 0.00\% & 0.00\% & 2.04\% & 0.00\% & 0.00\% \\
Curiosity To Gratitude & 0.00\% & 0.00\% & 2.04\% & 2.04\% & 0.00\% & 2.08\% & 0.00\% \\
Curiosity To Relief & 0.00\% & 0.00\% & 2.04\% & 2.04\% & 2.04\% & 0.00\% & 0.00\% \\
Factual To Confusion & 2.04\% & 0.00\% & 0.00\% & 2.04\% & 0.00\% & 0.00\% & 0.00\% \\
Factual To Frustration & 6.12\% & 0.00\% & 2.04\% & 2.04\% & 4.08\% & 0.00\% & 0.00\% \\
Factual To Gratitude & 0.00\% & 0.00\% & 10.20\% & 4.08\% & 4.08\% & 0.00\% & 0.00\% \\
Factual To Relief & 6.12\% & 0.00\% & 0.00\% & 2.04\% & 2.04\% & 0.00\% & 0.00\% \\
Frustration To Anxiety & 0.00\% & 0.00\% & 0.00\% & 2.04\% & 0.00\% & 4.17\% & 0.00\% \\
Frustration To Confusion & 2.04\% & 0.00\% & 0.00\% & 2.04\% & 0.00\% & 0.00\% & 0.00\% \\
Frustration To Factual & 0.00\% & 0.00\% & 0.00\% & 0.00\% & 2.04\% & 0.00\% & 0.00\% \\
Frustration To Frustration & 0.00\% & 0.00\% & 2.04\% & 2.04\% & 2.04\% & 0.00\% & 0.00\% \\
Frustration To Gratitude & 2.04\% & 0.00\% & 12.24\% & 2.04\% & 12.24\% & 8.33\% & 4.08\% \\
Frustration To Relief & 14.29\% & 10.00\% & 4.08\% & 12.24\% & 14.29\% & 8.33\% & 20.41\% \\
\bottomrule
\end{tabular}
\caption{Frequency distribution of \textbf{Customer Emotion Arc} categories for real and synthetic transcripts on the evaluation dataset across generation pipelines (used for G-test and Chi-Square Test) - French-Canadian. Distributions from the training set of original transcripts are also provided for reference.}
\label{freq_dist_customer_emotion_french_canadian}
\end{table*}

\begin{table*}[h!]
\centering
\fontsize{6}{8}\selectfont
\begin{tabular}{lccccccc}
\toprule
\multirow{2}{*}{\textbf{Category}} & \multicolumn{2}{c}{\textbf{Original Transcripts}} & \multicolumn{5}{c}{\textbf{Generation Methods}} \\
\cmidrule(lr){2-3} \cmidrule(lr){4-8}
 & Test Set & Train Set & Direct & Chunked & Characteristic Aware & ConvoGen & NoteChat \\
\midrule
Anxiety To Curiosity & 0.00\% & 0.00\% & 0.00\% & 4.00\% & 0.00\% & 0.00\% & 0.00\% \\
Anxiety To Frustration & 0.00\% & 0.00\% & 0.00\% & 2.00\% & 0.00\% & 0.00\% & 0.00\% \\
Anxiety To Gratitude & 0.00\% & 0.00\% & 4.00\% & 2.00\% & 4.35\% & 3.85\% & 4.00\% \\
Anxiety To Relief & 6.00\% & 25.00\% & 12.00\% & 16.00\% & 13.04\% & 19.23\% & 40.00\% \\
Confusion To Confusion & 2.00\% & 0.00\% & 0.00\% & 0.00\% & 0.00\% & 0.00\% & 0.00\% \\
Confusion To Factual & 2.00\% & 0.00\% & 0.00\% & 0.00\% & 0.00\% & 0.00\% & 0.00\% \\
Confusion To Frustration & 2.00\% & 0.00\% & 0.00\% & 2.00\% & 2.17\% & 0.00\% & 0.00\% \\
Confusion To Gratitude & 2.00\% & 0.00\% & 8.00\% & 8.00\% & 4.35\% & 19.23\% & 10.00\% \\
Confusion To Relief & 6.00\% & 16.67\% & 4.00\% & 6.00\% & 2.17\% & 11.54\% & 6.00\% \\
Curiosity To Confusion & 2.00\% & 0.00\% & 0.00\% & 0.00\% & 0.00\% & 0.00\% & 0.00\% \\
Curiosity To Factual & 4.00\% & 0.00\% & 2.00\% & 0.00\% & 2.17\% & 11.54\% & 0.00\% \\
Curiosity To Gratitude & 16.00\% & 0.00\% & 10.00\% & 18.00\% & 15.22\% & 11.54\% & 16.00\% \\
Curiosity To Relief & 0.00\% & 8.33\% & 2.00\% & 4.00\% & 4.35\% & 0.00\% & 4.00\% \\
Factual To Anxiety & 2.00\% & 0.00\% & 0.00\% & 0.00\% & 0.00\% & 0.00\% & 0.00\% \\
Factual To Confusion & 2.00\% & 0.00\% & 0.00\% & 2.00\% & 0.00\% & 0.00\% & 0.00\% \\
Factual To Curiosity & 6.00\% & 0.00\% & 2.00\% & 0.00\% & 0.00\% & 0.00\% & 0.00\% \\
Factual To Factual & 6.00\% & 8.33\% & 2.00\% & 2.00\% & 2.17\% & 0.00\% & 0.00\% \\
Factual To Frustration & 4.00\% & 16.67\% & 0.00\% & 0.00\% & 0.00\% & 0.00\% & 0.00\% \\
Factual To Gratitude & 20.00\% & 25.00\% & 40.00\% & 22.00\% & 28.26\% & 7.69\% & 6.00\% \\
Factual To Relief & 10.00\% & 0.00\% & 4.00\% & 2.00\% & 2.17\% & 3.85\% & 0.00\% \\
Frustration To Factual & 0.00\% & 0.00\% & 0.00\% & 0.00\% & 2.17\% & 0.00\% & 2.00\% \\
Frustration To Frustration & 0.00\% & 0.00\% & 2.00\% & 0.00\% & 2.17\% & 0.00\% & 0.00\% \\
Frustration To Gratitude & 0.00\% & 0.00\% & 4.00\% & 8.00\% & 6.52\% & 3.85\% & 0.00\% \\
Frustration To Relief & 8.00\% & 0.00\% & 2.00\% & 2.00\% & 4.35\% & 7.69\% & 12.00\% \\
Gratitude To Gratitude & 0.00\% & 0.00\% & 2.00\% & 0.00\% & 2.17\% & 0.00\% & 0.00\% \\
Gratitude To Relief & 0.00\% & 0.00\% & 0.00\% & 0.00\% & 2.17\% & 0.00\% & 0.00\% \\
\bottomrule
\end{tabular}
\caption{Frequency distribution of \textbf{Customer Emotion Arc} categories for real and synthetic transcripts on the evaluation dataset across generation pipelines (used for G-test and Chi-Square Test) - Spanish. Distributions from the training set of original transcripts are also provided for reference.}
\label{freq_dist_customer_emotion_spanish}
\end{table*}

\begin{table*}[h!]
\centering
\fontsize{6}{8}\selectfont
\begin{tabular}{llccccccc}
\toprule
\multirow{2}{*}{\textbf{Language}} & \multirow{2}{*}{\textbf{Category}} & \multicolumn{2}{c}{\textbf{Original Transcripts}} & \multicolumn{5}{c}{\textbf{Generation Methods}} \\
\cmidrule(lr){3-4} \cmidrule(lr){5-9}
 &  & Test Set & Train Set & Direct & Chunked & Characteristic Aware & ConvoGen & NoteChat \\
\midrule
\multirow{17}{*}{\texttt{English}} & Anxiety To Gratitude & 2.13\% & 2.00\% & 0.00\% & 0.00\% & 0.00\% & 0.00\% & 0.00\% \\
 & Confusion To Factual & 4.26\% & 4.97\% & 0.00\% & 0.00\% & 0.00\% & 0.00\% & 0.00\% \\
 & Confusion To Gratitude & 0.00\% & 1.45\% & 0.00\% & 2.13\% & 0.00\% & 0.00\% & 4.26\% \\
 & Curiosity To Factual & 4.26\% & 4.38\% & 2.13\% & 2.13\% & 0.00\% & 6.38\% & 0.00\% \\
 & Curiosity To Gratitude & 6.38\% & 5.44\% & 4.26\% & 2.13\% & 0.00\% & 2.13\% & 4.26\% \\
 & Empathy To Gratitude & 0.00\% & 0.31\% & 2.13\% & 0.00\% & 0.00\% & 0.00\% & 0.00\% \\
 & Factual To Anxiety & 2.13\% & 1.73\% & 0.00\% & 4.26\% & 0.00\% & 0.00\% & 0.00\% \\
 & Factual To Concern & 2.13\% & 2.41\% & 0.00\% & 0.00\% & 0.00\% & 0.00\% & 0.00\% \\
 & Factual To Confusion & 0.00\% & 1.19\% & 0.00\% & 2.13\% & 0.00\% & 2.13\% & 0.00\% \\
 & Factual To Curiosity & 0.00\% & 1.40\% & 0.00\% & 4.26\% & 0.00\% & 0.00\% & 0.00\% \\
 & Factual To Factual & 25.53\% & 20.40\% & 25.53\% & 36.17\% & 31.91\% & 48.94\% & 19.15\% \\
 & Factual To Frustration & 2.13\% & 2.50\% & 0.00\% & 2.13\% & 0.00\% & 0.00\% & 0.00\% \\
 & Factual To Gratitude & 27.66\% & 30.99\% & 63.83\% & 42.55\% & 63.83\% & 38.30\% & 72.34\% \\
 & Factual To Relief & 0.00\% & 0.42\% & 0.00\% & 2.13\% & 2.13\% & 2.13\% & 0.00\% \\
 & Gratitude To Factual & 17.02\% & 14.69\% & 0.00\% & 0.00\% & 2.13\% & 0.00\% & 0.00\% \\
 & Gratitude To Frustration & 2.13\% & 1.84\% & 0.00\% & 0.00\% & 0.00\% & 0.00\% & 0.00\% \\
 & Gratitude To Gratitude & 4.26\% & 3.88\% & 2.13\% & 0.00\% & 0.00\% & 0.00\% & 0.00\% \\
\midrule
\multirow{10}{*}{\texttt{French}} & Confusion To Factual & 0.00\% & 0.00\% & 0.00\% & 0.00\% & 0.00\% & 0.00\% & 2.17\% \\
 & Curiosity To Factual & 2.17\% & 0.00\% & 6.52\% & 4.35\% & 2.17\% & 12.50\% & 0.00\% \\
 & Curiosity To Gratitude & 2.17\% & 0.00\% & 15.22\% & 10.87\% & 4.35\% & 22.50\% & 4.35\% \\
 & Empathy To Gratitude & 0.00\% & 0.00\% & 0.00\% & 0.00\% & 0.00\% & 2.50\% & 0.00\% \\
 & Factual To Confusion & 0.00\% & 8.33\% & 0.00\% & 0.00\% & 0.00\% & 0.00\% & 0.00\% \\
 & Factual To Factual & 34.78\% & 25.00\% & 4.35\% & 26.09\% & 8.70\% & 27.50\% & 13.04\% \\
 & Factual To Gratitude & 56.52\% & 58.33\% & 71.74\% & 56.52\% & 76.09\% & 32.50\% & 80.43\% \\
 & Factual To Relief & 2.17\% & 0.00\% & 0.00\% & 2.17\% & 8.70\% & 2.50\% & 0.00\% \\
 & Frustration To Relief & 2.17\% & 0.00\% & 0.00\% & 0.00\% & 0.00\% & 0.00\% & 0.00\% \\
 & Gratitude To Gratitude & 0.00\% & 8.33\% & 2.17\% & 0.00\% & 0.00\% & 0.00\% & 0.00\% \\
\midrule
\multirow{16}{*}{\texttt{French-Canadian}} & Anxiety To Factual & 0.00\% & 0.00\% & 0.00\% & 2.04\% & 0.00\% & 0.00\% & 0.00\% \\
 & Confusion To Confusion & 2.04\% & 0.00\% & 0.00\% & 0.00\% & 0.00\% & 0.00\% & 0.00\% \\
 & Confusion To Factual & 2.04\% & 0.00\% & 0.00\% & 0.00\% & 0.00\% & 0.00\% & 0.00\% \\
 & Curiosity To Confusion & 2.04\% & 0.00\% & 0.00\% & 0.00\% & 0.00\% & 0.00\% & 0.00\% \\
 & Curiosity To Factual & 10.20\% & 0.00\% & 4.08\% & 4.08\% & 2.04\% & 20.83\% & 0.00\% \\
 & Curiosity To Gratitude & 0.00\% & 0.00\% & 16.33\% & 4.08\% & 2.04\% & 14.58\% & 4.08\% \\
 & Factual To Anxiety & 2.04\% & 0.00\% & 0.00\% & 4.08\% & 0.00\% & 0.00\% & 0.00\% \\
 & Factual To Confusion & 2.04\% & 20.00\% & 0.00\% & 2.04\% & 0.00\% & 0.00\% & 0.00\% \\
 & Factual To Curiosity & 0.00\% & 0.00\% & 0.00\% & 2.04\% & 0.00\% & 0.00\% & 0.00\% \\
 & Factual To Empathy & 0.00\% & 0.00\% & 0.00\% & 2.04\% & 0.00\% & 0.00\% & 0.00\% \\
 & Factual To Factual & 46.94\% & 70.00\% & 18.37\% & 24.49\% & 28.57\% & 41.67\% & 12.24\% \\
 & Factual To Frustration & 8.16\% & 10.00\% & 0.00\% & 0.00\% & 0.00\% & 0.00\% & 0.00\% \\
 & Factual To Gratitude & 24.49\% & 0.00\% & 59.18\% & 53.06\% & 63.27\% & 22.92\% & 83.67\% \\
 & Factual To Relief & 0.00\% & 0.00\% & 2.04\% & 2.04\% & 0.00\% & 0.00\% & 0.00\% \\
 & Gratitude To Factual & 0.00\% & 0.00\% & 0.00\% & 0.00\% & 2.04\% & 0.00\% & 0.00\% \\
 & Gratitude To Gratitude & 0.00\% & 0.00\% & 0.00\% & 0.00\% & 2.04\% & 0.00\% & 0.00\% \\
\midrule
\multirow{13}{*}{\texttt{Spanish}} & Confusion To Factual & 2.00\% & 0.00\% & 0.00\% & 0.00\% & 0.00\% & 0.00\% & 2.00\% \\
 & Confusion To Gratitude & 2.00\% & 0.00\% & 0.00\% & 0.00\% & 0.00\% & 0.00\% & 2.00\% \\
 & Confusion To Relief & 2.00\% & 0.00\% & 0.00\% & 0.00\% & 0.00\% & 0.00\% & 0.00\% \\
 & Curiosity To Factual & 2.00\% & 0.00\% & 0.00\% & 2.00\% & 0.00\% & 3.85\% & 0.00\% \\
 & Curiosity To Frustration & 2.00\% & 0.00\% & 0.00\% & 0.00\% & 0.00\% & 0.00\% & 0.00\% \\
 & Curiosity To Gratitude & 2.00\% & 0.00\% & 6.00\% & 4.00\% & 0.00\% & 3.85\% & 0.00\% \\
 & Factual To Anxiety & 0.00\% & 0.00\% & 0.00\% & 2.00\% & 0.00\% & 0.00\% & 0.00\% \\
 & Factual To Curiosity & 0.00\% & 0.00\% & 0.00\% & 2.00\% & 0.00\% & 0.00\% & 2.00\% \\
 & Factual To Factual & 38.00\% & 33.33\% & 16.00\% & 20.00\% & 19.57\% & 38.46\% & 18.00\% \\
 & Factual To Gratitude & 34.00\% & 50.00\% & 66.00\% & 52.00\% & 56.52\% & 42.31\% & 76.00\% \\
 & Factual To Relief & 2.00\% & 0.00\% & 0.00\% & 0.00\% & 2.17\% & 3.85\% & 0.00\% \\
 & Gratitude To Factual & 6.00\% & 8.33\% & 6.00\% & 2.00\% & 2.17\% & 3.85\% & 0.00\% \\
 & Gratitude To Gratitude & 8.00\% & 8.33\% & 6.00\% & 16.00\% & 19.57\% & 3.85\% & 0.00\% \\
\bottomrule
\end{tabular}
\caption{Frequency distribution of \textbf{Agent Emotion Arc} categories for real and synthetic transcripts on the evaluation dataset across generation pipelines (used for G-test and Chi-Square Test). Distributions from the training set of original transcripts are also provided for reference.}
\label{freq_dist_agent_emotion}
\end{table*}

\begin{table*}[h!]
\centering
\fontsize{6}{8}\selectfont
\begin{tabular}{llccccccc}
\toprule
\multirow{2}{*}{\textbf{Language}} & \multirow{2}{*}{\textbf{Category}} & \multicolumn{2}{c}{\textbf{Original Transcripts}} & \multicolumn{5}{c}{\textbf{Generation Methods}} \\
\cmidrule(lr){3-4} \cmidrule(lr){5-9}
 &  & Test Set & Train Set & Direct & Chunked & Characteristic Aware & ConvoGen & NoteChat \\
\midrule
\multirow{7}{*}{\texttt{English}} & Negative To Negative & 12.77\% & 12.74\% & 4.26\% & 0.00\% & 6.38\% & 4.26\% & 0.00\% \\
 & Negative To Neutral & 6.38\% & 7.91\% & 6.38\% & 6.38\% & 2.13\% & 6.38\% & 2.13\% \\
 & Negative To Positive & 8.51\% & 9.76\% & 12.77\% & 14.89\% & 27.66\% & 40.43\% & 55.32\% \\
 & Neutral To Negative & 21.28\% & 23.23\% & 4.26\% & 10.64\% & 8.51\% & 8.51\% & 0.00\% \\
 & Neutral To Neutral & 12.77\% & 11.56\% & 17.02\% & 10.64\% & 0.00\% & 10.64\% & 10.64\% \\
 & Neutral To Positive & 38.30\% & 34.68\% & 53.19\% & 55.32\% & 53.19\% & 29.79\% & 31.91\% \\
 & Positive To Positive & 0.00\% & 0.12\% & 2.13\% & 2.13\% & 2.13\% & 0.00\% & 0.00\% \\
\midrule
\multirow{7}{*}{\texttt{French}} & Negative To Negative & 13.04\% & 16.67\% & 0.00\% & 2.17\% & 0.00\% & 2.50\% & 0.00\% \\
 & Negative To Neutral & 6.52\% & 16.67\% & 0.00\% & 2.17\% & 0.00\% & 0.00\% & 0.00\% \\
 & Negative To Positive & 54.35\% & 50.00\% & 80.43\% & 76.09\% & 71.74\% & 85.00\% & 86.96\% \\
 & Neutral To Negative & 4.35\% & 8.33\% & 0.00\% & 0.00\% & 0.00\% & 0.00\% & 0.00\% \\
 & Neutral To Neutral & 2.17\% & 0.00\% & 4.35\% & 0.00\% & 0.00\% & 0.00\% & 0.00\% \\
 & Neutral To Positive & 19.57\% & 8.33\% & 15.22\% & 19.57\% & 26.09\% & 12.50\% & 13.04\% \\
 & Positive To Positive & 0.00\% & 0.00\% & 0.00\% & 0.00\% & 2.17\% & 0.00\% & 0.00\% \\
\midrule
\multirow{5}{*}{\texttt{French-Canadian}} & Negative To Negative & 28.57\% & 70.00\% & 4.08\% & 8.16\% & 12.24\% & 16.67\% & 0.00\% \\
 & Negative To Neutral & 4.08\% & 0.00\% & 8.16\% & 4.08\% & 2.04\% & 0.00\% & 0.00\% \\
 & Negative To Positive & 53.06\% & 30.00\% & 71.43\% & 73.47\% & 71.43\% & 81.25\% & 100.00\% \\
 & Neutral To Negative & 8.16\% & 0.00\% & 2.04\% & 4.08\% & 6.12\% & 0.00\% & 0.00\% \\
 & Neutral To Positive & 6.12\% & 0.00\% & 14.29\% & 10.20\% & 8.16\% & 2.08\% & 0.00\% \\
\midrule
\multirow{7}{*}{\texttt{Spanish}} & Negative To Negative & 4.00\% & 0.00\% & 2.00\% & 4.00\% & 4.35\% & 0.00\% & 0.00\% \\
 & Negative To Neutral & 2.00\% & 0.00\% & 0.00\% & 4.00\% & 2.17\% & 0.00\% & 2.00\% \\
 & Negative To Positive & 22.00\% & 41.67\% & 34.00\% & 42.00\% & 34.78\% & 65.38\% & 72.00\% \\
 & Neutral To Negative & 10.00\% & 16.67\% & 0.00\% & 2.00\% & 0.00\% & 0.00\% & 0.00\% \\
 & Neutral To Neutral & 16.00\% & 8.33\% & 6.00\% & 2.00\% & 4.35\% & 11.54\% & 0.00\% \\
 & Neutral To Positive & 46.00\% & 33.33\% & 56.00\% & 46.00\% & 50.00\% & 23.08\% & 26.00\% \\
 & Positive To Positive & 0.00\% & 0.00\% & 2.00\% & 0.00\% & 4.35\% & 0.00\% & 0.00\% \\
\bottomrule
\end{tabular}
\caption{Frequency distribution of \textbf{Customer Sentiment Arc} categories for real and synthetic transcripts on the evaluation dataset across generation pipelines (used for G-test and Chi-Square Test). Distributions from the training set of original transcripts are also provided for reference.}
\label{freq_dist_customer_sentiment}
\end{table*}

\begin{table*}[h!]
\centering
\fontsize{6}{8}\selectfont
\begin{tabular}{llccccccc}
\toprule
\multirow{2}{*}{\textbf{Language}} & \multirow{2}{*}{\textbf{Category}} & \multicolumn{2}{c}{\textbf{Original Transcripts}} & \multicolumn{5}{c}{\textbf{Generation Methods}} \\
\cmidrule(lr){3-4} \cmidrule(lr){5-9}
 &  & Test Set & Train Set & Direct & Chunked & Characteristic Aware & ConvoGen & NoteChat \\
\midrule
\multirow{8}{*}{\texttt{English}} & Negative To Neutral & 4.26\% & 4.28\% & 0.00\% & 0.00\% & 0.00\% & 0.00\% & 0.00\% \\
 & Negative To Positive & 2.13\% & 2.66\% & 0.00\% & 2.13\% & 0.00\% & 0.00\% & 4.26\% \\
 & Neutral To Negative & 4.26\% & 4.93\% & 0.00\% & 8.51\% & 0.00\% & 2.13\% & 0.00\% \\
 & Neutral To Neutral & 31.91\% & 35.11\% & 27.66\% & 42.55\% & 31.91\% & 55.32\% & 19.15\% \\
 & Neutral To Positive & 34.04\% & 31.07\% & 70.21\% & 46.81\% & 65.96\% & 42.55\% & 76.60\% \\
 & Positive To Negative & 2.13\% & 1.94\% & 0.00\% & 0.00\% & 0.00\% & 0.00\% & 0.00\% \\
 & Positive To Neutral & 17.02\% & 14.83\% & 0.00\% & 0.00\% & 2.13\% & 0.00\% & 0.00\% \\
 & Positive To Positive & 4.26\% & 5.17\% & 2.13\% & 0.00\% & 0.00\% & 0.00\% & 0.00\% \\
\midrule
\multirow{6}{*}{\texttt{French}} & Negative To Neutral & 0.00\% & 0.00\% & 0.00\% & 0.00\% & 0.00\% & 0.00\% & 2.17\% \\
 & Negative To Positive & 2.17\% & 0.00\% & 0.00\% & 0.00\% & 0.00\% & 0.00\% & 0.00\% \\
 & Neutral To Negative & 0.00\% & 8.33\% & 0.00\% & 0.00\% & 0.00\% & 0.00\% & 0.00\% \\
 & Neutral To Neutral & 36.96\% & 25.00\% & 10.87\% & 30.43\% & 10.87\% & 40.00\% & 13.04\% \\
 & Neutral To Positive & 60.87\% & 58.33\% & 86.96\% & 69.57\% & 89.13\% & 60.00\% & 84.78\% \\
 & Positive To Positive & 0.00\% & 8.33\% & 2.17\% & 0.00\% & 0.00\% & 0.00\% & 0.00\% \\
\midrule
\multirow{7}{*}{\texttt{French-Canadian}} & Negative To Negative & 2.04\% & 0.00\% & 0.00\% & 0.00\% & 0.00\% & 0.00\% & 0.00\% \\
 & Negative To Neutral & 2.04\% & 0.00\% & 0.00\% & 2.04\% & 0.00\% & 0.00\% & 0.00\% \\
 & Neutral To Negative & 14.29\% & 30.00\% & 0.00\% & 6.12\% & 0.00\% & 0.00\% & 0.00\% \\
 & Neutral To Neutral & 57.14\% & 70.00\% & 22.45\% & 32.65\% & 30.61\% & 62.50\% & 12.24\% \\
 & Neutral To Positive & 24.49\% & 0.00\% & 77.55\% & 59.18\% & 65.31\% & 37.50\% & 87.76\% \\
 & Positive To Neutral & 0.00\% & 0.00\% & 0.00\% & 0.00\% & 2.04\% & 0.00\% & 0.00\% \\
 & Positive To Positive & 0.00\% & 0.00\% & 0.00\% & 0.00\% & 2.04\% & 0.00\% & 0.00\% \\
\midrule
\multirow{7}{*}{\texttt{Spanish}} & Negative To Neutral & 2.00\% & 0.00\% & 0.00\% & 0.00\% & 0.00\% & 0.00\% & 2.00\% \\
 & Negative To Positive & 4.00\% & 0.00\% & 0.00\% & 0.00\% & 0.00\% & 0.00\% & 2.00\% \\
 & Neutral To Negative & 2.00\% & 0.00\% & 0.00\% & 2.00\% & 0.00\% & 0.00\% & 0.00\% \\
 & Neutral To Neutral & 40.00\% & 33.33\% & 16.00\% & 24.00\% & 19.57\% & 42.31\% & 20.00\% \\
 & Neutral To Positive & 38.00\% & 50.00\% & 72.00\% & 56.00\% & 58.70\% & 50.00\% & 76.00\% \\
 & Positive To Neutral & 6.00\% & 8.33\% & 6.00\% & 2.00\% & 2.17\% & 3.85\% & 0.00\% \\
 & Positive To Positive & 8.00\% & 8.33\% & 6.00\% & 16.00\% & 19.57\% & 3.85\% & 0.00\% \\
\bottomrule
\end{tabular}
\caption{Frequency distribution of \textbf{Agent Sentiment Arc} categories for real and synthetic transcripts on the evaluation dataset across generation pipelines (used for G-test and Chi-Square Test). Distributions from the training set of original transcripts are also provided for reference.}
\label{freq_dist_agent_sentiment}
\end{table*}

\begin{table*}[h!]
\centering
\fontsize{6}{8}\selectfont
\begin{tabular}{llccccccc}
\toprule
\multirow{2}{*}{\textbf{Language}} & \multirow{2}{*}{\textbf{Category}} & \multicolumn{2}{c}{\textbf{Original Transcripts}} & \multicolumn{5}{c}{\textbf{Generation Methods}} \\
\cmidrule(lr){3-4} \cmidrule(lr){5-9}
 &  & Test Set & Train Set & Direct & Chunked & Characteristic Aware & ConvoGen & NoteChat \\
\midrule
\multirow{4}{*}{\texttt{English}} & Simple Informal Language & 78.23\% & 76.37\% & 52.29\% & 65.61\% & 56.31\% & 72.89\% & 49.48\% \\
 & Simple Formal Language & 13.96\% & 15.55\% & 40.35\% & 28.59\% & 34.61\% & 22.48\% & 35.23\% \\
 & Complex Informal Language & 6.52\% & 5.91\% & 2.45\% & 3.82\% & 2.41\% & 4.41\% & 9.07\% \\
 & Complex Formal Language & 1.29\% & 2.17\% & 4.91\% & 1.98\% & 6.67\% & 0.22\% & 6.22\% \\
\midrule
\multirow{4}{*}{\texttt{French}} & Simple Informal Language & 77.52\% & 79.71\% & 51.27\% & 68.56\% & 54.02\% & 59.82\% & 34.91\% \\
 & Simple Formal Language & 17.77\% & 17.39\% & 44.10\% & 29.18\% & 40.02\% & 38.19\% & 58.65\% \\
 & Complex Informal Language & 3.59\% & 2.50\% & 0.87\% & 1.49\% & 0.80\% & 1.43\% & 2.62\% \\
 & Complex Formal Language & 1.12\% & 0.40\% & 3.76\% & 0.77\% & 5.15\% & 0.55\% & 3.82\% \\
\midrule
\multirow{4}{*}{\texttt{French-Canadian}} & Simple Informal Language & 70.88\% & 71.73\% & 48.96\% & 70.09\% & 55.34\% & 54.64\% & 34.22\% \\
 & Simple Formal Language & 24.08\% & 21.49\% & 48.71\% & 28.16\% & 38.60\% & 44.74\% & 60.76\% \\
 & Complex Informal Language & 4.74\% & 5.40\% & 0.61\% & 1.21\% & 0.80\% & 0.63\% & 1.06\% \\
 & Complex Formal Language & 0.29\% & 1.38\% & 1.72\% & 0.53\% & 5.26\% & 0.00\% & 3.97\% \\
\midrule
\multirow{4}{*}{\texttt{Spanish}} & Simple Informal Language & 73.83\% & 75.24\% & 48.08\% & 61.89\% & 53.23\% & 51.12\% & 36.79\% \\
 & Simple Formal Language & 19.84\% & 16.48\% & 46.52\% & 33.38\% & 39.46\% & 44.60\% & 52.85\% \\
 & Complex Informal Language & 4.66\% & 6.69\% & 0.82\% & 2.34\% & 1.80\% & 2.85\% & 2.97\% \\
 & Complex Formal Language & 1.67\% & 1.59\% & 4.58\% & 2.39\% & 5.51\% & 1.43\% & 7.39\% \\
\bottomrule
\end{tabular}
\caption{Frequency distribution of \textbf{Language Complexity} categories for real and synthetic transcripts on the evaluation dataset across generation pipelines (used for G-test and Chi-Square Test). Distributions from the training set of original transcripts are also provided for reference.}
\label{freq_dist_language_complexity}
\end{table*}

\begin{table*}[h!]
\centering
\fontsize{6}{8}\selectfont
\begin{tabular}{llccccccc}
\toprule
\multirow{2}{*}{\textbf{Language}} & \multirow{2}{*}{\textbf{Category}} & \multicolumn{2}{c}{\textbf{Original Transcripts}} & \multicolumn{5}{c}{\textbf{Generation Methods}} \\
\cmidrule(lr){3-4} \cmidrule(lr){5-9}
 &  & Test Set & Train Set & Direct & Chunked & Characteristic Aware & ConvoGen & NoteChat \\
\midrule
\multirow{5}{*}{\texttt{English}} & 1 & 0.00\% & 0.00\% & 0.00\% & 0.00\% & 0.00\% & 0.00\% & 0.00\% \\
 & 2 & 6.00\% & 2.00\% & 0.00\% & 2.00\% & 6.00\% & 0.00\% & 0.00\% \\
 & 3 & 30.00\% & 48.00\% & 46.00\% & 58.00\% & 60.00\% & 24.00\% & 50.00\% \\
 & 4 & 60.00\% & 46.00\% & 48.00\% & 38.00\% & 34.00\% & 72.00\% & 48.00\% \\
 & 5 & 4.00\% & 4.00\% & 6.00\% & 2.00\% & 0.00\% & 4.00\% & 2.00\% \\
\midrule
\multirow{5}{*}{\texttt{French}} & 1 & 0.00\% & 0.00\% & 0.00\% & 0.00\% & 0.00\% & 0.00\% & 0.00\% \\
 & 2 & 18.00\% & 20.00\% & 8.00\% & 2.00\% & 10.00\% & 2.44\% & 8.00\% \\
 & 3 & 50.00\% & 62.00\% & 48.00\% & 72.00\% & 62.00\% & 31.71\% & 60.00\% \\
 & 4 & 32.00\% & 18.00\% & 44.00\% & 24.00\% & 26.00\% & 63.41\% & 30.00\% \\
 & 5 & 0.00\% & 0.00\% & 0.00\% & 2.00\% & 2.00\% & 2.44\% & 2.00\% \\
\midrule
\multirow{5}{*}{\texttt{French-Canadian}} & 1 & 0.00\% & 1.00\% & 0.00\% & 0.00\% & 0.00\% & 0.00\% & 0.00\% \\
 & 2 & 8.00\% & 2.00\% & 0.00\% & 0.00\% & 0.00\% & 0.00\% & 2.00\% \\
 & 3 & 10.00\% & 25.00\% & 12.00\% & 22.00\% & 32.00\% & 12.24\% & 12.00\% \\
 & 4 & 78.00\% & 71.00\% & 70.00\% & 62.00\% & 60.00\% & 63.27\% & 66.00\% \\
 & 5 & 4.00\% & 1.00\% & 18.00\% & 16.00\% & 8.00\% & 24.49\% & 20.00\% \\
\midrule
\multirow{5}{*}{\texttt{Spanish}} & 1 & 0.00\% & 0.00\% & 0.00\% & 0.00\% & 0.00\% & 0.00\% & 0.00\% \\
 & 2 & 2.00\% & 2.15\% & 0.00\% & 6.00\% & 6.00\% & 3.85\% & 2.00\% \\
 & 3 & 26.00\% & 25.81\% & 32.00\% & 38.00\% & 44.00\% & 19.23\% & 24.00\% \\
 & 4 & 72.00\% & 66.67\% & 54.00\% & 52.00\% & 44.00\% & 65.38\% & 72.00\% \\
 & 5 & 0.00\% & 5.38\% & 14.00\% & 4.00\% & 6.00\% & 11.54\% & 2.00\% \\
\bottomrule
\end{tabular}
\caption{Frequency distribution of \textbf{Technical Density$^{\dagger}$} categories for real and synthetic transcripts on the evaluation dataset across generation pipelines (used for G-test and Chi-Square Test). Distributions from the training set of original transcripts are also provided for reference. $^{\dagger}$ denotes transcript-level metric.}
\label{freq_dist_technical_density}
\end{table*}

\begin{table*}[h!]
\centering
\fontsize{6}{8}\selectfont
\begin{tabular}{llccccccc}
\toprule
\multirow{2}{*}{\textbf{Language}} & \multirow{2}{*}{\textbf{Category}} & \multicolumn{2}{c}{\textbf{Original Transcripts}} & \multicolumn{5}{c}{\textbf{Generation Methods}} \\
\cmidrule(lr){3-4} \cmidrule(lr){5-9}
 &  & Test Set & Train Set & Direct & Chunked & Characteristic Aware & ConvoGen & NoteChat \\
\midrule
\multirow{5}{*}{\texttt{English}} & 1 & 0.00\% & 0.00\% & 0.00\% & 0.00\% & 0.00\% & 0.00\% & 0.00\% \\
 & 2 & 2.00\% & 0.00\% & 0.00\% & 0.00\% & 2.00\% & 0.00\% & 0.00\% \\
 & 3 & 16.00\% & 14.00\% & 0.00\% & 0.00\% & 8.00\% & 0.00\% & 4.00\% \\
 & 4 & 76.00\% & 84.00\% & 70.00\% & 94.00\% & 88.00\% & 94.00\% & 96.00\% \\
 & 5 & 6.00\% & 2.00\% & 30.00\% & 6.00\% & 2.00\% & 6.00\% & 0.00\% \\
\midrule
\multirow{5}{*}{\texttt{French}} & 1 & 0.00\% & 0.00\% & 0.00\% & 0.00\% & 0.00\% & 0.00\% & 0.00\% \\
 & 2 & 8.00\% & 25.00\% & 0.00\% & 0.00\% & 0.00\% & 0.00\% & 0.00\% \\
 & 3 & 4.00\% & 6.00\% & 0.00\% & 0.00\% & 4.00\% & 0.00\% & 0.00\% \\
 & 4 & 88.00\% & 69.00\% & 82.00\% & 98.00\% & 96.00\% & 87.80\% & 100.00\% \\
 & 5 & 0.00\% & 0.00\% & 18.00\% & 2.00\% & 0.00\% & 12.20\% & 0.00\% \\
\midrule
\multirow{5}{*}{\texttt{French-Canadian}} & 1 & 0.00\% & 0.00\% & 0.00\% & 0.00\% & 0.00\% & 0.00\% & 0.00\% \\
 & 2 & 2.00\% & 8.00\% & 0.00\% & 0.00\% & 2.00\% & 0.00\% & 0.00\% \\
 & 3 & 22.00\% & 16.00\% & 0.00\% & 0.00\% & 8.00\% & 0.00\% & 0.00\% \\
 & 4 & 74.00\% & 76.00\% & 82.00\% & 98.00\% & 90.00\% & 93.88\% & 100.00\% \\
 & 5 & 2.00\% & 0.00\% & 18.00\% & 2.00\% & 0.00\% & 6.12\% & 0.00\% \\
\midrule
\multirow{5}{*}{\texttt{Spanish}} & 1 & 0.00\% & 0.00\% & 0.00\% & 0.00\% & 0.00\% & 0.00\% & 0.00\% \\
 & 2 & 0.00\% & 1.08\% & 0.00\% & 0.00\% & 4.00\% & 0.00\% & 0.00\% \\
 & 3 & 16.00\% & 31.18\% & 0.00\% & 2.00\% & 2.00\% & 0.00\% & 2.00\% \\
 & 4 & 82.00\% & 67.74\% & 70.00\% & 96.00\% & 92.00\% & 84.62\% & 98.00\% \\
 & 5 & 2.00\% & 0.00\% & 30.00\% & 2.00\% & 2.00\% & 15.38\% & 0.00\% \\
\bottomrule
\end{tabular}
\caption{Frequency distribution of \textbf{Sentence Complexity$^{\dagger}$} categories for real and synthetic transcripts on the evaluation dataset across generation pipelines (used for G-test and Chi-Square Test). Distributions from the training set of original transcripts are also provided for reference. $^{\dagger}$ denotes transcript-level metric.}
\label{freq_dist_sentence_complexity}
\end{table*}

\begin{table*}[h!]
\centering
\fontsize{6}{8}\selectfont
\begin{tabular}{llccccccc}
\toprule
\multirow{2}{*}{\textbf{Language}} & \multirow{2}{*}{\textbf{Category}} & \multicolumn{2}{c}{\textbf{Original Transcripts}} & \multicolumn{5}{c}{\textbf{Generation Methods}} \\
\cmidrule(lr){3-4} \cmidrule(lr){5-9}
 &  & Test Set & Train Set & Direct & Chunked & Characteristic Aware & ConvoGen & NoteChat \\
\midrule
\multirow{5}{*}{\texttt{English}} & 1 & 0.00\% & 1.00\% & 0.00\% & 0.00\% & 0.00\% & 0.00\% & 0.00\% \\
 & 2 & 2.00\% & 2.00\% & 0.00\% & 0.00\% & 0.00\% & 0.00\% & 0.00\% \\
 & 3 & 20.00\% & 16.00\% & 0.00\% & 2.00\% & 6.00\% & 0.00\% & 0.00\% \\
 & 4 & 72.00\% & 79.00\% & 74.00\% & 90.00\% & 88.00\% & 90.00\% & 96.00\% \\
 & 5 & 6.00\% & 2.00\% & 26.00\% & 8.00\% & 6.00\% & 10.00\% & 4.00\% \\
\midrule
\multirow{5}{*}{\texttt{French}} & 1 & 0.00\% & 0.00\% & 0.00\% & 0.00\% & 0.00\% & 0.00\% & 0.00\% \\
 & 2 & 16.00\% & 28.00\% & 0.00\% & 0.00\% & 0.00\% & 0.00\% & 0.00\% \\
 & 3 & 26.00\% & 34.00\% & 0.00\% & 2.00\% & 4.00\% & 0.00\% & 0.00\% \\
 & 4 & 58.00\% & 38.00\% & 92.00\% & 96.00\% & 96.00\% & 100.00\% & 100.00\% \\
 & 5 & 0.00\% & 0.00\% & 8.00\% & 2.00\% & 0.00\% & 0.00\% & 0.00\% \\
\midrule
\multirow{5}{*}{\texttt{French-Canadian}} & 1 & 4.00\% & 8.00\% & 0.00\% & 0.00\% & 0.00\% & 0.00\% & 0.00\% \\
 & 2 & 44.00\% & 40.00\% & 0.00\% & 0.00\% & 0.00\% & 0.00\% & 0.00\% \\
 & 3 & 44.00\% & 36.00\% & 2.00\% & 2.00\% & 6.00\% & 0.00\% & 2.00\% \\
 & 4 & 8.00\% & 16.00\% & 84.00\% & 96.00\% & 90.00\% & 95.92\% & 96.00\% \\
 & 5 & 0.00\% & 0.00\% & 14.00\% & 2.00\% & 4.00\% & 4.08\% & 2.00\% \\
\midrule
\multirow{5}{*}{\texttt{Spanish}} & 1 & 2.00\% & 3.23\% & 0.00\% & 0.00\% & 0.00\% & 0.00\% & 0.00\% \\
 & 2 & 2.00\% & 1.08\% & 0.00\% & 0.00\% & 2.00\% & 0.00\% & 0.00\% \\
 & 3 & 34.00\% & 24.73\% & 0.00\% & 8.00\% & 2.00\% & 3.85\% & 2.00\% \\
 & 4 & 62.00\% & 70.97\% & 68.00\% & 88.00\% & 94.00\% & 76.92\% & 98.00\% \\
 & 5 & 0.00\% & 0.00\% & 32.00\% & 4.00\% & 2.00\% & 19.23\% & 0.00\% \\
\bottomrule
\end{tabular}
\caption{Frequency distribution of \textbf{Overall Readability Score$^{\dagger}$} categories for real and synthetic transcripts on the evaluation dataset across generation pipelines (used for G-test and Chi-Square Test). Distributions from the training set of original transcripts are also provided for reference. $^{\dagger}$ denotes transcript-level metric.}
\label{freq_dist_overall_readability_score}
\end{table*}

\begin{table*}[h!]
\centering
\fontsize{6}{8}\selectfont
\begin{tabular}{llccccccc}
\toprule
\multirow{2}{*}{\textbf{Language}} & \multirow{2}{*}{\textbf{Category}} & \multicolumn{2}{c}{\textbf{Original Transcripts}} & \multicolumn{5}{c}{\textbf{Generation Methods}} \\
\cmidrule(lr){3-4} \cmidrule(lr){5-9}
 &  & Test Set & Train Set & Direct & Chunked & Characteristic Aware & ConvoGen & NoteChat \\
\midrule
\multirow{5}{*}{\texttt{English}} & 1 & 2.00\% & 2.00\% & 0.00\% & 0.00\% & 0.00\% & 0.00\% & 0.00\% \\
 & 2 & 6.00\% & 6.00\% & 0.00\% & 0.00\% & 0.00\% & 0.00\% & 0.00\% \\
 & 3 & 42.00\% & 33.00\% & 0.00\% & 18.00\% & 4.00\% & 26.00\% & 12.00\% \\
 & 4 & 40.00\% & 42.00\% & 0.00\% & 36.00\% & 22.00\% & 40.00\% & 22.00\% \\
 & 5 & 10.00\% & 17.00\% & 100.00\% & 46.00\% & 74.00\% & 34.00\% & 66.00\% \\
\midrule
\multirow{5}{*}{\texttt{French}} & 1 & 0.00\% & 5.00\% & 0.00\% & 0.00\% & 0.00\% & 0.00\% & 0.00\% \\
 & 2 & 20.00\% & 25.00\% & 0.00\% & 0.00\% & 0.00\% & 0.00\% & 0.00\% \\
 & 3 & 16.00\% & 23.00\% & 0.00\% & 20.00\% & 0.00\% & 7.32\% & 4.00\% \\
 & 4 & 44.00\% & 42.00\% & 0.00\% & 58.00\% & 10.00\% & 14.63\% & 16.00\% \\
 & 5 & 20.00\% & 5.00\% & 100.00\% & 22.00\% & 90.00\% & 78.05\% & 80.00\% \\
\midrule
\multirow{5}{*}{\texttt{French-Canadian}} & 1 & 6.00\% & 12.00\% & 0.00\% & 0.00\% & 0.00\% & 0.00\% & 0.00\% \\
 & 2 & 66.00\% & 55.00\% & 2.00\% & 2.00\% & 0.00\% & 0.00\% & 0.00\% \\
 & 3 & 18.00\% & 19.00\% & 6.00\% & 62.00\% & 18.00\% & 22.45\% & 10.00\% \\
 & 4 & 10.00\% & 14.00\% & 10.00\% & 26.00\% & 28.00\% & 44.90\% & 34.00\% \\
 & 5 & 0.00\% & 0.00\% & 82.00\% & 10.00\% & 54.00\% & 32.65\% & 56.00\% \\
\midrule
\multirow{5}{*}{\texttt{Spanish}} & 1 & 2.00\% & 3.23\% & 0.00\% & 0.00\% & 0.00\% & 0.00\% & 0.00\% \\
 & 2 & 12.00\% & 7.53\% & 0.00\% & 0.00\% & 0.00\% & 0.00\% & 0.00\% \\
 & 3 & 48.00\% & 41.94\% & 2.00\% & 24.00\% & 8.00\% & 7.69\% & 18.00\% \\
 & 4 & 32.00\% & 35.48\% & 0.00\% & 46.00\% & 28.00\% & 19.23\% & 18.00\% \\
 & 5 & 6.00\% & 11.83\% & 98.00\% & 30.00\% & 64.00\% & 73.08\% & 64.00\% \\
\bottomrule
\end{tabular}
\caption{Frequency distribution of \textbf{Discourse Flow$^{\dagger}$} categories for real and synthetic transcripts on the evaluation dataset across generation pipelines (used for G-test and Chi-Square Test). Distributions from the training set of original transcripts are also provided for reference. $^{\dagger}$ denotes transcript-level metric.}
\label{freq_dist_discourse_flow}
\end{table*}

\begin{table*}[h!]
\centering
\fontsize{6}{8}\selectfont
\begin{tabular}{llccccccc}
\toprule
\multirow{2}{*}{\textbf{Language}} & \multirow{2}{*}{\textbf{Category}} & \multicolumn{2}{c}{\textbf{Original Transcripts}} & \multicolumn{5}{c}{\textbf{Generation Methods}} \\
\cmidrule(lr){3-4} \cmidrule(lr){5-9}
 &  & Test Set & Train Set & Direct & Chunked & Characteristic Aware & ConvoGen & NoteChat \\
\midrule
\multirow{3}{*}{\texttt{English}} & Neutral Proactivity & 70.04\% & 72.37\% & 90.57\% & 80.43\% & 83.32\% & 84.50\% & 77.68\% \\
 & Overstated Proactivity & 7.25\% & 5.45\% & 4.95\% & 6.50\% & 9.73\% & 6.56\% & 9.02\% \\
 & Understated Proactivity & 22.71\% & 22.18\% & 4.48\% & 13.07\% & 6.95\% & 8.94\% & 13.30\% \\
\midrule
\multirow{3}{*}{\texttt{French}} & Neutral Proactivity & 76.19\% & 79.99\% & 92.97\% & 81.63\% & 86.90\% & 89.22\% & 84.92\% \\
 & Overstated Proactivity & 1.51\% & 0.58\% & 3.51\% & 3.72\% & 6.76\% & 3.21\% & 7.09\% \\
 & Understated Proactivity & 22.30\% & 19.43\% & 3.51\% & 14.65\% & 6.35\% & 7.57\% & 7.99\% \\
\midrule
\multirow{3}{*}{\texttt{French-Canadian}} & Neutral Proactivity & 65.00\% & 68.16\% & 88.64\% & 78.25\% & 84.77\% & 84.24\% & 79.67\% \\
 & Overstated Proactivity & 1.47\% & 1.41\% & 3.03\% & 4.07\% & 4.80\% & 4.62\% & 10.16\% \\
 & Understated Proactivity & 33.52\% & 30.42\% & 8.33\% & 17.68\% & 10.44\% & 11.14\% & 10.16\% \\
\midrule
\multirow{3}{*}{\texttt{Spanish}} & Neutral Proactivity & 75.14\% & 72.87\% & 90.43\% & 82.81\% & 84.27\% & 86.53\% & 83.06\% \\
 & Overstated Proactivity & 2.53\% & 3.85\% & 6.21\% & 5.23\% & 6.88\% & 2.45\% & 8.47\% \\
 & Understated Proactivity & 22.33\% & 23.28\% & 3.37\% & 11.97\% & 8.85\% & 11.02\% & 8.47\% \\
\bottomrule
\end{tabular}
\caption{Frequency distribution of \textbf{Proactivity} categories for real and synthetic transcripts on the evaluation dataset across generation pipelines (used for G-test and Chi-Square Test). Distributions from the training set of original transcripts are also provided for reference.}
\label{freq_dist_proactivity}
\end{table*}

\begin{table*}[h!]
\centering
\fontsize{6}{8}\selectfont
\begin{tabular}{llccccccc}
\toprule
\multirow{2}{*}{\textbf{Language}} & \multirow{2}{*}{\textbf{Category}} & \multicolumn{2}{c}{\textbf{Original Transcripts}} & \multicolumn{5}{c}{\textbf{Generation Methods}} \\
\cmidrule(lr){3-4} \cmidrule(lr){5-9}
 &  & Test Set & Train Set & Direct & Chunked & Characteristic Aware & ConvoGen & NoteChat \\
\midrule
\multirow{3}{*}{\texttt{English}} & Emotion Focused & 11.14\% & 9.81\% & 13.88\% & 11.97\% & 13.74\% & 16.82\% & 11.67\% \\
 & Fact Focused & 83.94\% & 85.42\% & 68.73\% & 70.00\% & 67.85\% & 53.74\% & 76.56\% \\
 & Balanced & 4.93\% & 4.77\% & 17.39\% & 18.02\% & 18.42\% & 29.44\% & 11.76\% \\
\midrule
\multirow{3}{*}{\texttt{French}} & Emotion Focused & 8.46\% & 6.28\% & 15.51\% & 12.73\% & 14.31\% & 17.87\% & 10.53\% \\
 & Fact Focused & 87.73\% & 91.42\% & 66.76\% & 70.05\% & 68.98\% & 57.16\% & 82.40\% \\
 & Balanced & 3.81\% & 2.30\% & 17.73\% & 17.21\% & 16.71\% & 24.97\% & 7.07\% \\
\midrule
\multirow{3}{*}{\texttt{French-Canadian}} & Emotion Focused & 16.47\% & 14.79\% & 22.97\% & 18.41\% & 20.43\% & 22.49\% & 13.33\% \\
 & Fact Focused & 77.63\% & 81.40\% & 49.31\% & 57.54\% & 58.66\% & 44.85\% & 73.44\% \\
 & Balanced & 5.90\% & 3.81\% & 27.72\% & 24.05\% & 20.91\% & 32.66\% & 13.23\% \\
\midrule
\multirow{3}{*}{\texttt{Spanish}} & Emotion Focused & 8.53\% & 10.88\% & 14.49\% & 13.69\% & 13.39\% & 14.32\% & 11.67\% \\
 & Fact Focused & 87.25\% & 84.15\% & 64.58\% & 67.12\% & 66.94\% & 56.83\% & 77.67\% \\
 & Balanced & 4.22\% & 4.98\% & 20.93\% & 19.19\% & 19.67\% & 28.85\% & 10.66\% \\
\bottomrule
\end{tabular}
\caption{Frequency distribution of \textbf{Emphasis} categories for real and synthetic transcripts on the evaluation dataset across generation pipelines (used for G-test and Chi-Square Test). Distributions from the training set of original transcripts are also provided for reference.}
\label{freq_dist_emphasis}
\end{table*}

\begin{table*}[h!]
\centering
\fontsize{6}{8}\selectfont
\begin{tabular}{llccccccc}
\toprule
\multirow{2}{*}{\textbf{Language}} & \multirow{2}{*}{\textbf{Category}} & \multicolumn{2}{c}{\textbf{Original Transcripts}} & \multicolumn{5}{c}{\textbf{Generation Methods}} \\
\cmidrule(lr){3-4} \cmidrule(lr){5-9}
 &  & Test Set & Train Set & Direct & Chunked & Characteristic Aware & ConvoGen & NoteChat \\
\midrule
\multirow{4}{*}{\texttt{English}} & No Question & 61.55\% & 61.62\% & 67.98\% & 68.54\% & 74.52\% & 69.54\% & 42.16\% \\
 & Closed Question & 15.87\% & 16.54\% & 16.75\% & 13.02\% & 11.36\% & 12.23\% & 27.64\% \\
 & Informational Question & 18.37\% & 18.65\% & 9.95\% & 14.40\% & 9.81\% & 12.31\% & 21.37\% \\
 & Conversational Question & 4.21\% & 3.19\% & 5.32\% & 4.04\% & 4.31\% & 5.92\% & 8.83\% \\
\midrule
\multirow{4}{*}{\texttt{French}} & No Question & 74.57\% & 72.99\% & 69.64\% & 67.84\% & 72.83\% & 69.88\% & 44.74\% \\
 & Closed Question & 9.47\% & 10.52\% & 13.74\% & 13.70\% & 10.12\% & 12.00\% & 22.85\% \\
 & Informational Question & 13.63\% & 13.14\% & 11.72\% & 15.19\% & 13.46\% & 12.82\% & 24.89\% \\
 & Conversational Question & 2.33\% & 3.35\% & 4.90\% & 3.27\% & 3.59\% & 5.29\% & 7.51\% \\
\midrule
\multirow{4}{*}{\texttt{French-Canadian}} & No Question & 67.49\% & 67.72\% & 74.05\% & 75.21\% & 78.48\% & 70.82\% & 41.97\% \\
 & Closed Question & 8.67\% & 9.10\% & 12.58\% & 9.38\% & 8.66\% & 9.37\% & 19.29\% \\
 & Informational Question & 19.08\% & 20.37\% & 8.39\% & 12.11\% & 8.41\% & 13.25\% & 26.07\% \\
 & Conversational Question & 4.76\% & 2.81\% & 4.98\% & 3.30\% & 4.45\% & 6.56\% & 12.67\% \\
\midrule
\multirow{4}{*}{\texttt{Spanish}} & No Question & 60.58\% & 64.43\% & 65.92\% & 55.20\% & 63.59\% & 60.18\% & 31.72\% \\
 & Closed Question & 11.15\% & 10.49\% & 13.69\% & 15.81\% & 13.35\% & 14.22\% & 24.61\% \\
 & Informational Question & 24.00\% & 20.51\% & 14.66\% & 22.43\% & 18.41\% & 19.47\% & 32.15\% \\
 & Conversational Question & 4.27\% & 4.57\% & 5.73\% & 6.55\% & 4.65\% & 6.13\% & 11.52\% \\
\bottomrule
\end{tabular}
\caption{Frequency distribution of \textbf{Question Type} categories for real and synthetic transcripts on the evaluation dataset across generation pipelines (used for G-test and Chi-Square Test). Distributions from the training set of original transcripts are also provided for reference.}
\label{freq_dist_question_type}
\end{table*}

\begin{table*}[h!]
\centering
\fontsize{6}{8}\selectfont
\begin{tabular}{llccccccc}
\toprule
\multirow{2}{*}{\textbf{Language}} & \multirow{2}{*}{\textbf{Category}} & \multicolumn{2}{c}{\textbf{Original Transcripts}} & \multicolumn{5}{c}{\textbf{Generation Methods}} \\
\cmidrule(lr){3-4} \cmidrule(lr){5-9}
 &  & Test Set & Train Set & Direct & Chunked & Characteristic Aware & ConvoGen & NoteChat \\
\midrule
\multirow{3}{*}{\texttt{English}} & Solution Oriented & 18.89\% & 19.13\% & 32.68\% & 34.10\% & 34.47\% & 36.42\% & 34.11\% \\
 & Process Oriented & 40.53\% & 39.88\% & 43.22\% & 41.53\% & 42.58\% & 47.98\% & 41.41\% \\
 & No Solution & 40.59\% & 40.99\% & 24.10\% & 24.38\% & 22.95\% & 15.61\% & 24.48\% \\
\midrule
\multirow{3}{*}{\texttt{French}} & Solution Oriented & 23.82\% & 22.70\% & 35.32\% & 30.69\% & 36.27\% & 36.90\% & 33.33\% \\
 & Process Oriented & 33.62\% & 35.00\% & 45.39\% & 43.83\% & 45.19\% & 48.06\% & 42.85\% \\
 & No Solution & 42.56\% & 42.30\% & 19.28\% & 25.48\% & 18.54\% & 15.03\% & 23.82\% \\
\midrule
\multirow{3}{*}{\texttt{French-Canadian}} & Solution Oriented & 21.67\% & 21.02\% & 31.50\% & 27.34\% & 31.93\% & 32.70\% & 31.77\% \\
 & Process Oriented & 37.59\% & 37.21\% & 46.78\% & 45.23\% & 42.72\% & 53.24\% & 38.91\% \\
 & No Solution & 40.74\% & 41.77\% & 21.72\% & 27.44\% & 25.35\% & 14.05\% & 29.32\% \\
\midrule
\multirow{3}{*}{\texttt{Spanish}} & Solution Oriented & 15.22\% & 19.31\% & 26.53\% & 28.74\% & 30.16\% & 35.37\% & 26.88\% \\
 & Process Oriented & 43.48\% & 40.25\% & 51.66\% & 45.97\% & 49.59\% & 48.78\% & 45.99\% \\
 & No Solution & 41.30\% & 40.44\% & 21.82\% & 25.29\% & 20.25\% & 15.85\% & 27.14\% \\
\bottomrule
\end{tabular}
\caption{Frequency distribution of \textbf{Solution} categories for real and synthetic transcripts on the evaluation dataset across generation pipelines (used for G-test and Chi-Square Test). Distributions from the training set of original transcripts are also provided for reference.}
\label{freq_dist_solution}
\end{table*}

\begin{table*}[h!]
\centering
\fontsize{6}{8}\selectfont
\begin{tabular}{llccccccc}
\toprule
\multirow{2}{*}{\textbf{Language}} & \multirow{2}{*}{\textbf{Category}} & \multicolumn{2}{c}{\textbf{Original Transcripts}} & \multicolumn{5}{c}{\textbf{Generation Methods}} \\
\cmidrule(lr){3-4} \cmidrule(lr){5-9}
 &  & Test Set & Train Set & Direct & Chunked & Characteristic Aware & ConvoGen & NoteChat \\
\midrule
\multirow{5}{*}{\texttt{English}} & Speech Repair Repetition & 7.74\% & 7.25\% & 0.47\% & 4.87\% & 2.91\% & 1.10\% & 5.06\% \\
 & Hesitation Fillers & 4.84\% & 4.91\% & 2.21\% & 14.29\% & 4.79\% & 74.23\% & 22.69\% \\
 & Interactional Disfluency & 27.00\% & 28.20\% & 0.63\% & 1.88\% & 3.07\% & 0.51\% & 6.64\% \\
 & Comprehension Clarity Issues & 3.09\% & 3.38\% & 2.13\% & 5.74\% & 3.34\% & 2.42\% & 1.19\% \\
 & No Disfluency & 57.32\% & 56.26\% & 94.56\% & 73.22\% & 85.89\% & 21.73\% & 64.43\% \\
\midrule
\multirow{5}{*}{\texttt{French}} & Speech Repair Repetition & 6.84\% & 6.71\% & 0.26\% & 7.50\% & 4.40\% & 1.42\% & 3.70\% \\
 & Hesitation Fillers & 4.74\% & 5.02\% & 2.46\% & 15.53\% & 5.51\% & 57.95\% & 19.45\% \\
 & Interactional Disfluency & 26.14\% & 25.31\% & 0.26\% & 2.14\% & 2.33\% & 0.44\% & 4.35\% \\
 & Comprehension Clarity Issues & 2.60\% & 2.94\% & 2.11\% & 7.06\% & 3.34\% & 1.63\% & 1.02\% \\
 & No Disfluency & 59.68\% & 60.01\% & 94.91\% & 67.76\% & 84.43\% & 38.56\% & 71.48\% \\
\midrule
\multirow{5}{*}{\texttt{French-Canadian}} & Speech Repair Repetition & 9.20\% & 8.53\% & 2.17\% & 8.00\% & 4.21\% & 2.67\% & 6.66\% \\
 & Hesitation Fillers & 9.36\% & 7.44\% & 4.81\% & 19.98\% & 6.34\% & 60.15\% & 19.33\% \\
 & Interactional Disfluency & 24.16\% & 26.05\% & 1.44\% & 3.12\% & 3.85\% & 0.49\% & 5.92\% \\
 & Comprehension Clarity Issues & 8.34\% & 6.83\% & 6.14\% & 8.30\% & 6.20\% & 4.13\% & 1.81\% \\
 & No Disfluency & 48.94\% & 51.15\% & 85.44\% & 60.59\% & 79.40\% & 32.56\% & 66.28\% \\
\midrule
\multirow{5}{*}{\texttt{Spanish}} & Speech Repair Repetition & 9.08\% & 9.24\% & 0.46\% & 7.56\% & 5.15\% & 1.40\% & 5.14\% \\
 & Hesitation Fillers & 6.50\% & 6.83\% & 2.47\% & 16.89\% & 6.40\% & 50.50\% & 21.36\% \\
 & Interactional Disfluency & 16.32\% & 19.31\% & 0.27\% & 1.43\% & 3.96\% & 2.40\% & 6.09\% \\
 & Comprehension Clarity Issues & 4.85\% & 3.98\% & 1.65\% & 5.88\% & 4.17\% & 2.00\% & 1.03\% \\
 & No Disfluency & 63.25\% & 60.64\% & 95.16\% & 68.24\% & 80.33\% & 43.71\% & 66.38\% \\
\bottomrule
\end{tabular}
\caption{Frequency distribution of \textbf{Disfluency} categories for real and synthetic transcripts on the evaluation dataset across generation pipelines (used for G-test and Chi-Square Test). Distributions from the training set of original transcripts are also provided for reference.}
\label{freq_dist_disfluency}
\end{table*}

\begin{table*}[h!]
\centering
\fontsize{6}{8}\selectfont
\begin{tabular}{llccccccc}
\toprule
\multirow{2}{*}{\textbf{Language}} & \multirow{2}{*}{\textbf{Category}} & \multicolumn{2}{c}{\textbf{Original Transcripts}} & \multicolumn{5}{c}{\textbf{Generation Methods}} \\
\cmidrule(lr){3-4} \cmidrule(lr){5-9}
 &  & Test Set & Train Set & Direct & Chunked & Characteristic Aware & ConvoGen & NoteChat \\
\midrule
\multirow{3}{*}{\texttt{English}} & No Repetition & 79.70\% & 80.76\% & 84.85\% & 80.13\% & 81.96\% & 78.52\% & 82.67\% \\
 & Agent Repetition & 10.93\% & 10.64\% & 9.87\% & 12.42\% & 11.66\% & 12.45\% & 17.33\% \\
 & Customer Repetition & 9.36\% & 8.60\% & 5.29\% & 7.46\% & 6.38\% & 9.02\% & 0.00\% \\
\midrule
\multirow{3}{*}{\texttt{French}} & No Repetition & 82.39\% & 82.24\% & 84.47\% & 76.89\% & 82.51\% & 83.22\% & 89.95\% \\
 & Agent Repetition & 8.95\% & 8.90\% & 8.46\% & 13.18\% & 10.14\% & 7.89\% & 10.05\% \\
 & Customer Repetition & 8.66\% & 8.86\% & 7.07\% & 9.93\% & 7.35\% & 8.88\% & 0.00\% \\
\midrule
\multirow{3}{*}{\texttt{French-Canadian}} & No Repetition & 77.06\% & 76.71\% & 78.55\% & 72.30\% & 77.97\% & 73.88\% & 86.15\% \\
 & Agent Repetition & 11.78\% & 11.62\% & 10.78\% & 15.45\% & 11.74\% & 14.25\% & 13.85\% \\
 & Customer Repetition & 11.15\% & 11.67\% & 10.66\% & 12.25\% & 10.28\% & 11.88\% & 0.00\% \\
\midrule
\multirow{3}{*}{\texttt{Spanish}} & No Repetition & 79.42\% & 77.14\% & 87.93\% & 81.19\% & 82.71\% & 83.37\% & 86.58\% \\
 & Agent Repetition & 10.44\% & 11.99\% & 7.95\% & 11.54\% & 10.41\% & 10.34\% & 13.42\% \\
 & Customer Repetition & 10.14\% & 10.88\% & 4.11\% & 7.26\% & 6.88\% & 6.29\% & 0.00\% \\
\bottomrule
\end{tabular}
\caption{Frequency distribution of \textbf{Repetition} categories for real and synthetic transcripts on the evaluation dataset across generation pipelines (used for G-test and Chi-Square Test). Distributions from the training set of original transcripts are also provided for reference.}
\label{freq_dist_repetition}
\end{table*}

\begin{table*}[h!]
\centering
\fontsize{6}{8}\selectfont
\begin{tabular}{llccccccc}
\toprule
\multirow{2}{*}{\textbf{Language}} & \multirow{2}{*}{\textbf{Category}} & \multicolumn{2}{c}{\textbf{Original Transcripts}} & \multicolumn{5}{c}{\textbf{Generation Methods}} \\
\cmidrule(lr){3-4} \cmidrule(lr){5-9}
 &  & Test Set & Train Set & Direct & Chunked & Characteristic Aware & ConvoGen & NoteChat \\
\midrule
\multirow{4}{*}{\texttt{English}} & No Noise & 72.47\% & 68.91\% & 100.00\% & 94.56\% & 96.79\% & 67.88\% & 85.16\% \\
 & Substitution & 6.51\% & 8.12\% & 0.00\% & 1.12\% & 1.03\% & 1.04\% & 3.12\% \\
 & Insertion & 3.14\% & 2.66\% & 0.00\% & 3.47\% & 0.65\% & 30.56\% & 10.96\% \\
 & Deletion & 17.87\% & 20.30\% & 0.00\% & 0.86\% & 1.52\% & 0.52\% & 0.76\% \\
\midrule
\multirow{4}{*}{\texttt{French}} & No Noise & 67.53\% & 71.56\% & 99.65\% & 97.17\% & 97.64\% & 92.44\% & 95.73\% \\
 & Substitution & 5.03\% & 4.30\% & 0.35\% & 0.65\% & 0.64\% & 0.55\% & 1.10\% \\
 & Insertion & 1.57\% & 1.43\% & 0.00\% & 1.57\% & 1.07\% & 6.90\% & 2.43\% \\
 & Deletion & 25.86\% & 22.71\% & 0.00\% & 0.61\% & 0.64\% & 0.11\% & 0.74\% \\
\midrule
\multirow{4}{*}{\texttt{French-Canadian}} & No Noise & 40.14\% & 43.66\% & 97.87\% & 96.74\% & 96.29\% & 85.25\% & 91.45\% \\
 & Substitution & 28.26\% & 24.24\% & 0.00\% & 1.21\% & 1.26\% & 0.38\% & 1.28\% \\
 & Insertion & 1.73\% & 1.42\% & 2.13\% & 1.73\% & 1.41\% & 14.37\% & 6.84\% \\
 & Deletion & 29.86\% & 30.68\% & 0.00\% & 0.32\% & 1.04\% & 0.00\% & 0.43\% \\
\midrule
\multirow{4}{*}{\texttt{Spanish}} & No Noise & 72.87\% & 72.42\% & 99.73\% & 94.94\% & 96.31\% & 91.48\% & 94.56\% \\
 & Substitution & 7.59\% & 8.34\% & 0.27\% & 1.25\% & 0.67\% & 2.23\% & 0.74\% \\
 & Insertion & 2.97\% & 3.45\% & 0.00\% & 3.42\% & 1.12\% & 6.29\% & 3.79\% \\
 & Deletion & 16.57\% & 15.79\% & 0.00\% & 0.39\% & 1.90\% & 0.00\% & 0.91\% \\
\bottomrule
\end{tabular}
\caption{Frequency distribution of \textbf{ASR Noise Type} categories for real and synthetic transcripts on the evaluation dataset across generation pipelines (used for G-test and Chi-Square Test). Distributions from the training set of original transcripts are also provided for reference.}
\label{freq_dist_asr_noise_type}
\end{table*}

\begin{table*}[h!]
\centering
  \fontsize{8}{9}\selectfont
\begin{tabular}{lcccccccc}
\toprule
\multirow{2}{*}{Method} &
  \multicolumn{2}{c}{Proactivity} &
  \multicolumn{2}{c}{Emphasis} &
  \multicolumn{2}{c}{Question Type} &
  \multicolumn{2}{c}{Solution} \\
\cmidrule(lr){2-3} \cmidrule(lr){4-5} \cmidrule(lr){6-7} \cmidrule(lr){8-9}
&
\multicolumn{1}{c}{$\chi^2$/G (p)} &
\multicolumn{1}{c}{JS-Div} &
\multicolumn{1}{c}{$\chi^2$/G (p)} &
\multicolumn{1}{c}{JS-Div} &
\multicolumn{1}{c}{$\chi^2$/G (p)} &
\multicolumn{1}{c}{JS-Div} &
\multicolumn{1}{c}{$\chi^2$/G (p)} &
\multicolumn{1}{c}{JS-Div} \\
\midrule
  Direct  & 0.000 & 0.021 & 0.000 & 0.017 & 0.000 & 0.006 & 0.000 & 0.018  \\
  Chunked  & 0.000 & 0.003 & 0.000 & 0.015 & 0.000 & 0.001 & 0.000 & 0.014  \\
  Characteristic Aware  & 0.000 & 0.014 & 0.000 & 0.042 & 0.000 & 0.063 & 0.000 & 0.013  \\
  ConvoGen  & 0.000 & 0.012 & 0.000 & 0.033 & 0.000 & 0.016 & 0.000 & 0.094  \\
  NoteChat  & 0.000 & 0.098 & 0.000 & 0.023 & 0.000 & 0.052 & 0.000 & 0.026  \\
 \bottomrule
\end{tabular}%
\caption{Comparison of methods for transcript generation across \textbf{English} language and metrics in the \textbf{Interaction Style and Operational} category on the expanded set of 200 English examples.}
\label{results_interaction_style_english_200}
\end{table*}

\begin{table*}[t]
  \centering
  \fontsize{8}{9}\selectfont
\begin{tabular}{lcccccc}
\toprule
\multirow{2}{*}{Method} &
  \multicolumn{2}{c}{Disfluency} &
  \multicolumn{2}{c}{Repetition} &
  \multicolumn{2}{c}{ASR Noise} \\
\cmidrule(lr){2-3} \cmidrule(lr){4-5} \cmidrule(lr){6-7}
&
{$\chi^2$/G (p)} & JS-Div &
{$\chi^2$/G (p)} & JS-Div &
{$\chi^2$/G (p)} & JS-Div \\
\midrule
Direct  & 0.000 & 0.252 & 0.000 & 0.131 & 0.000 & 0.063 \\
Chunked & 0.000 & 0.032 & 0.014 & 0.002 & 0.000 & 0.031 \\
Characteristic Aware & 0.000 & 0.023 & 0.000 & 0.001 & 0.000 & 0.014 \\
ConvoGen & 0.000 & 0.515 & \textcolor{magenta}{\textbf{0.126}} & 0.002 & 0.002 & 0.101 \\
NoteChat & 0.000 & 0.052 & 0.023 & 0.006 & 0.000 & 0.026 \\
\bottomrule
\end{tabular}
\caption{Comparison of methods for transcript generation across \textbf{English} language and metrics in the \textbf{Conversational Properties} category on the expanded set of 200 English examples.}
\label{results_conversational_properties_english_200}
\end{table*}

%============

\begin{table*}[ht]
  \centering
\resizebox{0.9\textwidth}{!}{%
\large
\begin{tabular}{lcccccccccc}
\toprule
\multirow{2}{*}{Method} &
  \multicolumn{2}{c}{Sentiment} &
  \multicolumn{2}{c}{Customer Emotion Arc$^{\dagger}$} &
  \multicolumn{2}{c}{Agent Emotion Arc$^{\dagger}$} &
  \multicolumn{2}{c}{Customer Sentiment Arc$^{\dagger}$} &
  \multicolumn{2}{c}{Agent Sentiment Arc$^{\dagger}$} \\
\cmidrule(lr){2-3} \cmidrule(lr){4-5} \cmidrule(lr){6-7}
\cmidrule(lr){8-9} \cmidrule(lr){10-11}
&
\multicolumn{1}{c}{$\chi^2$/G (p)} &
\multicolumn{1}{c}{JS-Div} &
\multicolumn{1}{c}{$\chi^2$/G (p)} &
\multicolumn{1}{c}{JS-Div} &
\multicolumn{1}{c}{$\chi^2$/G (p)} &
\multicolumn{1}{c}{JS-Div} &
\multicolumn{1}{c}{$\chi^2$/G (p)} &
\multicolumn{1}{c}{JS-Div} &
\multicolumn{1}{c}{$\chi^2$/G (p)} &
\multicolumn{1}{c}{JS-Div} \\
\midrule
  Direct  & 0.000 & 0.013 & \textcolor{magenta}{\textbf{0.152}} & 0.102 & 0.012 & 0.051 & \textcolor{magenta}{\textbf{0.251}} & 0.002 & 0.012 & 0.032  \\
  Chunked  & 0.000 & 0.010 & \textcolor{magenta}{\textbf{0.255}} & 0.053 & \textcolor{magenta}{\textbf{0.072}} & 0.002 & \textcolor{magenta}{\textbf{0.063}} & 0.026 & 0.044 & 0.238  \\
  Characteristic Aware  & 0.000 & 0.012 & 0.025 & 0.132 & 0.005 & 0.133 & 0.012 & 0.207 & 0.022 & 0.219  \\
  ConvoGen  & 0.000 & 0.012 & 0.000 & 0.530 & 0.017 & 0.109 & 0.021 & 0.062 & 0.003 & 0.323  \\
  NoteChat  & 0.000 & 0.021 & 0.000 & 0.590 & 0.003 & 0.510 & 0.000 & 0.421 & 0.000 & 0.244  \\
 \bottomrule
\end{tabular}%
}
\caption{Comparison of methods for transcript generation across \textbf{English} language and metrics in the \textbf{Sentiment and Emotion} category on the expanded set of 200 English examples. $^{\dagger}$ denotes transcript-level metrics; unmarked metrics are turn-level traits.}
\label{results_sent_and_emotion_english_200}
\end{table*}

\begin{table*}[ht]
  \centering
\resizebox{0.9\textwidth}{!}{%
\large
\begin{tabular}{lcccccccccc}
\toprule
\multirow{2}{*}{Method} &
  \multicolumn{2}{c}{Language Complexity} &
  \multicolumn{2}{c}{Technical Density$^{\dagger}$} &
  \multicolumn{2}{c}{Sentence Complexity$^{\dagger}$} &
  \multicolumn{2}{c}{Overall Readability$^{\dagger}$} &
  \multicolumn{2}{c}{Discourse Flow$^{\dagger}$} \\
\cmidrule(lr){2-3} \cmidrule(lr){4-5} \cmidrule(lr){6-7}
\cmidrule(lr){8-9} \cmidrule(lr){10-11}
&
\multicolumn{1}{c}{$\chi^2$/G (p)} &
\multicolumn{1}{c}{JS-Div} &
\multicolumn{1}{c}{$\chi^2$/G (p)} &
\multicolumn{1}{c}{JS-Div} &
\multicolumn{1}{c}{$\chi^2$/G (p)} &
\multicolumn{1}{c}{JS-Div} &
\multicolumn{1}{c}{$\chi^2$/G (p)} &
\multicolumn{1}{c}{JS-Div} &
\multicolumn{1}{c}{$\chi^2$/G (p)} &
\multicolumn{1}{c}{JS-Div} \\
\midrule
  Direct  & 0.000 & 0.117 & \textcolor{magenta}{\textbf{0.236}} & 0.012 & 0.002 & 0.073 & 0.004 & 0.063 & 0.000 & 0.156  \\
  Chunked  & 0.000 & 0.038 & 0.031 & 0.098 & 0.023 & 0.044 & 0.018 & 0.095 & 0.000 & 0.205  \\
  Characteristic Aware  & 0.000 & 0.283 & 0.022 & 0.016 & \textcolor{magenta}{\textbf{0.339}} & 0.055 & \textcolor{magenta}{\textbf{0.087}} & 0.095 & 0.000 & 0.311  \\
  ConvoGen  & 0.000 & 0.005 & \textcolor{magenta}{\textbf{0.149}} & 0.067 & 0.029 & 0.032 & 0.010 & 0.019 & 0.008 & 0.017  \\
  NoteChat  & 0.000 & 0.044 & \textcolor{magenta}{\textbf{0.088}} & 0.026 & 0.023 & 0.155 & 0.002 & 0.007 & 0.000 & 0.651  \\
 \bottomrule
\end{tabular}%
}
\caption{Comparison of methods for transcript generation across \textbf{English} language and metrics in the \textbf{Linguistic Complexity and Content Density} category  on the expanded set of 200 English examples. $^{\dagger}$ denotes transcript-level metrics; unmarked metrics are turn-level.}
\label{results_linguistic_complexity_english_200}
\end{table*}

% ============================ Compact Implementation Details Table ============================
\begin{table*}
  \centering
  \small
  \setlength{\tabcolsep}{3pt}
  \begin{tabular}{llc}
    \hline
    \textbf{Category} & \textbf{Metric} & \textbf{$\kappa$} \\
    \hline
    \textbf{Sentiment and Emotion} & Customer Emotion Arc (T) & 0.667 \\
     & Customer Sentiment Arc (T) & 0.812 \\
     & Agent Emotion Arc (T) & 0.750 \\
     & Agent Sentiment Arc (T) & 0.883 \\
     & Sentiment (Turn) & 0.880 \\
    \hline
    \textbf{Linguistic Complexity and Content Density} & Vocabulary Complexity (T) & 0.739 \\
     & Sentence Complexity (T) & 0.733 \\
     & Technical Density (T) & 0.651 \\
     & Discourse Flow (T) & 0.632 \\
     & Overall Readability Score (T) & 0.665 \\
     & Language Complexity (Turn) & 0.668 \\
    \hline
    \textbf{Interaction Style and Operational} & Proactivity (Turn) & 0.610 \\
     & Emphasis (Turn) & 0.636 \\
     & Question Type (Turn) & 0.808 \\
     & Solution (Turn) & 0.657 \\
    \hline
    \textbf{Conversational Properties} & Disfluency$^*$ (Turn) & 0.612 \\
     & Repetition (Turn) & 0.607 \\
     & ASR Noise$^*$ (Turn) & 0.670 \\
    \hline
  \end{tabular}
  \caption{Inter-annotator agreement measured by Cohen's Kappa ($\kappa$) for transcript-level (T) and turn-level (Turn) metrics. Values closer to 1.0 indicate higher agreement. $^*$Multi-select metrics (Disfluency, ASR Noise) use macro-averaged $\kappa$ scores calculated individually for each label.}
  \label{tab:kappa_scores}
\end{table*}

\begin{table*}[!hbt]
\centering
\small
\setlength\tabcolsep{6pt}
\renewcommand{\arraystretch}{1.1}
\begin{tabular}{llcccc}
\toprule
\textbf{Stage} & \textbf{Model} & \textbf{Provider} & \textbf{Temp.} & \textbf{Top-p} & \textbf{Max Tokens} \\
\midrule

\multicolumn{6}{l}{\textbf{Synthetic Transcript Generation}} \\
Convogen & GPT-4.1-mini & OpenAI & 0.7 & 1.0 & 32,768 \\
Notechat & GPT-4.1-mini & OpenAI & 0.7 & 1.0 & 32,768 \\
Chunked Enhancement & GPT-4.1-mini & OpenAI & 0.7 & 1.0 & 32,768 \\
Char-Aware Enhancement & GPT-4.1-mini & OpenAI & 0.7 & 1.0 & 32,768 \\

\midrule
\multicolumn{6}{l}{\textbf{Prompt Optimization (Generation \& Downstream)}} \\
Task Model\textsuperscript{*} & \parbox{4cm}{GPT-4.1-mini, GPT-4.1, GPT-4o, Claude-v3-Haiku} & \parbox{2cm}{OpenAI, Anthropic} & 0.7 & 1.0 & 32,768 \\
Prompt Model & Claude 3.5 Sonnet & Anthropic & 1.0 & 1.0 & 8192 \\

\midrule
\multicolumn{6}{l}{\textbf{Generation Evaluation}} \\
Reconstruction & Claude 3.5 Sonnet & Anthropic & 1.0 & 1.0 & 8192 \\

\midrule
\multicolumn{6}{l}{\textbf{Downstream Task (AutoQA)}} \\
Answer Generation\textsuperscript{*} & \parbox{4cm}{GPT-4.1-mini, GPT-4.1, GPT-4o, Claude-v3-Haiku} & \parbox{2cm}{OpenAI, Anthropic} & 0.7 & 1.0 & 1024 \\

\bottomrule
\end{tabular}
\caption{LLM configurations used across pipeline stages. \textsuperscript{*}Task models and answer generation use a suite of four diverse models (GPT-4.1-mini, GPT-4.1, GPT-4o, Claude-v3-Haiku) evaluated separately and averaged for model-agnostic assessment.}
\label{tab:llm_pipeline_parameters}
\end{table*}

\begin{table*}[t]
  \centering
  \small
  \setlength{\tabcolsep}{4pt}
  \renewcommand{\arraystretch}{1.3}
  \begin{tabular}{lp{2.2cm}p{2.2cm}p{7cm}}
    \hline
    \textbf{Hyperparameter} &
    \textbf{Synthetic Transcript Generation} &
    \textbf{AutoQA Downstream Task} &
    \textbf{Description} \\
    \hline
    \texttt{max\_bootstrapped\_demos} & 1 & 0 & Number of bootstrapped demonstrations generated per program. \\
    \texttt{max\_labeled\_demos}      & 0 & 3 & Number of labeled (few-shot) demonstrations allowed. \\
    \texttt{num\_candidates}         & 5 & 5 & Number of prompt candidates proposed per optimization step. \\
    \texttt{num\_threads}            & 4 & 10 & Number of parallel threads used during optimization. \\
    \texttt{num\_trials}             & 10 & 50 & Number of optimization trials executed. \\
    \hline
  \end{tabular}
  \caption{\label{tab:dspy_hyperparameters}
  DSPy \texttt{MIPROv2} optimization hyperparameters used for (i) synthetic transcript generation and (ii) downstream AutoQA prompt optimization.}
  
\end{table*}

\begin{table*}[!hbt]
\centering
\small
\begin{tabular}{p{12cm}cc}
\toprule
\textbf{Question} & \textbf{Yes (\%)} & \textbf{No (\%)} \\
\midrule
\multicolumn{3}{l}{\textit{Test Set}} \\
\midrule
Did the agent offer a proper greeting? & 88.2 & 11.8 \\
Did the agent properly close the call? & 85.7 & 14.3 \\
Did the agent acknowledge and determine the reason for the call & 89.8 & 10.2 \\
Did the agent complete all necessary transactions required on the account? & 69.4 & 30.6 \\
Did the agent probe effectively to accurately diagnose and confirm the issue & 84.4 & 15.6 \\
Did the agent take control of the call/re-phrase to check for customer's understanding without using leading questions? & 94.9 & 5.1 \\
Did the agent actively listen to the customer to avoid repetition & 67.9 & 32.1 \\
\midrule
\multicolumn{3}{l}{\textit{Train Set}} \\
\midrule
Did the agent avoid interruption or talking over the customer & 88.2 & 11.8 \\
Did the agent actively listen to the caller and ask the proper probing questions? & 90.9 & 9.1 \\
Did the agent open the call in a positive and optimistic tone and make sure to offer assistance & 90.9 & 9.1 \\
Did the agent use the customers name at least once during the call? & 88.9 & 11.1 \\
\bottomrule
\end{tabular}
\caption{Question types and answer distribution in the AutoQA test and train sets.}
\label{tab:question_distribution}
\end{table*}

\begin{table*}[t]
  \centering
  \small
  \renewcommand{\arraystretch}{1.2}
  \setlength{\tabcolsep}{8pt}
  \begin{tabular}{lcccc}
    \hline
    \textbf{Language} & \textbf{Very Short} & \textbf{Short} & \textbf{Medium} & \textbf{Long} \\
    \hline
    English          & 65.07  & 188.50 & 421.73 & 495.54 \\
    French           & 99.42  & 111.13 & 304.00 & 558.75 \\
    French Canadian  & 66.87  & 173.33 & 283.00 & 637.00 \\
    Spanish          & 79.61  & 173.67 & 201.17 & 385.00 \\
    \hline
  \end{tabular}
  \caption{\label{tab:call_length_turn_distribution}
    Mean number of turns in the tuning data of real transcripts for different call length categories across languages. Call lengths are grouped into four bins: \textit{Very Short}, \textit{Short}, \textit{Medium}, and \textit{Long}.
  }
\end{table*}

\clearpage
\twocolumn
\subsection{What Do Structured Inputs Mean and How are They Obtained?}
\label{desc_struc_inputs}

To address challenges in synthetic contact center transcript generation and limitations of existing methods, our pipeline conditions generation on modular, interpretable supervision signals derived from real transcripts and routinely produced in call center operations, such as summarization \citep{probing-cc-llm}, QA auto-answering \citep{probing-qa}, and call segmentation \citep{calls-enhancing-segmentation}. We focus on four supervision types commonly available in structured call attributes: intent-specific summaries, topic flows, and QA scores as question–response pairs, providing consistent semantic and behavioral guidance across datasets. We also inject ASR noise and disfluencies, which are characteristic of real-world transcripts but typically absent in clean, LLM-generated outputs.

\begin{enumerate}[leftmargin=5.25mm, itemsep=0.5mm]
    \item \textbf{Intent-Specific Summaries} capture the semantic backbone of a conversation, such as complaints, key events, or resolutions. They ensure inclusion of core content and act as anchors that prevent hallucinations.

    \item \textbf{Topic Flow} provides a global discourse plan, such as a progression from greeting to complaint to troubleshooting and resolution. This supports coherent turn transitions, enhances discourse structure, and models speaker role changes over time.

    \item \textbf{Quality assurance (QA)} forms supply structured behavioral annotations for each call. In routine QA processes, every interaction is evaluated against questions such as “Did the agent demonstrate empathy with the customer?” or “Did the agent propose a solution without prompting?” These scores capture how agents actually perform across dimensions like empathy, proactivity, and script adherence. Because different agents often exhibit distinct behaviors even when handling the same call intent, we use these QA-derived labels to induce behavioral variation in generation. Conditioning on these annotations enables our pipeline to produce a diverse set of synthetic transcripts that faithfully reflect the range of real-world agent interactions. 

\end{enumerate}

These components collectively support the generation of synthetic data that is faithful to the structural, stylistic, and ASR characteristics of real contact center conversations. Examples of these attributes are shown in Tables~\ref{examples_dual_stage_v1_1} - \ref{examples_dual_stage_v1_2}, and detailed descriptions are provided in section~\ref{desc_struc_inputs}.

\begin{enumerate}[leftmargin=5.25mm, itemsep=0.5mm]
    \item \textbf{Intent-Specific Summaries}: We sample a set of carefully chosen intents and obtain their summaries using in-house systems \citep{probing-cc-llm} capable of automatically generating intent-specific summaries. The details of these systems are beyond the scope of this work. The selected intents are diverse enough to collectively capture key characteristics of contact center conversations while minimizing overlap. These intents include:

    \begin{enumerate}
        \item \textit{Customer Complaints}: Summary of any complaints raised by the customer, if any.

        \item \textit{Key Events}: Summary of key events that occurred during the call.

        \item \textit{Next Steps}: Summary of the next steps or actions agreed upon, if any.

        \item \textit{Reason for Call}: Summary of the primary reason the customer initiated the call.

        \item \textit{Key Entities}: Summary of the key entities mentioned in the call.

        \item \textit{Hold and Transfer}: Summary related to holds or transfers that occurred.

        \item \textit{Resolution}: Summary of the call's resolution or outcome, if any.
    \end{enumerate}

    \item \textbf{Topic Flow}: We obtain the ordered sequence of meaningful topics using in-house systems capable of automatically segmenting calls into topic sequences. Each identified topic is accompanied by a name and a corresponding description. The details of these internal systems are beyond the scope of this work.

    \item \textbf{QA Evaluation}: The questions are sourced from call evaluation forms, each designed to assess various aspects of agent performance during interactions. The answers are obtained from in-house systems \citep{probing-qa}, the details of which are beyond the scope of this work. For both the generation (prompt tuning) and evaluation datasets, we first sampled a set of real contact center interactions and used their mapped evaluation forms to retrieve all applicable QA question-answer pairs for that call. 
    
    Call evaluation forms are structured assessment templates used by Quality Assurance (QA) teams in contact centers to systematically evaluate agent performance during customer interactions. These forms are designed by QA experts and supervisors, typically based on business rules, compliance requirements, and customer service quality standards. Each form contains a set of predefined questions, often grouped by dimensions such as professionalism, problem resolution, empathy, adherence to script, escalation handling, and closing behavior. Each question is paired with a limited set of answer choices—usually in the form of Likert scales, binary options (Yes/No), or categorical ratings.
    
    In operational settings, every completed interaction is either randomly sampled or algorithmically selected for quality review, and the corresponding call (or interaction) is mapped to one of the evaluation forms. Trained evaluators manually or semi-automatically fill out the answers based on call recordings or transcripts, which are then used for agent training, performance analytics, and compliance tracking.
    
    These evaluation forms are also increasingly leveraged in AI workflows to both benchmark and steer generation. In our work, they serve as supervision scaffolds: for each sampled interaction in our generation and evaluation datasets, we retrieve its associated evaluation form and extract all question-answer pairs to guide synthetic transcript generation. This process ensures that key behavioral and functional traits - such as adherence to resolution procedures or maintenance of a professional tone - are embedded into the generated dialogue. While in-house systems \citep{probing-qa} can automatically generate answers to QA questions, detailing these systems is beyond our scope; our focus is on leveraging these attributes to produce synthetic transcripts without relying on the original transcripts.

\end{enumerate}

Examples of the formatted input attributes used can be found in Tables \ref{examples_dual_stage_v1_1} - \ref{examples_dual_stage_v1_2}.

\subsection{Characteristic-Aware Enhancement Pipeline}
\label{characteristic_aware_pipeline}

This advanced pipeline generates synthetic transcripts \( \mathcal{T}_S \) whose turn-level linguistic features match those of a real-world corpus \( \mathcal{T}_R \), by systematically controlling the distribution of conversational characteristics.

\subsubsection*{Stage 1: Characteristic-Aware Base Generation}

A modified generation model \( G'_{\text{base}} \) is conditioned on both standard call attributes and sampled transcript-level characteristics (e.g., \texttt{vocabulary\_complexity}, \texttt{customer\_emotion}) to produce a base transcript \( T'_{\text{base}} \) with the desired global tone and structure.

\subsubsection*{Stage 2: Chunking and Conversational Extension}

The transcript \( T'_{\text{base}} \) is segmented into coherent chunks  
\( \mathcal{C}_{\text{chunk}} = (\chi_1, \chi_2, \ldots, \chi_k) \), and each chunk undergoes conversational extension. This increases interactivity by adding natural back-and-forth exchanges and breaking down long monologues.

\subsubsection*{Stage 3: Controlled Application of Turn-Level Characteristics}

A set of turn-level features are applied in three steps:

\begin{enumerate}
    \item \textbf{Candidate Identification:}  
    For each characteristic dimension \( C_d \in \mathcal{C} \) (e.g., Sentiment, Question Type), an LLM identifies eligible turns:
    \[
    \text{Cand}(c_{d,j}) \subseteq \text{Turns}, \quad \forall c_{d,j} \in C_d
    \]

    \item \textbf{Global Probabilistic Sampling:}  
    Let \( P_{\text{target}} \) denote the desired distribution of characteristics derived from real data. For each \( c_{d,j} \in C_d \), sample a set of turns \( \mathcal{U}_S(c_{d,j}) \subseteq \text{Cand}(c_{d,j}) \) such that:
    \[
    \frac{|\mathcal{U}_S(c_{d,j})|}{|\mathcal{U}_S|} \approx P_{\text{target}}(c_{d,j}) \quad \forall c_{d,j} \in C_d, \forall C_d \in \mathcal{C}
    \]

    \item \textbf{Targeted Application:}  
    Each selected turn in \( \mathcal{U}_S(c_{d,j}) \) is modified to exhibit the characteristic \( c_{d,j} \), while preserving context. Unselected turns remain unchanged.
\end{enumerate}

\subsubsection*{Stage 4: Recombination}

The modified chunks are recombined to yield the final synthetic transcript:
\[
T_{\text{final}} = \chi'_1 \oplus \chi'_2 \oplus \ldots \oplus \chi'_k
\]

This characteristic-aware pipeline provides an explicit mechanism for generating synthetic data with verifiable linguistic properties aligned with a target dataset. It is a powerful tool for constructing high-fidelity training and evaluation corpora for robust speech and language models.

\subsection{Reconstruction Score}
\label{loss_function_prompt_tuning}

To systematically measure how accurately synthetic transcripts reflect the explicit call attributes provided as input during generation, we introduce a comprehensive evaluation framework centered on reconstruction accuracy and conversational realism. Unlike traditional surface-level comparisons, this framework evaluates structural, semantic, and stylistic fidelity to assess whether the generated conversation fulfills the intended goals and content. Reconstruction accuracy focuses on how well the transcript adheres to provided specifications—such as key events, topic sequence, and summarized intents—while conversational realism captures the naturalness of speech patterns. These dimensions are quantified using automated LLM-based evaluations, orchestrated via the \texttt{dspy} framework \citep{khattab2023dspycompilingdeclarativelanguage}, and combined into a single, interpretable \textbf{Reconstruction Score}. This score serves as a grounded, attribute-aware measure of generation quality.

\subsubsection{Topic Flow Adherence}

This metric evaluates whether the synthetic transcript follows a predefined narrative structure. It is calculated by having an LLM assess the synthetic transcript to confirm the presence and correct sequencing of all specified topics, assigning a raw score, \(S_{TS_{raw}}\), on a scale of 1 to 10. A perfect score indicates that all topics are covered in the correct order with logical transitions.

\subsubsection{Intent Summary Fulfillment}

This metric verifies that the synthetic transcript contains the specific information detailed in various intent summaries. To calculate the score, we distinguish between primary and secondary intents. The adherence to ``Key Events'' yields a raw score, \(S_{KE_{raw}}\). The adherence to other intents (e.g., \textit{next steps}, \textit{resolution}) are also scored from 1 to 10, and their scores are averaged to produce an \textit{Average Score per Summary Intent}, \(S_{Summ\_Intent_{raw}}\). The evaluation penalizes both missing information and the inclusion of extraneous details. The prompt used for calculating this metric can be found in Table \ref{prompts_for_transcript_generation_3}.

\subsubsection{QA Scenario Replication}

This metric ensures the synthetic transcript is factually consistent from an external reviewer's perspective by measuring its alignment with a set of pre-defined Quality Assurance (QA) question-answer pairs. It is calculated by having an LLM assign a binary score (1 for a match, 0 for a mismatch) for each QA pair based on the synthetic transcript content. These individual scores are then averaged to produce the final QA score, \(S_{QA}\), which ranges from 0 to 1. The final QA score can be interpreted as the proportion of QA questions for which there's alignment with respect to the synthetic transcript. The prompt used for calculating this metric can be found in Table \ref{prompts_for_transcript_generation_3}.

\subsubsection{Conversational Realism Metrics}

A transcript can be factually correct but sound artificial. The \textbf{Conversational Realism} metrics assess the naturalness of the dialogue by measuring key speech characteristics.

\begin{itemize}
    \item \textbf{Input}: The synthetic transcript.
    \item \textbf{Evaluation}: An LLM assesses the transcript for three characteristics of natural human speech:
    \begin{enumerate}
        \item \textbf{Interruptions}: The presence of natural, well-placed speaker overlaps.
        \item \textbf{Disfluencies}: The use of filler words, hesitations, and self-corrections.
        \item \textbf{ASR Noise}: The inclusion of plausible artifacts typical of Automatic Speech Recognition systems (e.g., homophone errors).
    \end{enumerate}
\end{itemize}

Each characteristic is scored from 1 to 10. These three scores are then averaged to produce the \textit{Average Score per Speech Characteristic}, \(S_{Speech\_Char_{raw}}\). The prompt used for calculating this metric can be found in Table \ref{prompts_for_transcript_generation_3}.

\subsubsection{Score Normalization and Aggregation}
\label{reconstruction_score_calculation}

To combine these diverse metrics into a single Reconstruction Score, we first normalize them to a common scale of [0, 1]. Scores originally on a 1--10 scale are normalized using min-max scaling:

\[
S_{norm} = \frac{S_{raw} - 1}{9}
\]

This is applied to \(S_{TS_{raw}}\), \(S_{KE_{raw}}\), \(S_{Summ\_Intent_{raw}}\), and \(S_{Speech\_Char_{raw}}\). The QA score, \(S_{QA}\), is already in a [0, 1] range and requires no normalization.

The final \textbf{Reconstruction Score} is a weighted sum of the normalized component scores, reflecting their relative importance in our evaluation framework:

{\small
\begin{align*}
\text{Score}_{\text{Recon}} ={}&\  w_{TS} \cdot S_{TS_{\text{norm}}} + w_{QA} \cdot S_{QA} 
+ w_{KE} \cdot S_{KE_{\text{norm}}} \\ & + w_{Summ} \cdot S_{Summ\_Intent_{\text{norm}}} \\
& + w_{Speech} \cdot S_{Speech\_Char_{\text{norm}}}
\end{align*}
}

\noindent Where the weights are defined as:
\begin{itemize}
    \item \(w_{TS} = 0.25\) (Topic Sequence)
    \item \(w_{QA} = 0.15\) (QA Scenario)
    \item \(w_{KE} = 0.25\) (Key Events)
    \item \(w_{Summ} = 0.15\) (Average Summary Intent)
    \item \(w_{Speech} = 0.20\) (Average Speech Characteristic)
\end{itemize}

The composite score was tuned on the test split of the prompt tuning dataset and was also used to optimize prompts for the generation pipelines in Section~\ref{transcript_gen_pipeline}. It balances fidelity to source attributes with linguistic realism to provide a holistic measure of synthetic transcript quality.

\subsection{Methodology - Downstream Evaluation}
\label{downstream_eval_methodology}

\begin{enumerate}
    \item \textbf{Real downstream dataset construction.}
    Following the process in Section~\ref{dataset}, we mine a set of real production calls across four languages.
    Each call $i$ yields a transcript $x_i^{\text{real}}$ and a set of \emph{input call attributes} $A_i$ used by the generation pipelines.
    The attributes $A_i$ include a QA form containing question--answer pairs $\{(q_{ij}, y_{ij})\}_j$.
    We flatten each call into triplets $(x_i^{\text{real}}, q_{ij}, y_{ij})$ and define the real downstream dataset
    \[
        \mathcal{D}^{\text{real}} \;=\; \{(x_i^{\text{real}}, q_{ij}, y_{ij})\}_{i,j}.
    \]
    The task is framed as a classification problem where the model must select the correct label $y_{ij}$ from the predefined options. We normalize answer strings into $\{\texttt{Yes},\texttt{No}\}$ (language-specific variants mapped to canonical labels) and discard any example where $y_{ij}\notin\{\texttt{Yes},\texttt{No}\}$.

    \item \textbf{Train/validation/test split.}
    We randomly split $\mathcal{D}^{\text{real}}$ into disjoint sets
    \[
        \mathcal{D}_{\text{train}}^{\text{real}},\quad
        \mathcal{D}_{\text{val}}^{\text{real}},\quad
        \mathcal{D}_{\text{test}}^{\text{real}}.
    \]
    The test set $\mathcal{D}_{\text{test}}^{\text{real}}$ is held fixed and shared across all methods.

    \item \textbf{Synthetic transcript generation (how the synthetic datasets are created).}
    Let $g_m(\cdot)$ denote a transcript generation pipeline, where
    \[
        m \in \{\textsf{direct},\ \textsf{chunked},\ \textsf{char-aware}\}.
    \]
    For each \emph{underlying real call} $i$ that appears in $\mathcal{D}_{\text{train}}^{\text{real}}$, we generate a synthetic transcript by conditioning on the same input call attributes:
    \[
        x_i^{m} \;=\; g_m(A_i).
    \]
    We then \emph{reuse the exact same questions and labels} from the real call $i$ and swap only the transcript, producing the synthetic training set
    \[
        \mathcal{D}_{\text{train}}^{m}
        \;=\;
        \{(x_i^{m},\, q_{ij},\, y_{ij}) : (x_i^{\text{real}}, q_{ij}, y_{ij}) \in \mathcal{D}_{\text{train}}^{\text{real}}\}.
    \]
    Thus, across $\mathcal{D}_{\text{train}}^{\text{real}}$ and $\{\mathcal{D}_{\text{train}}^{m}\}_m$, the supervision $(q_{ij},y_{ij})$ and call-level attributes $A_i$ are aligned, and only the transcript differs.

    \item \textbf{Prompt optimization.}
    We optimize an AutoQA prompting program separately for each training source
    \[
        S \in \Big\{\mathcal{D}_{\text{train}}^{\text{real}},\ \mathcal{D}_{\text{train}}^{\textsf{direct}},\ \mathcal{D}_{\text{train}}^{\textsf{chunked}},\ \mathcal{D}_{\text{train}}^{\textsf{char-aware}}\Big\}
    \]
    across a set of diverse model variants $\mathcal{M} = \{ \text{GPT-4.1-mini}, \text{GPT-4.1}, \text{Claude Haiku}, \text{GPT-4o} \}$ using DSPy MIPROv2, with hyperparameters reported in Table~\ref{tab:dspy_hyperparameters}.
    All optimizations use the same validation set $\mathcal{D}_{\text{val}}^{\text{real}}$ for model selection.

    \item \textbf{Evaluation.}
    Each resulting prompt (including the unoptimized baseline) is evaluated on the common held-out test set $\mathcal{D}_{\text{test}}^{\text{real}}$ for each model variant in $\mathcal{M}$.
    We report macro F1 scores averaged across all model variants in $\mathcal{M}$ to provide a model-agnostic indicator of downstream generalization.
\end{enumerate}

\subsection{Methodology - Evaluation pipeline}
\label{evaluation_pipeline_detailed}
To evaluate how well synthetic transcripts capture the deeper, implicit qualities of real call center conversations, we propose a dedicated analysis framework focused on latent conversational characteristics. These are properties not explicitly provided as input during generation—such as emotion shifts, discourse flow, and speaker behavior patterns—but are integral to the realism and authenticity of natural conversations. The framework assesses whether the synthetic data mirrors the nuanced statistical distribution of these latent features in real, human-generated transcripts. The analysis proceeds in four stages: (1) programmatic transcript chunking for contextual analysis, (2) LLM-based annotation of turn-level and transcript-level latent features, (3) construction of empirical frequency distributions, and (4) statistical comparison of these distributions between synthetic and real data.

\subsubsection{Transcript Processing and Chunking}
The foundation of the analysis rests on a comprehensive evaluation of all conversational turns. Let the corpus of real transcripts be denoted by $\mathcal{T}_R = \{t_{r,1}, t_{r,2}, \dots, t_{r,N}\}$ and the corresponding set of synthetic transcripts be $\mathcal{T}_S = \{t_{s,1}, t_{s,2}, \dots, t_{s,N}\}$. Each transcript $t$ is an ordered sequence of turns, $t = (u_1, u_2, \dots, u_{m})$, where $m$ is the total number of turns.

To handle long transcripts while maintaining local context for turn-level analysis, each transcript is programmatically segmented into contiguous, non-overlapping chunks. This ensures that every turn is analyzed. The chunking algorithm divides the sequence of $m$ turns into $n$ chunks, where the size of each chunk is between a predefined minimum $c_{\min}$ and maximum $c_{\max}$ number of turns. This approach avoids biases from transcripts of varying lengths and respects the context limitations of the language model.

For contextual understanding, each chunk of turns $(u_j, \dots, u_k)$ is presented to the classification model along with its surrounding conversational context. A context window of size $w$ is used, providing the model with the preceding turns $(u_{j-w}, \dots, u_{j-1})$ and the succeeding turns $(u_{k+1}, \dots, u_{k+w})$. This allows the model to make more informed judgments based on the immediate dialog flow.

\subsubsection{LLM-based Latent Feature Classification}
A comprehensive, multi-dimensional taxonomy of conversational characteristics is employed. This taxonomy, $\mathcal{C}$, consists of $D$ distinct dimensions, $\mathcal{C} = \{C_1, C_2, \dots, C_D\}$. Each dimension represents a specific aspect of the conversation. The analysis is performed at two granularities: turn-level and transcript-level.

\paragraph{Turn-Level Characteristics}
Each dimension $C_d$ is defined by a set of $V_d$ discrete categories, $\{c_{d,1}, c_{d,2}, \dots, c_{d,V_d}\}$. For most dimensions, these categories are mutually exclusive (single-label classification). However, for complex phenomena like conversational disfluencies, multiple categories can be assigned to a single turn (multi-label classification).

The core of the classification is a Large Language Model (LLM) that functions as a classifier, $f_{\text{LLM}}$. For each turn $u_i$ within a chunk, the model assigns a category or a set of categories for each dimension based on the chunk and its context.
\begin{itemize}
    \item For single-label dimensions: $f_{\text{LLM}}(u_i, \text{context}) \rightarrow c_{d,j}$ where $c_{d,j} \in C_d$.
    \item For multi-label dimensions: $f_{\text{LLM}}(u_i, \text{context}) \rightarrow \mathcal{P}_j \subseteq C_d$.
\end{itemize}
The taxonomy includes characteristics such as sentiment, proactivity, disfluency, and question type.

\paragraph{Transcript-Level Characteristics}
In addition to turn-level analysis, the framework assesses transcript-level characteristics to capture holistic conversational properties. A separate LLM-based function, $g_{\text{LLM}}$, analyzes the full transcript text to derive two types of metrics:
\begin{itemize}
    \item \textbf{Readability Scores:} A set of metrics $\mathcal{R}$ (vocabulary complexity, sentence complexity, technical density, discourse flow, and overall readability) are scored on a 1-5 scale. For a transcript $t$, $g_{\text{LLM}}(t) \rightarrow \mathbf{s} \in \{1, \dots, 5\}^{|\mathcal{R}|}$.
    \item \textbf{Emotion and Sentiment Arcs:} The trajectory of emotion and sentiment for both customer and agent from the beginning to the end of the conversation is captured as a descriptive string (e.g., "neutral to positive"). For a set of arc metrics $\mathcal{A}$, $g_{\text{LLM}}(t) \rightarrow \{a_1, a_2, \dots, a_{|\mathcal{A}|}\}$.
\end{itemize}

\subsubsection{Empirical Frequency Distribution Construction}
Following the classification, the results are aggregated to construct empirical frequency distributions for each conversational dimension. Let $\mathcal{U}_R = \bigcup_{t \in \mathcal{T}_R} t$ and $\mathcal{U}_S = \bigcup_{t \in \mathcal{T}_S} t$ be the complete sets of turns from real and synthetic transcripts, respectively.

For a given turn-level, single-label dimension $C_d$, the observed frequency of a category $c_{d,j}$ in the real transcripts is the count of turns assigned to that category:
\begin{equation}
O_{d,j} = \sum_{u \in \mathcal{U}_R} \mathbb{I}(f_{\text{LLM}}(u) = c_{d,j})
\end{equation}
where $\mathbb{I}(\cdot)$ is the indicator function. This produces a frequency vector for the real data, $\mathbf{O}_d = (O_{d,1}, O_{d,2}, \dots, O_{d,V_d})$. Similarly, a frequency vector is constructed for the synthetic data, $\mathbf{E}_d = (E_{d,1}, E_{d,2}, \dots, E_{d,V_d})$. The total number of observations is $N_R = \sum_{j=1}^{V_d} O_{d,j}$ and $N_S = \sum_{j=1}^{V_d} E_{d,j}$. For multi-label dimensions, frequencies are counted independently.

Similarly, for transcript-level metrics (readability scores and emotion arcs), frequencies are calculated by counting the occurrences of each category or score across the entire set of transcripts.

\subsubsection{Statistical Comparison of Distributions}
The goal is to quantify the similarity between the frequency distributions of real and synthetic data for each dimension $C_d$. The null hypothesis $H_0$ assumes both distributions are statistically indistinguishable.

\paragraph{Pearson's Chi-squared (\(\chi^2\)) Test}
This test evaluates the goodness-of-fit between observed frequencies from real data ($\mathbf{O}_d$) and expected counts scaled from synthetic data:
\begin{align}
\chi^2_d = \sum_{j=1}^{V_d} \frac{(O_{d,j} - E'_{d,j})^2}{E'_{d,j}}, \nonumber \\
\text{where} \quad E'_{d,j} = E_{d,j} \cdot \frac{N_R}{N_S}
\end{align}
The statistic follows a chi-squared distribution with $V_d - 1$ degrees of freedom. A low p-value indicates significant distributional differences.

\paragraph{G-test (Likelihood-Ratio Test)}
An alternative to the chi-squared test, the G-test uses the log-likelihood ratio:
\begin{equation}
G_d = 2 \sum_{j=1}^{V_d} O_{d,j} \ln\left(\frac{O_{d,j}}{E'_{d,j}}\right)
\end{equation}
It shares the same asymptotic distribution and interpretability as the $\chi^2$ statistic but can be more accurate for small sample sizes.

\paragraph{Jensen-Shannon (JS) Divergence}
To assess distributional similarity more directly, we compute the JS divergence between the normalized probability distributions $\mathbf{P}_d$ and $\mathbf{Q}_d$ (derived from $\mathbf{O}_d$ and $\mathbf{E}_d$):
\begin{align}
D_{\text{JS}}(\mathbf{P}_d || \mathbf{Q}_d) = \frac{1}{2} D_{\text{KL}}(\mathbf{P}_d || \mathbf{M}_d) + \frac{1}{2} D_{\text{KL}}(\mathbf{Q}_d || \mathbf{M}_d)
\end{align}
where $\mathbf{M}_d = \frac{1}{2}(\mathbf{P}_d + \mathbf{Q}_d)$ is the midpoint distribution and $D_{\text{KL}}$ is the Kullback-Leibler divergence. The JS score ranges from 0 (identical distributions) to 1 (fully dissimilar), offering an intuitive fidelity measure.

\subsection{Baselines}
\label{baselines}
Our work builds upon two prior methodologies for synthetic conversation generation, adapting them to the specific context of call center transcripts.

NoteChat \citep{Wang_2024}: This framework was originally designed to generate synthetic patient-physician dialogues from clinical notes using a three-stage pipeline: Planning, Roleplay, and Polish. For our adaptation, we re-contextualized this system for customer service interactions. The input was changed from clinical notes to the structured call attributes (e.g., reason for call, key entities), and the roles were modified to "Agent" and "Customer." The core logic of systematically covering predefined entities was preserved, while the final Polish module was repurposed to inject authentic spoken artifacts like disfluencies and simulated ASR errors.

ConvoGen \citep{gody2025convogenenhancingconversationalai}: This approach leverages the AutoGen \citep{wu2023autogenenablingnextgenllm} framework to simulate diverse, multi-party conversations by first generating experiences (personas, context) and then having agents role-play within that scenario. We specialized this flexible system for the two-party structure of a customer service call. Instead of generating open-ended experiences, our implementation uses predefined call attributes to establish the call context. Furthermore, we enforced a strict, turn-by-turn, single-sentence dialogue format and integrated instructions for generating disfluencies and ASR noise directly into the agents' prompts to produce a realistic call flow.

\subsection{Evaluation Metrics: Full Descriptions} \label{eval_metric_detailed_desc}

Evaluating the quality of synthetic conversational data is challenging, as traditional metrics like BLEU or ROUGE do not capture interaction nuances. We present a multi-dimensional evaluation framework assessing emotional arcs, linguistic complexity, interaction styles, and conversational properties. These metrics are computed using DSPy modules that analyze transcript chunks (for turn-level metrics) or the full transcript (for transcript-level metrics), employing language model predictions with predefined signatures.

\subsubsection{Emotional \& Sentiment Metrics}
These metrics capture affective characteristics at transcript and turn levels, obtained by analyzing beginning/ending segments or individual turns.

\begin{enumerate}[leftmargin=5.25mm, itemsep=0.5mm]
    \item \textit{Agent \& Customer Emotion Arc}: Tracks emotional trajectory from conversation start to end. Obtained by classifying emotions in beginning and ending segments (first/last 5 turns). Categories: \texttt{gratitude} (appreciation), \texttt{relief} (stress lifted), \texttt{factual} (objective), \texttt{curiosity} (seeking info), \texttt{confusion} (uncertainty), \texttt{frustration} (irritation), \texttt{anger} (hostile), \texttt{anxiety} (worried). Example: frustration to relief -- customer starts irritated but ends satisfied after resolution.
    
    \item \textit{Agent \& Customer Sentiment Arc}: Maps emotion arcs to positive/neutral/negative. Derived from emotion classifications.
    
    \item \textit{Sentiment (Turn-Level)}: Classifies each turn's tone. Obtained per turn. Categories: \texttt{positive\_sentiment} (pleased, e.g., "Thank you!"), \texttt{neutral\_sentiment} (factual, e.g., "What's the status?"), \texttt{negative\_sentiment} (annoyed, e.g., "I'm frustrated"), \texttt{N/A}.
\end{enumerate}

\subsubsection{Linguistic Complexity \& Content Density}
Measures language richness and accessibility, with turn-level classification and transcript-level scores (1-5, where 1 is complex/hard and 5 is simple/easy). Turn-level obtained per turn; transcript-level obtained on full transcript.

\begin{enumerate}[leftmargin=5.25mm, itemsep=0.5mm]
    \item \textit{Language Complexity (Turn-Level)}: Categories: \texttt{simple\_informal\_language} (conversational, e.g., "Yeah, no worries"), \texttt{simple\_formal\_language} (professional plain, e.g., "I have processed your request"), \texttt{complex\_informal\_language} (informal with jargon, e.g., "Awesome, you passed KYC"), \texttt{complex\_formal\_language} (formal complex, e.g., "Pursuant to the agreement"), \texttt{N/A}.

    \item \textit{Technical Density}: Transcript score measuring jargon prevalence.
    
    \item \textit{Sentence Complexity}: Transcript score evaluating structural complexity.
    
    \item \textit{Discourse Flow}: Transcript score assessing coherence.
    
    \item \textit{Overall Readability Score}: Combined transcript metric.
\end{enumerate}

\subsubsection{Interaction Style and Operational}
Assesses conversation management and resolution effectiveness, obtained per turn.

\begin{enumerate}[leftmargin=5.25mm, itemsep=0.5mm]
    \item \textit{Proactivity (Agent Turns)}: Agent's initiative level. Categories: \texttt{neutral\_proactivity} (appropriate, e.g., direct answer), \texttt{overstated\_proactivity} (excessive, e.g., unsolicited reassurance), \texttt{understated\_proactivity} (passive, e.g., minimal response), \texttt{N/A}.
    
    \item \textit{Emphasis}: Turn's focus. Categories: \texttt{emotion\_focused} (subjective, e.g., "I'm confused why"), \texttt{fact\_focused} (objective, e.g., "The rate is..."), \texttt{balanced} (mixed, e.g., "I expected the refund but was declined"), \texttt{N/A}.
    
    \item \textit{Question Type}: Classifies questions. Categories: \texttt{no\_question} (statement, e.g., "I've noted that"), \texttt{closed\_question} (yes/no, e.g., "Do you have bills?"), \texttt{informational\_question} (facts, e.g., "Pricing details?"), \texttt{conversational\_question} (rapport, e.g., "How can I assist?"), \texttt{N/A}.
    
    \item \textit{Solution}: Agent's contribution to resolving issues. Categories: \texttt{solution\_oriented} (direct fix, e.g., "Contact fraud department"), \texttt{process\_oriented} (logistics, e.g., "Schedule consultation"), \texttt{no\_solution} (acknowledgment, e.g., "Okay"), \texttt{N/A}.
\end{enumerate}

\subsubsection{Conversational Properties}
Captures naturalness and surface features, obtained per turn (multi-select for some).

\begin{enumerate}[leftmargin=5.25mm, itemsep=0.5mm]
    \item \textit{Repetition}: Information repetition patterns. Categories: \texttt{no\_repetition} (new info, e.g., providing details), \texttt{agent\_repetition} (agent restates, e.g., "Confirm address?"), \texttt{customer\_repetition} (customer restates, e.g., "You said click the link?"), \texttt{N/A}.
    
    \item \textit{Disfluency}: Detects speech disruptions (multi-select). Categories: \texttt{speech\_repair\_repetition} (corrections, e.g., "no no it's..."), \texttt{hesitation\_fillers} (pauses, e.g., "um..."), \texttt{interactional\_disfluency} (interruptions, e.g., "hold on—"), \texttt{comprehension\_clarity\_issues} (misunderstandings, e.g., "Sorry, repeat?"), \texttt{no\_disfluency} (clean, e.g., "Yes that's right"), \texttt{N/A}.
    
    \item \textit{ASR Noise Type}: Simulates transcription errors (multi-select). Categories: \texttt{no\_noise} (accurate, e.g., "I need help"), \texttt{substitution} (replacement, e.g., "older" for "order"), \texttt{insertion} (extra, e.g., "my my account"), \texttt{deletion} (missing, e.g., "need account"), \texttt{N/A}.
\end{enumerate}

This framework enables quantitative comparison of generation methods by analyzing synthetic transcripts holistically.

\subsection{Implementation Details}
\label{implementation_details}

Our synthetic transcript generation pipeline is implemented in Python, leveraging several open-source libraries and large language models (LLMs). The entire pipeline is designed to run on a local machine, ensuring reproducibility and control over the environment.

\subsubsection*{Prompt Engineering and Optimization with DSPy}

A core component of our methodology is the use of DSPy \cite{khattab2023dspycompilingdeclarativelanguage}, a framework for programming with foundation models. Instead of manually crafting prompts, we define the steps of our pipeline as \texttt{DSPy} modules and use its automatic optimization capabilities to generate and refine high-quality prompts.

This process involves a \texttt{MIPROv2} (Multi-prompt Instruction-driven Program Optimizer) optimizer \cite{opsahlong2024optimizinginstructionsdemonstrationsmultistage}, which explores various prompt candidates to maximize a given metric. The optimizer was configured with specific hyperparameters to guide the search for the most effective prompts for our generation task. The key \texttt{DSPy} hyperparameters used for prompt optimization are detailed in Table~\ref{tab:dspy_hyperparameters}.

\subsubsection*{LLM and Infrastructure}

The language models used in this research were accessed via \texttt{Bedrock} and \texttt{LiteLLM}, which provide a unified interface to various model providers. This setup allows for flexibility and easy substitution of models. Table~\ref{tab:llm_pipeline_parameters} provides a comprehensive list of the LLMs used in different stages of our pipeline---from initial data generation to final evaluation---along with their default parameters. The entire pipeline, including model inference and optimization, was executed on local infrastructure.

\end{document}